\documentclass{article}

\usepackage{iclr2027_conference,times}

\usepackage[utf8]{inputenc}
\usepackage[T1]{fontenc}
\usepackage{hyperref}
\usepackage{url}
\usepackage{booktabs}
\usepackage{amsfonts}
\usepackage{amsmath}
\usepackage{amssymb}
\usepackage{mathtools}
\usepackage{amsthm}
\usepackage{xspace}
\usepackage{amsmath}
\usepackage{nicefrac}
\usepackage{ragged2e}
\usepackage{microtype}
\usepackage{xcolor}
\usepackage{algorithm}
\usepackage{algorithmic}
\usepackage{graphicx}
\usepackage{multirow}
\usepackage{tabularx}
\usepackage{array}
\usepackage[most]{tcolorbox}
\usepackage{wrapfig}
\usepackage{float}
\usepackage{amsthm}
\usepackage{bbm}
\usepackage{changepage}
\usepackage{enumitem}
\usepackage{listings}
\usepackage{subcaption}
\usepackage{adjustbox}
\definecolor{sqrtred}{HTML}{C44E52}
\definecolor{adamwblue}{HTML}{4C72B0}
\definecolor{inkgrey}{HTML}{333333}
\newtcolorbox{takeaway}[2][]{enhanced, breakable, colback=#2!5, colframe=#2!75!black,
  boxrule=0.4pt, left=6pt, right=6pt, top=5pt, bottom=5pt, arc=1.5pt,
  borderline west={2pt}{0pt}{#2}, coltitle=white, fonttitle=\bfseries\small,
  colbacktitle=#2!85!black, title={#1}, before skip=7pt, after skip=7pt}

\usepackage{tikz}

\definecolor{critfill}{RGB}{255,249,210}

\newcounter{criterion}

\definecolor{boxblue}{RGB}{220,235,252}
\definecolor{boxgreen}{RGB}{220,248,228}
\definecolor{boxorange}{RGB}{253,237,213}
\definecolor{boxred}{RGB}{252,224,224}
\definecolor{boxyellow}{RGB}{255,249,210}

\definecolor{DarkPink}{rgb}{0.5,0.0,0.18}
\definecolor{DarkGreen}{rgb}{0.1,0.6,0.1}
\definecolor{DarkRed}{rgb}{0.5,0.1,0.1}
\definecolor{DarkBlue}{rgb}{0.1,0.1,0.7}
\definecolor{DarkYellow}{rgb}{.79,.79,0}
\definecolor{Cyan}{cmyk}{5,0,3,0}
\hypersetup{
    unicode=false,          %
    pdftoolbar=true,        %
    pdfmenubar=true,        %
    pdffitwindow=false,      %
    pdfnewwindow=true,      %
    colorlinks=true,       %
    linkcolor=DarkGreen,          %
    citecolor=DarkRed,        %
    filecolor=DarkRed,      %
    urlcolor=DarkBlue,          %
    pdftitle={},
    pdfauthor={},
}

\newtheorem{theorem}{Theorem}
\newtheorem{proposition}{Proposition}
\newtheorem{lemma}{Lemma}
\newtheorem{corollary}{Corollary}

\newtheorem{remark}{Remark}

\newtheorem{assumption}{Assumption}

\newcommand{\myparagraph}{\textbf}
\newcommand{\AdEMAMix}{\texttt{AdEMAMix}\xspace}
\newcommand{\AdamW}{\texttt{AdamW}\xspace}
\newcommand{\Muon}{\texttt{Muon}\xspace}
\newcommand{\Lion}{\texttt{Lion}\xspace}
\newcommand{\Signum}{\texttt{Signum}\xspace}
\newcommand{\SignSGD}{\texttt{SignSGD}\xspace}
\newcommand{\Sophia}{\texttt{Sophia}\xspace}
\newcommand{\MARS}{\texttt{MARS}\xspace}

\newcommand{\SOAP}{\texttt{SOAP}\xspace}
\newcommand{\ADOPT}{\texttt{ADOPT}\xspace}
\newcommand{\DMuon}{\texttt{D-Muon}\xspace}

\newcommand{\SFAdamW}{\texttt{SF-AdamW}\xspace}
\newcommand{\Prodigy}{\texttt{Prodigy}\xspace}
\newcommand{\Adam}{\texttt{Adam}\xspace}
\newcommand{\SGD}{\texttt{SGD}\xspace}

\newcommand{\btheta}{\boldsymbol{\theta}}
\newcommand{\bg}{\mathbf{g}}
\newcommand{\bu}{\mathbf{u}}
\newcommand{\bmv}{\mathbf{m}}
\newcommand{\bv}{\mathbf{v}}
\newcommand{\bhv}{\mathbf{h}}
\newcommand{\bc}{\mathbf{c}}

\newcommand{\bb}{\mathbf{b}}
\newcommand{\bs}{\mathbf{s}}
\newcommand{\bd}{\mathbf{d}}
\newcommand{\bx}{\mathbf{x}}

\newcommand{\bW}{\mathbf{W}}
\newcommand{\bP}{\mathbf{P}}
\newcommand{\bI}{\mathbf{I}}
\newcommand{\bU}{\mathbf{U}}
\newcommand{\bV}{\mathbf{V}}
\newcommand{\bH}{\mathbf{H}}
\newcommand{\bG}{\mathbf{G}}
\newcommand{\bM}{\mathbf{M}}
\newcommand{\bN}{\mathbf{N}}
\newcommand{\bO}{\mathbf{O}}

\newcommand{\bR}{\mathbf{R}}
\newcommand{\bC}{\mathbf{C}}
\newcommand{\bS}{\mathbf{S}}
\newcommand{\bD}{\mathbf{D}}
\newcommand{\bQ}{\mathbf{Q}}

\newcommand{\bX}{\mathbf{X}}

\newcommand{\bK}{\mathbf{K}}

\newcommand{\cS}{\mathcal{S}}
\newcommand{\cB}{\mathcal{B}}
\newcommand{\cD}{\mathcal{D}}
\newcommand{\cL}{\mathcal{L}}
\newcommand{\cV}{\mathcal{V}}
\newcommand{\cO}{\mathcal{O}}

\newcommand{\E}{\mathbb{E}}
\newcommand{\R}{\mathbb{R}}

\newcommand{\diag}{\mathrm{diag}}
\newcommand{\sign}{\mathrm{sign}}
\newcommand{\clip}{\mathrm{clip}}

\newcommand{\NS}{\mathrm{NS}}

\newcommand{\Var}{\mathrm{Var}}
\newcommand{\Cov}{\mathrm{Cov}}
\newcommand{\norm}[1]{\left\lVert #1 \right\rVert}
\newcommand{\inner}[2]{\langle #1,\, #2 \rangle}

\DeclareMathOperator{\msign}{msign}

\newcommand{\bigO}{\mathrm{O}}       

\newcommand{\dd}{d} 

\providecommand{\bA}{\mathbf{A}}
\providecommand{\bB}{\mathbf{B}}

\newcolumntype{Y}{>{\raggedright\arraybackslash}X}
\newcommand{\labelw}{0.26\linewidth}                        
\newcolumntype{L}{>{\raggedright\arraybackslash}p{\labelw}} 
\newcolumntype{Y}{>{\raggedright\arraybackslash}X}          

\title{\textit{OptiSelect}: How does \\the Optimizer Shape Data Curriculum?}

\author{Simin Fan$^{\star}$\quad Alireza Abdollahpoorrostam$^{\star}$\quad Martin Jaggi\\[3pt]
EPFL\\
\normalfont\small $^{\star}$Equal contribution}

\iclrfinalcopy 
\begin{document}

\maketitle
\vspace{-2em}
\begin{abstract}
\vspace{-1em}
Online data selection has demonstrated substantial efficiency gains for LLM pretraining by training on the most valuable candidates within each batch.
Since a candidate's value is realized through its effective model update, principled selection should account for the optimizer step, which reshapes the raw gradient before it updates model parameters. 
We formalize this optimizer-aware selection paradigm as \textit{OptiSelect} and present the first systematic study of how the optimizer shapes data selection.
Our theory establishes a \textit{selection gain principle} in which the advantage of online selection is governed by the \textit{discriminability} of the optimizer-induced utility scores.
We prove that sign-based and polar-tangential preconditioners of \Lion and \Muon would suffer from a \textit{discriminability collapse} which caps attainable gains from \textit{OptiSelect}, whereas diagonal-adaptive optimizers such as \AdamW and \Sophia admit strictly better upper bounds. 
The proposed principle also yields a quantitative derivation of the optimal candidate oversampling ratio.
Pretraining experiments on 124M and 720M models are consistent with our theoretical analysis
and show that \AdamW's diagonal-adaptive scoring geometry remains the strongest scoring geometry even with \Muon as optimizer. 
We further demonstrate that \textit{OptiSelect} retains its benefits under data rephrasing, a technique used in modern data processing pipelines.
Our findings provide theoretical foundations and practical guidance for co-designing optimizers and data selection in LLM pretraining.
\end{abstract}
\vspace{-1.5em}
\section{Introduction}
\label{sec:intro}
\vspace{-1em}
Data recipes and optimizers are two important levers determining the efficiency of LLM pretraining: they define what the model learns from and how it learns, respectively. 
These two design choices are often considered independently. Data curation pipelines filter or reweight the corpus before training begins, agnostic to the optimizer that will process it \citep{xie2023doremi, fan2024doge, li2024dclm, penedo2024fineweb, wang2024greats}, while optimizer benchmarks compare optimization algorithms on uniformly sampled data streams, agnostic to data selection strategies \citep{semenov2025benchmarking, wen2025fantasticpretrainingoptimizers}. 
The \textit{optimizer-aware online data selection} strategy \citep{wang2026opus} combines these two considerations, which we formalize as the \textit{OptiSelect} training paradigm: At every optimization step, each data sample from an oversampled candidate pool is scored by its marginal utility, and only the highest-utility batch is selected for  training. 
As a sample's utility is realized through the model update it induces, the data utilities should be computed with the optimizer preconditioner, which reshapes the raw gradient before updating model parameters. 
Therefore, the optimizer plays a dual role under \textit{OptiSelect}. It descends the loss landscape strategically while also designing the data curriculum, adapting the data utility scoring function according to its unique preconditioner. 

We analyze the interaction between optimizers and data selection across five optimizers spanning different preconditioning geometries: \AdamW rescales gradient coordinates by their estimated RMS \citep{loshchilov2019adamw}, \AdEMAMix mixes in a slow momentum buffer \citep{pagliardini2024ademamix}, \Sophia normalizes by estimated curvature \citep{liu2023sophia}, \Lion discards magnitudes and keeps only signs \citep{chen2023lion}, and \Muon orthogonalizes each update matrix \citep{jordan2024muon}.
Our study is organized around one central question from both theoretical and empirical perspectives: \looseness=-1
\begin{center}
\vspace{-0.5em}
\textit{How does the optimizer interact with online data selection,\\
and which optimization geometries can actually profit from it?}
\vspace{-0.5em}
\end{center}

\textbf{Contributions.} We summarize our contributions as three-fold.
\begin{itemize}[wide=0.65\parindent,topsep=1pt,itemsep=3pt]
\item \textbf{The \textit{OptiSelect} framework (\S\ref{sec:optiselect}).} We formalize the \textit{OptiSelect} training paradigm through an optimizer-aware \emph{transformation operator} $\cO_t$ and derive the corresponding optimizer-aware data utility functions across five optimizers spanning various geometries.
\item \textbf{A principled theory of the optimizer--selection interplay (\S\ref{sec:theory}).} We establish a \textit{selection gain principle}, demonstrating that the one-step advantage from data selection scales with the \textit{discriminability} of the utility scores (\autoref{thm:selection_gain}). 
We prove that a \textit{sign collapse} bounds the gain of \Lion and a \textit{radial collapse} bounds that of \DMuon, whereas \AdamW and \Sophia retain strictly larger score spread. The same principle also derives the optimum candidate oversampling ratio as $\lambda^\star = 2$.
\item \textbf{A comprehensive pretraining benchmark (\S\ref{sec:experiments}).} We evaluate five optimizers under standard training and \textit{OptiSelect} on 124M and 720M models, demonstrating a close agreement between measured gains and the theory. 
Decoupling the scoring geometry from the base optimizer reveals that the optimizer-matched geometry is not necessarily optimal, with \AdamW's diagonal-adaptive operator prevailing under \DMuon. The selection benefit persists under format rephrasing.
\end{itemize}

\vspace{-0.5em}
\section{Related Work}
\label{sec:related}
\vspace{-0.5em}
We position each work briefly in this section and provide a more detailed survey in Appendix~\ref{app:related_extended}.\looseness=-1

\textbf{Data selection for LLM pretraining.}
Research on training-data ordering and sample weighting predates LLMs, e.g. easy-to-hard curricula~\citep{bengio2009curriculum}, online loss-ranked sampling~\citep{loshchilov2015online}, importance sampling with learned gradient-norm proxies~\citep{katharopoulos2019samplescreatedequaldeep} whereas none of these strategies explicitly accounts for the optimizer.
At LLM scale, selection is dominated by \emph{static} pipelines that curate the corpus before training begins: hashed n-gram resampling (DSIR;~\citealp{xie2023dsir}), fastText quality classifiers (DCLM;~\citealp{li2024dclm}), distilled LLM quality judgments (FineWeb-Edu;~\citealp{penedo2024fineweb}), and proxy-model domain weights (DoReMi;~\citealp{xie2023doremi}, DoGE;~\citealp{fan2024doge}).
The online strategies include MATES~\citep{yu2024mates}, which predicts influence with a periodically refreshed auxiliary model; GREATS~\citep{wang2024greats} and OPUS~\citep{wang2026opus}, which compute gradient-alignment scores and optimizer-aware scores, respectively. \looseness=-1

\textbf{Optimizer-aware influence.}
Influence functions~\citep{koh2017understanding}, the first-order influence TracIn~\citep{pruthi2020tracin} and TRAK with random projections~\citep{park2023trak} attributed model behaviour to training samples under implicit \SGD\ dynamics.
Adaptive optimizer state changes influence: LESS~\citep{xia2024less} and TrackStar~\citep{chang2024scalableinfluencefacttracing} correct influence scores for \Adam's second moment; OPUS~\citep{wang2026opus} computes optimizer-aware attribution scores on \AdamW\ and \Muon, and uses them for per-step sample selection during LLM pretraining.
Yet, none compares the influence geometries different optimizers induce, which is the object of our theory (\S\ref{sec:theory}).
\begin{wrapfigure}{r}{0.5\textwidth}
\vspace{-1.5em}
\begin{minipage}{0.5\textwidth}
\begin{algorithm}[H]
\caption{\small \textit{OptiSelect}: one pretraining step with optimizer-aware data selection}
\label{alg:opus}
\small
\begin{algorithmic}[1]
\REQUIRE Candidate batch $\tilde{\cB}_t$, parameters $\btheta_t$, optimizer state, target batch size $B$, temperature $\tau$
\ENSURE Updated parameters $\btheta_{t+1}$
\STATE \textbf{// Candidate forward-backward pass}
\FOR{each $z \in \tilde{\cB}_t$}
    \STATE Compute per-sample gradient $g_z \gets \nabla_{\btheta}\, \ell(z;\btheta_t)$
\ENDFOR
\STATE \textbf{// Influence scoring}
\FOR{each $z \in \tilde{\cB}_t$}
    \STATE Assign utility $U(z, t)$ using $g_z$ and the current optimizer state
\ENDFOR
\STATE \textbf{// Selection via Boltzmann distribution}
\STATE Define $p_t(z) \propto \exp\!\big(U(z, t)/\tau\big)$ for $z \in \tilde{\cB}_t$
\STATE Sample $\cB_t \subset \tilde{\cB}_t$ with $|\cB_t| = B$ according to $p_t$
\STATE \textbf{// Optimizer update}
\STATE $\btheta_{t+1} \gets \textsc{OptimizerStep}\!\left(\btheta_t,\, \{g_z\}_{z \in \cB_t}\right)$
\RETURN $\btheta_{t+1}$
\end{algorithmic}
\end{algorithm}
\vspace{-2em}
\end{minipage}
\vspace{-\intextsep}
\vspace{-1em}
\end{wrapfigure}
\vspace{-1em}

\textbf{Optimizers for LLM pretraining.}
The families we study span diagonal-adaptive (\AdamW; \AdEMAMix;~\citealp{pagliardini2024ademamix}), curvature-aware (\Sophia;~\citealp{liu2023sophia}), sign-based (\Lion, \Signum), and matrix-geometry methods (\Muon;~\citealp{jordan2024muon}), with \SOAP~\citep{vyas2025soap}, \MARS~\citep{yuan2025mars}.
Systematic benchmarks evaluate these under standard pretraining only~\citep{semenov2025benchmarking,wen2025fantasticpretrainingoptimizers}, where rankings already shift with scale and batch size.
No prior work explores how optimizer algorithms affect preferences in \emph{data selection}, which we analyze both theoretically (\S\ref{sec:theory}) and empirically (\S\ref{sec:experiments}) in this work.
\vspace{-1em}
\section{The \textit{OptiSelect} Paradigm}
\label{sec:optiselect}
\vspace{-1em}
In this section, we first formalize the \textit{OptiSelect} training loop (\S \ref{sec:pretrain}), then detail the optimizer-induced preconditioning that defines the data attribution score for online selection (\S \ref{sec:scoring}). 
\vspace{-1em}
\subsection{Data Schemes in LLM Pretraining}
\label{sec:pretrain}
\vspace{-0.5em}
\myparagraph{Standard pretraining.}
Let $\cD$ denote the pretraining data corpus.
Standard pretraining minimizes the next-token prediction loss for language modeling, i.e. $\cL(\btheta) = \E_{z \sim \cD}[\ell(z; \btheta)]$, where $\ell(z; \btheta)$ is the cross-entropy loss on data instance~$z$.
At each step~$t$, a mini-batch $\cB_t \subset \cD$ of size~$B$ is drawn uniformly at random and used to compute a stochastic gradient $\hat{\bg}_t = \frac{1}{B} \sum_{z \in \cB_t} \nabla_{\btheta} \ell(z; \btheta_t)$.
The model parameters are updated as $\btheta_{t+1} = \btheta_t - \eta_t \cdot \text{Opt}(\hat{\bg}_t, \cS_t)$, where $\text{Opt}$ is the optimizer's step function corresponding to the current optimizer state $\cS_t$, e.g. momentum buffers and variance estimates. \looseness=-1

\myparagraph{\textit{OptiSelect}: Online in-batch data selection.}
While uniform sampling provides no control over data quality, \textit{OptiSelect} applies \textit{online data selection} at each optimization step.
Specifically, denoting the effective training batch size by $B$, a larger \emph{candidate batch} $\tilde{\cB}_t$ of size $|\tilde{\cB}_t| = \lambda B$ is drawn at every step $t$. Each datapoint is ranked by the optimizer-aware influence score $U(z, t)$; the top-$B$ subset $\cB_t \subset \tilde{\cB}_t$ is then selected for the model update.
The full procedure is presented in Algorithm~\ref{alg:opus}.
Under \textit{OptiSelect} the optimizer plays a two-fold role: it performs the optimization step on the loss landscape, and simultaneously shapes data attribution by defining $U(z, t)$. We detail the optimizer-transformed utility under various instantiations in the following section. \looseness=-1

\vspace{-0.5em}
\subsection{Optimizer-Aware Influence Scoring}
\label{sec:scoring}
\vspace{-0.5em}
\paragraph{Preliminaries and notation.}
\label{sec:notation}
We begin by establishing a unified notation for all following derivations.
At step~$t$, we denote $\btheta_t \in \R^{\dd}$ as the model parameters, the per-sample loss on document~$z$ as $\ell(z;\btheta_t)$ and its gradient as $\bg_{z,t} \;\triangleq\; \nabla_{\btheta}\, \ell(z;\btheta_t)$ and the stochastic mini-batch gradient $\hat{\bg}_t = \frac{1}{|\cB_t|}\sum_{z\in\cB_t}\bg_{z,t}$.
We denote element-wise multiplication, division and square-root operations by~$\odot$,~$\oslash$, and~$\sqrt{\cdot}$ respectively.
We define the frozen-state scoring operator \(\mathcal O_t\) as the map from raw gradients to directions used for per-sample attribution, conditional on the optimizer state \(\cS_t\). Given a candidate \(z\), its scoring direction is \(u_{z,t}=\mathcal O_t(g_{z,t})\). This operator need not coincide with the actual optimizer update map \(\text{Opt}(\cdot,\cS_t)\); the relation between them, including the approximations used for scoring, is specified in~\autoref{app:operator_derivations}.
\myparagraph{Optimizer-transformed Data Influence.}
During online data selection, we assess the influence of a candidate sample~$z$ at step~$t$ using its \textit{per-sample scoring direction} $\bu_{z,t}$ induced by the optimizer state $\cS_t$ and a proxy set from a predefined target distribution.
Let $\cV$ be a small held-out validation set representative of a target distribution. We define the validation loss
$F(\btheta) = \frac{1}{|\cV|}\sum_{z\in\cV}\ell(z;\btheta)$.
Selection proceeds with greedy influence scoring at every step: writing $B^{(t)}$ as the selected batch at time step $t$, the parameter vector predicted by the additive scoring model is $\btheta_t^{\mathrm{sel}} \triangleq \btheta_t - \eta_t\sum_{z'\in\cS^{(t)}}\cO_t(\bg_{z',t})$. We estimate the \emph{marginal utility} of adding sample~$z$ into $B^{(t)}$ as the additional reduction in~$F$:
$U(z,t) \;=\; F\!\bigl(\btheta_t^{\mathrm{sel}}\bigr) - F\!\bigl(\btheta_t^{\mathrm{sel}} - \eta_t\,\cO_t(\bg_{z,t})\bigr)$.
Expanding $F$ to second order around~$\btheta_t$, denoting $\bg_{\cV,t} = \nabla_{\btheta}F(\btheta_t)$ as the validation gradient proxy and $\bH_F = \nabla^2 F(\btheta_t)$ as the validation Hessian, the marginal utility is defined as
$U(z,t) = \eta_t\inner{\cO_t(\bg_{z,t})}{\bg_{\cV,t}} - \eta_t^2\,|B^{(t)}|\inner{\cO_t(\bg_{z,t})}{\bH_F\,\bG_t} + \cO(\eta_t^2\norm{\cO_t(\bg_{z,t})}^2)$.
Following OPUS~\citep{wang2026opus}, we approximate the intractable $\dd\times\dd$ Hessian by the identity matrix $\bH_F \approx \bI$ in the redundancy term
, giving
\begin{equation}
  U(z,t) \;\approx\; \eta_t\,\inner{\cO_t(\bg_{z,t})}{\bg_{\cV,t}}
  \;-\; \eta_t^2\,|B^{(t)}|\inner{\cO_t(\bg_{z,t})}{\bG_t},
\label{eq:opus_score}
\end{equation}
where
$\bG_t = \frac{1}{|B^{(t)}|}\sum_{z'\in\cS^{(t)}}\cO_t(\bg_{z',t})$
is the average scoring direction of the selected samples.
The first term measures \textit{alignment} between sample~$z$ and the target distribution, while the second penalizes \textit{redundancy} among the peer candidates in the same batch.
\vspace{-0.3em}
\myparagraph{Instantiation of \texorpdfstring{$\cO_t$}{Ot} across various optimizer families.}
\label{sec:opt_geometries}
Each optimizer algorithm induces a specific operator $\cO_t$ for online data attribution at step $t$.
We write $\bx$ for the \emph{generic argument} of $\cO_t$, i.e. the per-sample raw gradient vector $\bx = \bg_{z,t} \in \R^{\dd}$ or the mini-batch gradient $\bx = \hat{\bg}_t \in \R^{\dd}$, where $\bd_t = \cO_t(\hat{\bg}_t)$ denotes the batch direction represented by the scoring map. For \Muon\ the argument is matrix-valued, i.e. $\bG \in \R^{m\times n}$, acting on each 2-D weight block in the transformer model.
The remaining symbols are read from the optimizer state $\cS_t$, which is identical for all candidates within one scoring pass at step $t$.
We present the transformation operators for the benchmarked optimizers, with detailed derivations provided in~\autoref{app:operator_derivations}: 
\vspace{-0.2em}
\begin{align}
  \cO_t^{\mathrm{\AdamW}}(\bx) &= \cO_t^{\mathrm{\AdEMAMix}}(\bx) = \bc_t \odot \bx,
  &\cO_t^{\mathrm{\Sophia}}(\bx) &= \bx \oslash \max\{\gamma\,\bhv_t,\, \varepsilon\},\nonumber\\[2pt]
  \cO_t^{\mathrm{\Lion}}(\bx) &= \sign\bigl(\beta\,\bmv_{t-1} + (1-\beta)\,\bx\bigr),
  &
  \cO_t^{\mathrm{\Lion},\,\mathrm{lin}}(\bx) &= \sign(\bmv_{t-1}) \odot \bx,
  \nonumber\\[2pt]
  \cO_t^{\mathrm{\Muon,lin}}(\bG) &= \mathrm{D}\msign(\bA_t)[\bG],
  &
  \cO_t^{\mathrm{\Muon,rp}}(\bG) &= \bG\,(\bA_t^\top\bA_t + \varepsilon\bI)^{-1/2},
\label{eq:operator_zoo}
\vspace{-0.5em}
\end{align}
Specifically, for \AdamW / \AdEMAMix (Appendix~\ref{app:op_adamw}, \ref{app:op_ademamix}), $\bc_t = \mathbf{1}\oslash(\sqrt{\hat{\bv}_t}+\varepsilon)$ denotes the inverse root of the bias-corrected second moment $\hat{\bv}_t$. \AdEMAMix\ induces the \emph{same} operator; for \Sophia (Appendix~\ref{app:op_sophia}), $\bhv_t$ denotes the estimate of diagonal curvature with trust scale $\gamma$;
for \Lion (Appendix~\ref{app:op_lion}), $\bmv_{t-1}$ denotes the sign methods' momentum buffer, with decay $\beta$. With $\beta \gg 1-\beta$; for \Muon (Appendix~\ref{app:op_dmuon}), $\bA_t$ denotes the momentum matrix.
$\mathrm{D}\msign(\bA_t)[\,\cdot\,]$ is the faithful Fr\'{e}chet derivative in $\cO_t^{\mathrm{\Muon,lin}}$, while $\cO_t^{\mathrm{\Muon,rp}}$ is a cheaper right-preconditioned proxy. 
\vspace{-1em}
\section{What Makes an Optimizer Good at Selecting Data?}
\label{sec:theory}
\vspace{-1em}
Since each optimizer shapes the influence function through its own transformation $\cO_t$, a natural question arises: \textit{how does the choice of optimizer interact with data selection?} 
To study this question, we introduce \textbf{\textit{discriminability}} (\autoref{thm:selection_gain}), a principled criterion that determines an upper bound on the gain of \textit{OptiSelect} over standard training under a given optimizer. 
We then analyze several optimization geometries through the lens of this \textit{selection gain principle}: we find that sign collapse and radial collapse cap the achievable selection gain for the sign-based optimizer \Lion and the polar-tangential optimizer \DMuon, respectively, whereas diagonal-adaptive optimizers such as \AdamW and \Sophia enjoy strictly better upper bounds on the selection gain.
Our proof builds on a Gaussian model of the data candidates. We detail the setup as follows.

\begin{wrapfigure}{r}{0.66\textwidth}
\centering
\vspace{-1.8em}
\includegraphics[width=\linewidth]{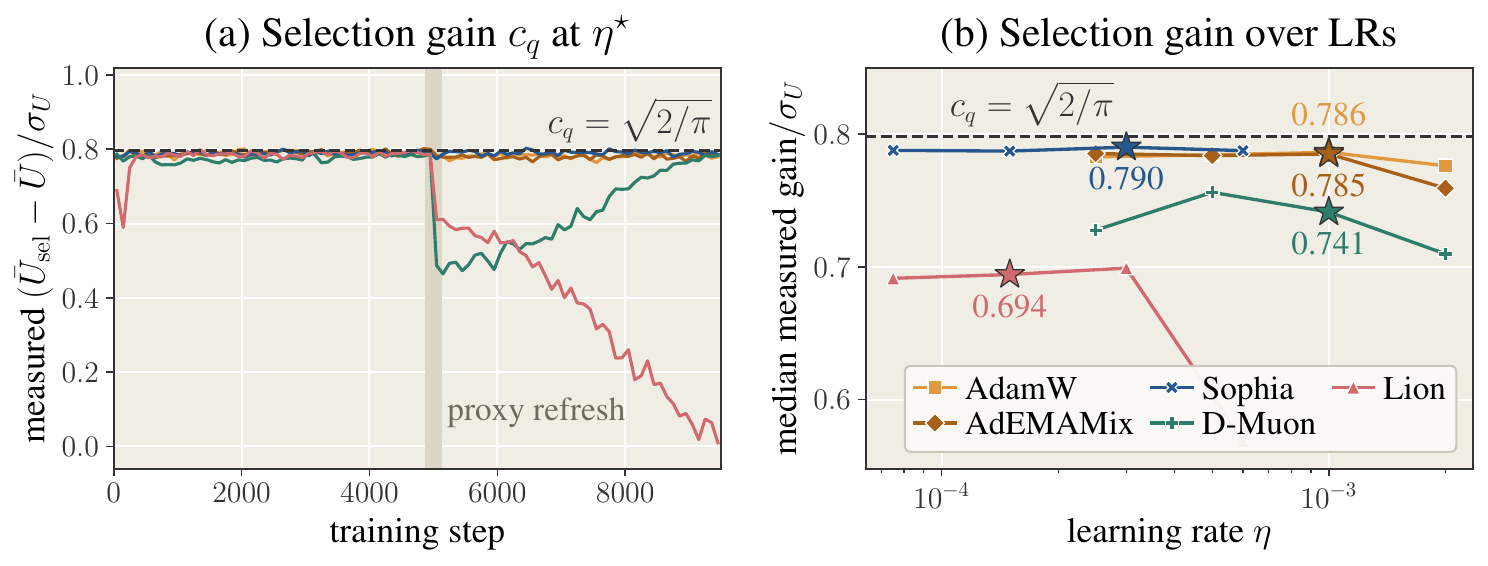}
\vspace{-2em}
\caption{\small \textbf{\textit{Selection gain}~(\autoref{thm:selection_gain}) measured during 124M LLM pretraining.}
(a)~\textbf{Running medians at tuned $\eta^\star$}: the measured gain mostly sits close to $c_q$ for \emph{every} optimizer; while \Lion decays after proxy refresh.
(b)~\textbf{Measured gain from learning rate sweep}: the selection gain is flat in various learning rate $\eta$.
}
\vspace{-1.8em}
\label{fig:selection_gain_llm}
\end{wrapfigure}
\myparagraph{Setup and Notations.}
At step~$t$, the candidate batch $\tilde{\cB}_t$ of size $\lambda B$ is drawn i.i.d.\ from the data distribution with utility scores follows a Gaussian distribution $\{U(z,t)\}_{z\in\tilde{\cB}_t} \sim \mathcal{N}(\mu_U(t), \sigma_U^2(t))$.
Consider that candidate gradients are also sampled from a Gaussian distribution $\bg_z \sim \mathcal{N}(\boldsymbol{\mu}, \boldsymbol{\Sigma})$, with diagonal covariance $\boldsymbol{\Sigma} = \diag(\sigma_1^2,\ldots,\sigma_\dd^2)$. 
Given the validation proxy $\bg_\cV$, we denote $\hat{\bg}_\cV=\bg_\cV-\eta_tG_t$ to absorb the redundancy term in~\eqref{eq:opus_score}. 
Denote $\boldsymbol{\pi}$ as the diagonal operator applied by $\cO_t$ transformation, the utility score on the sample $z$ can be written as
$U_z = \eta_t\inner{\boldsymbol{\pi}\odot\bg_z}{\hat{\bg}_\cV} = \eta_t\inner{\bg_z}{\boldsymbol{\pi}\odot\hat{\bg}_\cV}$, i.e. $\boldsymbol{\pi} = \mathbf{1}$ for \SGD, $\boldsymbol{\pi} = \bc_t$~\eqref{eq:operator_zoo} for \AdamW, and $\boldsymbol{\pi} = \sign(\bmv_{t-1})$ for linearized sign-based optimizers~(\autoref{prop:sign_neutral}). \looseness=-1

\begin{takeaway}[]{sqrtred}
\vspace{-0.5em}
\begin{theorem}[Selection-gain Principle]
\label{thm:selection_gain}
Consider the candidate utilities distribution $\mathcal{N}(\mu_U(t), \sigma_U^2(t))$ and standardized top-$q$ superquantile $c_q>0$\footnotemark
, where $q=1/\lambda$ is the retained fraction from OptiSelect.
The expected one-step reduction on the validation loss $F$, $R_t\triangleq\E[F(\btheta_t)-F(\btheta_{t+1})]$, satisfies
\begin{equation}
  R_t^{\mathrm{rand}} = \mu_U(t) + \bigO(\eta_t^2),
  \qquad
  R_t^{\mathrm{sel}} = \mu_U(t) + c_q\,\sigma_U(t) + \bigO(\eta_t^2),
\label{eq:per_step_reduction}
\end{equation}
which gives the \textbf{selection gain}, i.e. the extra loss reduction obtained from data selection as
\begin{equation}
  G_t \;\triangleq\; R_t^{\mathrm{sel}} - R_t^{\mathrm{rand}}
  \;=\; c_q\,\sigma_U(t) \;+\; \bigO(\eta_t^2).
\label{eq:selection_gain}
\end{equation}
\end{theorem}
\end{takeaway}
\footnotetext{After standardizing the candidate utilities to zero mean and unit variance, the \emph{standardized top-$q$ superquantile} ($c_q$) refers to the mean utility of the top $q = 1/\lambda$ candidates. For Gaussian utilities at $\lambda = 2$, $c_q$ is the mean of the upper half of a standard normal, $\sqrt{2/\pi} \approx 0.798$.}

The proof is given in~\autoref{app:selection_gain_proof}. Since $c_q$ depends only on the retained fraction $q=1/\lambda$, the absolute gain is governed by the score spread $\sigma_U(t)$. 
We validate~\autoref{thm:selection_gain} on $15$ \textit{OptiSelect} LLM pretraining runs and compute $(\bar{U}_{\mathrm{sel}} - \bar{U})/\sigma_U$ on the attribution scores, where the proxy gradient is refreshed at step $5k$. \autoref{fig:selection_gain_llm} shows that the normalized score gain $(\bar U_{\mathrm{sel}}-\bar U)/\sigma_U$ remains consistently close to $c_q$ and nearly invariant across various learning rates. 
Since $c_q$ depends only on the retained fraction $q=1/\lambda$, the absolute gain is governed by the score spread $\sigma_U(t)$. We further decompose the variance by coordinates and define the coordinate weights on \SGD as
$\displaystyle{w_j
  \;\triangleq\;
  \frac{
    \sigma_j^2[\hat{\bg}_{\cV}]_j^2
  }{
    \sum_k
    \sigma_k^2[\hat{\bg}_{\cV}]_k^2
  }}$. In the rest of this section, we adopt the \textit{variance ratio} $\displaystyle{
  \frac{
    \Var[U_z^{(\boldsymbol{\pi})}]
  }{
    \Var[U_z^{\mathrm{\SGD}}]
  }
  =
  \sum_{j=1}^{\dd}w_j\pi_j^2}$ to compare the \textit{discriminability} benefits of linear operators against \SGD.

\myparagraph{Sign Collapse.}
\label{sec:sign_collapse}
Sign-based optimizers can lose \textit{discriminability} since the momentum buffer easily dominates the sign of gradients, diminishing the candidate-aware differences. 
The selector therefore has little information for ranking them, which approaches a \emph{uniform} selection strategy. 
\autoref{thm:sign_collapse} formalizes this \textit{sign collapse} by comparing the exact score with the linear surrogate score $U_z^{\mathrm{\Lion,lin}}$.
\looseness=-1

\begin{takeaway}[]{sqrtred}
\begin{theorem}[Sign collapse]
\label{thm:sign_collapse}
Condition on the frozen optimizer state and scoring reference
$\hat{\bg}_{\cV}$, and let
$\bg_z\sim\mathcal{N}(\boldsymbol{\mu},\boldsymbol{\Sigma})$ with
$\boldsymbol{\Sigma}=\diag(\sigma_1^2,\ldots,\sigma_\dd^2)$.

\textbf{(i)} (\emph{Spread neutrality of the linear surrogate score.})\\
For a sign pattern
$\bs_t\triangleq\sign(\bmv_{t-1})\in\{-1,+1\}^{\dd}$, 
\begin{equation}
\vspace{-0.3em}
  U_z^{\mathrm{\Lion,lin}}
  \;\triangleq\;
  \eta_t\inner{\bg_z}{\bs_t\odot\hat{\bg}_{\cV}}
  \quad\Longrightarrow\quad
  \Var\bigl[U_z^{\mathrm{\Lion,lin}}\bigr]
  =
  \eta_t^2
  \sum_{j=1}^{\dd}
  \sigma_j^2[\hat{\bg}_{\cV}]_j^2
  =
  \Var\bigl[U_z^{\mathrm{\SGD}}\bigr].
\label{eq:var_lion}
\end{equation}

\textbf{(ii)} (\emph{Exponential collapse of the exact score.})
Suppose $\exists \kappa\geq1$ satisfying Assumption \ref{ass:momentum_dominance}, 
the exact \Lion\ score
$U_z^{\mathrm{\Lion}}
  \;\triangleq\;
  \eta_t
  \inner{
    \sign\bigl(
      \beta_1\bmv_{t-1}
      +(1-\beta_1)\bg_z
    \bigr)
  }{
    \hat{\bg}_{\cV}
  }$
satisfies
\begin{equation}
  \Var\bigl[U_z^{\mathrm{\Lion}}\bigr]
  \;\leq\;
  8e^{-\kappa^2/2}
  \eta_t^2
  \norm{\hat{\bg}_{\cV}}_2^2.
\label{eq:var_lion_true}
\end{equation}
Consequently, when
$\norm{\boldsymbol{\Sigma}^{1/2}\hat{\bg}_{\cV}}_2>0$ and $\kappa^2
  \;\geq\;
  2\log\!\left(
    \frac{
      8\norm{\hat{\bg}_{\cV}}_2^2
    }{
      \norm{
        \boldsymbol{\Sigma}^{1/2}
        \hat{\bg}_{\cV}
      }_2^2
    }
  \right)$, we have 
\begin{equation}
  \Var\bigl[U_z^{\mathrm{\Lion}}\bigr]
  \;\leq\;
  \Var\bigl[U_z^{\mathrm{\Lion,lin}}\bigr] \;=\; \Var\bigl[U_z^{\mathrm{\SGD}}\bigr].
\end{equation}
\end{theorem}
\end{takeaway}
\autoref{thm:sign_collapse}~\textup{(i)} shows that a frozen sign pattern preserves the \SGD\ score variance but cannot adaptively amplify discriminability;~\autoref{thm:sign_collapse}~\textup{(ii)} shows that the exact score can degrade this baseline spread under momentum dominance. The full proof are deferred to Appendix~\ref{app:proof_sign}. 

\myparagraph{Diagonal-adaptive advantage.} \autoref{thm:diag_advantage} further compares the optimizers along their linear diagonal operators. We define the coordinate weights 
$\displaystyle{w_j
  \;\triangleq\;
  \frac{
    \sigma_j^2[\hat{\bg}_{\cV}]_j^2
  }{
    \sum_k
    \sigma_k^2[\hat{\bg}_{\cV}]_k^2
  }}$ with variance ratio $\displaystyle{
  \frac{
    \Var[U_z^{(\boldsymbol{\pi})}]
  }{
    \Var[U_z^{\mathrm{\SGD}}]
  }
  =
  \sum_{j=1}^{\dd}w_j\pi_j^2}$ to compare the benefits of linear operators against \SGD.

\begin{takeaway}[]{sqrtred}
\begin{theorem}[Diagonal-adaptive advantage, and the discriminability ranking]
\label{thm:diag_advantage}
Under~\autoref{ass:variance_normalization} and its curvature analogue for \Sophia with its hessian estimates,
{%
\setlength{\abovedisplayskip}{2pt plus 1pt minus 1pt}%
\setlength{\belowdisplayskip}{2pt plus 1pt minus 1pt}%
\setlength{\abovedisplayshortskip}{0pt plus 1pt}%
\setlength{\belowdisplayshortskip}{1pt plus 1pt minus 1pt}%
\begin{equation}
  \frac{\Var[U_z^{\mathrm{\AdamW}}]}{\Var[U_z^{\mathrm{\SGD}}]}
  \;=\; \sum_{j=1}^{\dd} w_j\,[\bc_t]_j^2 \;>\; 1 - \bigO(\varepsilon'),
  \qquad
  \frac{\Var[U_z^{\mathrm{\Sophia}}]}{\Var[U_z^{\mathrm{\SGD}}]}
  \;>\; 1 - \bigO(\varepsilon''),
\label{eq:diag_amplification}
\end{equation}
}%
with $\varepsilon', \varepsilon''$ the respective numerical floors.
Combined with~\autoref{thm:sign_collapse}, in the regime $\kappa^2 \geq 2\log\bigl(8\norm{\bg_\cV}_2^2/\norm{\boldsymbol{\Sigma}^{1/2}\bg_\cV}_2^2\bigr)$ the spreads of the operator classes are ranked:
{%
\setlength{\abovedisplayskip}{2pt plus 1pt minus 1pt}%
\setlength{\belowdisplayskip}{2pt plus 1pt minus 1pt}%
\setlength{\abovedisplayshortskip}{0pt plus 1pt}%
\setlength{\belowdisplayshortskip}{1pt plus 1pt minus 1pt}%
\begin{equation*}
  \underbrace{\Var[U_z^{\mathrm{\Lion}}]}_{\text{exact sign}}
  \;\leq\;
  \underbrace{\Var[U_z^{\mathrm{\Lion,lin}}]}_{\text{linearized sign}}
  \;=\;
  \Var[U_z^{\mathrm{\SGD}}]
  \;<\;
  \Var[U_z^{\mathrm{\AdamW}}],\;
  \Var[U_z^{\mathrm{\Sophia}}].
\end{equation*}
}%
\end{theorem} 
\end{takeaway}
The final inequality holding separately for each of the two diagonal-adaptive scores from \AdamW and \Sophia, which the \textit{discriminability} does not order against each other.\looseness=-1

\myparagraph{Radial collapse in the ideal polar geometry of \Muon.}
\label{sec:muon_spectral}
The ideal \Muon\ map removes the stretching of the momentum matrix
while retaining its orientation. To see this, write the polar
decomposition as $\bA_t=\bQ\bH$: the ideal update is
$\bQ=\msign(\bA_t)$, whereas $\bH\succ0$ describes the stretching
that the map discards. A candidate can therefore change the stretching
without turning the update. We call such changes \emph{radial};
changing singular values while keeping their directions fixed is a
simple example. \looseness=-1

For a frozen full-column-rank momentum $\bA_t$ and scoring reference
$\bR=\hat{\bG}_\cV$, decompose each candidate gradient into its radial
component and the orthogonal remainder,
$\bG_z=\bG_z^{\mathrm{rad}}+\bG_z^{\mathrm{tan}}$.
The latter is its \emph{tangential} component, which can change the
update's orientation. The precise decomposition is given in
Appendix~\ref{app:proof_muon_spectral}.
Let $\mathcal L_t=\mathrm D\msign(\bA_t)$ denote the local linear
response of the ideal map.

\begin{takeaway}[]{sqrtred}
\begin{theorem}[Radial selectivity of the ideal polar score]
\label{thm:muon_spread}
At the frozen state, suppose $\bG_z$ has finite second moment and
pairwise uncorrelated entries in the singular-vector coordinates
specified in Appendix~\ref{app:proof_muon_spectral}.
For $\eta_t>0$ and candidate weight $\alpha\ge0$, define
$
U_z^{\Muon}
=\eta_t\alpha\inner{\mathcal L_t[\bG_z]}{\bR}_F$,
$
U_z^{\mathrm{tan}}
=\eta_t\inner{\bG_z^{\mathrm{tan}}}{\bR}_F$. Then $U_z^{\Muon}$ depends on the candidate only through
$\bG_z^{\mathrm{tan}}$, and
\begin{equation}
\Var(U_z^{\Muon})
\le
\frac{\alpha^2}{\sigma_{\min}(\bA_t)^2}
\Var(U_z^{\mathrm{tan}}),
\label{eq:muon_ranking_slot}
\end{equation}
where $\sigma_{\min}(\bA_t)>0$ is the smallest singular value
of the frozen momentum.
\end{theorem}
\end{takeaway}

Thus, two candidates that differ \emph{only radially} receive the same
first-order score. The collapse is selective: radial differences
are ignored, while tangential differences are reweighted according
to the momentum spectrum. 
The \emph{practical} right-preconditioned proxy used in our experiments~\eqref{eq:operator_zoo}, behaves differently: it rescales the gradient
without removing its radial component, and can therefore distinguish
candidates that the ideal first-order score treats as identical. \autoref{thm:muon_spread} implies neither an unconditional variance ordering against \SGD nor a radial-collapse guarantee for the experimental proxy.
The exact variance formulas, proof, and proxy discrepancy are given
in Appendices~\ref{app:proof_muon_spectral} and~\ref{app:muon_approx}. \looseness=-1

\vspace{-1em}
\section{Experiments}
\label{sec:experiments}
\vspace{-0.5em}
We conduct extensive experiments on LLM pretraining and give the first optimizer benchmarking along the \textit{OptiSelect} paradigm (\S\ref{sec:optiselect-bench}). Notably, we also show that the optimizer-aligned geometry may not be optimal for data selection (\S\ref{sec:geometry-ablation}).
We further assess the impact of data format on \textit{OptiSelect} with rephrasing, showing that the benefits remains upon modern data synthesis pipeline (\S\ref{sec:finephrase}). 

\myparagraph{Models and training.}
\label{sec:opt_configs}
We train 124M and 720M decoder-only transformers models following Llama architecture~\citep{touvron2023llama} adopted in ~\citet{semenov2025benchmarking}. We benchmark seven core optimizers, \AdamW, \AdEMAMix, \Sophia, \Lion and \DMuon under both standard training and \textit{OptiSelect} paradigm, incurring a direct comparison of the effectiveness of online data selection. We train
\begin{wraptable}{r}{0.76\textwidth}
\centering
\vspace{-1em}
\caption{\small\textbf{Data selection benefits across optimizers.} Optimizers are measured according to their downstream losses and loss gap $\Delta = \cL_\mathrm{standard} - \cL_\mathrm{OptiSelect}$, followed by their ranks. Larger $\Delta$ indicates larger data selection gain.}
\label{tab:loss_rank}
\centering
\small
\vspace{-1em}
\adjustbox{max width=0.80\textwidth}{
\begin{tabular}{l rrr c rrr}
\toprule
& \multicolumn{3}{c}{\textbf{124M}} & & \multicolumn{3}{c}{\textbf{720M}} \\
\cmidrule(lr){2-4}\cmidrule(lr){6-8}
\textbf{Optimizer} & standard & \textit{OptiSelect} & $\Delta$ & & standard & \textit{OptiSelect} & $\Delta$ \\
\midrule
\AdEMAMix & 3.410 \textsubscript{(2)} & 3.274 \textsubscript{(2)} & $+\mathbf{0.135}$ & & 3.004 \textsubscript{(3)} & 2.908 \textsubscript{(2)} & +$\mathbf{0.096}$ \\
\AdamW & 3.361 \textsubscript{(1)} & 3.241 \textsubscript{(1)} & $+\mathbf{0.119}$ & & 2.995 \textsubscript{(2)} & 2.891 \textsubscript{(1)} & +$\mathbf{0.104}$ \\
\Sophia & 3.486 \textsubscript{(5)} & 3.458 \textsubscript{(4)} & +$\mathbf{0.028}$ & & 3.049 \textsubscript{(5)} & 3.026 \textsubscript{(5)} & +$\mathbf{0.023}$ \\
\Lion & 3.417 \textsubscript{(3)} & 3.399 \textsubscript{(3)} & +$\mathbf{0.018}$ & & 3.018 \textsubscript{(4)} & 3.022 \textsubscript{(4)} & $\textcolor{red}{-\mathbf{0.004}}$ \\
\DMuon & 3.455 \textsubscript{(4)} & 3.471 \textsubscript{(5)} & $\textcolor{red}{-\mathbf{0.015}}$ & & 2.993 \textsubscript{(1)} & 2.944 \textsubscript{(3)} & +$\mathbf{0.049}$ \\
\bottomrule
\end{tabular}
}
\vspace{-1.1em}
\end{wraptable}
124M and 720M models for 4.2B and 14B tokens respectively, matching their Chichilla~\citep{hoffmann2022trainingcomputeoptimallargelanguage} optimal token amount. We apply the optimal learning rate tuned from a grid search on both 124M and 720M models, detailed in Appendix~\ref{app:lr_sweep}.
Complete configurations are provided in Appendix~\ref{app:exp_configs}.

\myparagraph{Data and selection proxy.}
All the models are trained on FineWeb-HQ\footnote{\url{https://huggingface.co/datasets/epfml/FineWeb-HQ}}, a high-quality, model-filtered pretraining dataset derived as a subset of FineWeb. 
The selection proxy $\cV$ is a fixed set of $4{,}096$ densely packed sequences drawn from a \emph{downstream} task mixture---ARC-Easy, ARC-Challenge~\citep{clark2018arc}, HellaSwag~\citep{zellers2019hellaswag}, PIQA and SciQ, which serves as a targeted guidance of data selection. 
The proxy gradient is re-computed every $1{,}000$ steps. The ablation of proxy-refresh interval is presented in Appendix~\ref{app:refresh_sweep}.

We report two axes of evaluation: (1) \emph{\textbf{Downstream loss}} computes the loss of the correct continuation given the prompt on held-out validation sets, averaged across five proxy tasks; 
(2) \emph{\textbf{Zero-shot accuracy}} as the exact match rate measured by \texttt{lm-evaluation-harness}~\citep{eval-harness} toolkit. We report the accuracy scores over nine tasks, including the 5 \textit{in-domain} proxy tasks plus 4 \textit{out-of-domain} ones WinoGrande, WSC, CommonsenseQA and OpenBookQA, which the proxy never sees.
\vspace{-0.5em}
\subsection{The \textit{OptiSelect} Benchmark: Selection Gain Across Optimizers and Scale}
\label{sec:optiselect-bench}
\begin{figure}[ht!]
\centering
\begin{subfigure}{\linewidth}
  \centering
  \vspace{-1em}
  \includegraphics[width=\linewidth]{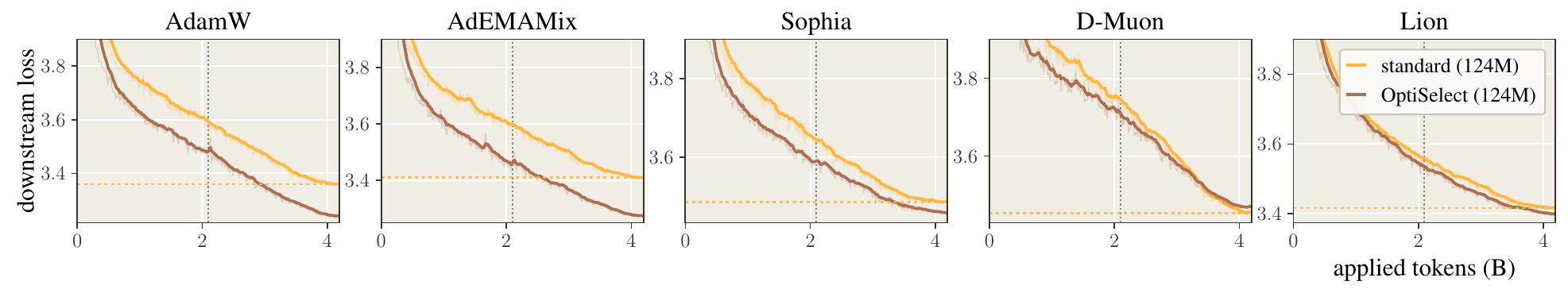}
  \vspace{-2em}
\end{subfigure}\\[8pt]
\begin{subfigure}{\linewidth}
  \centering
  \vspace{-1em}
  \includegraphics[width=\linewidth]{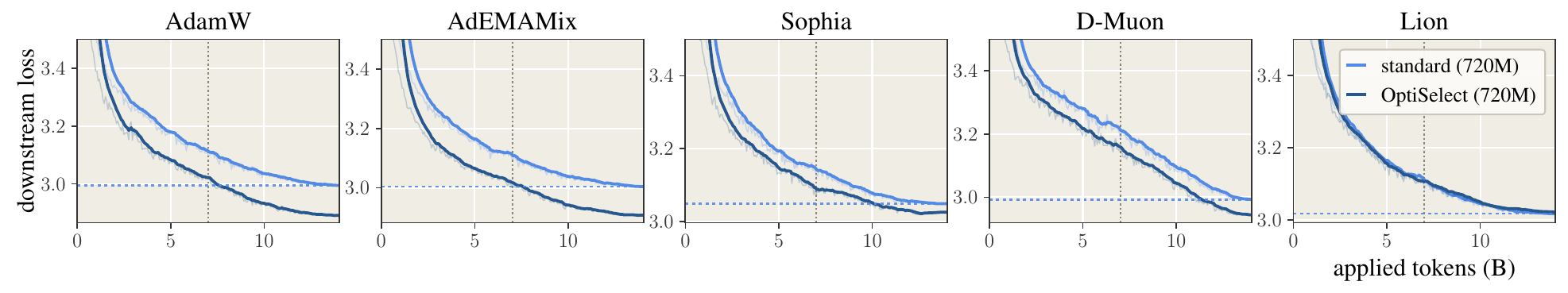}
  \vspace{-1.5em}
\end{subfigure}
\vspace{-2em}
\caption{\small \textbf{Downstream loss on 124M and 720M models across optimizers.} Models are trained on $4.2$B/$14.03$B tokens under standard and \textit{OptiSelect} paradigm, respectively. The intersection of two dotted curves denotes the processed-matched baseline with equal processed tokens under $\lambda=2$.}\label{fig:optiselect_loss}
\vspace{-1.5em}
\end{figure}

\begin{wrapfigure}{r}{0.70\textwidth} 
\vspace{-1em} 
\includegraphics[width=\linewidth]{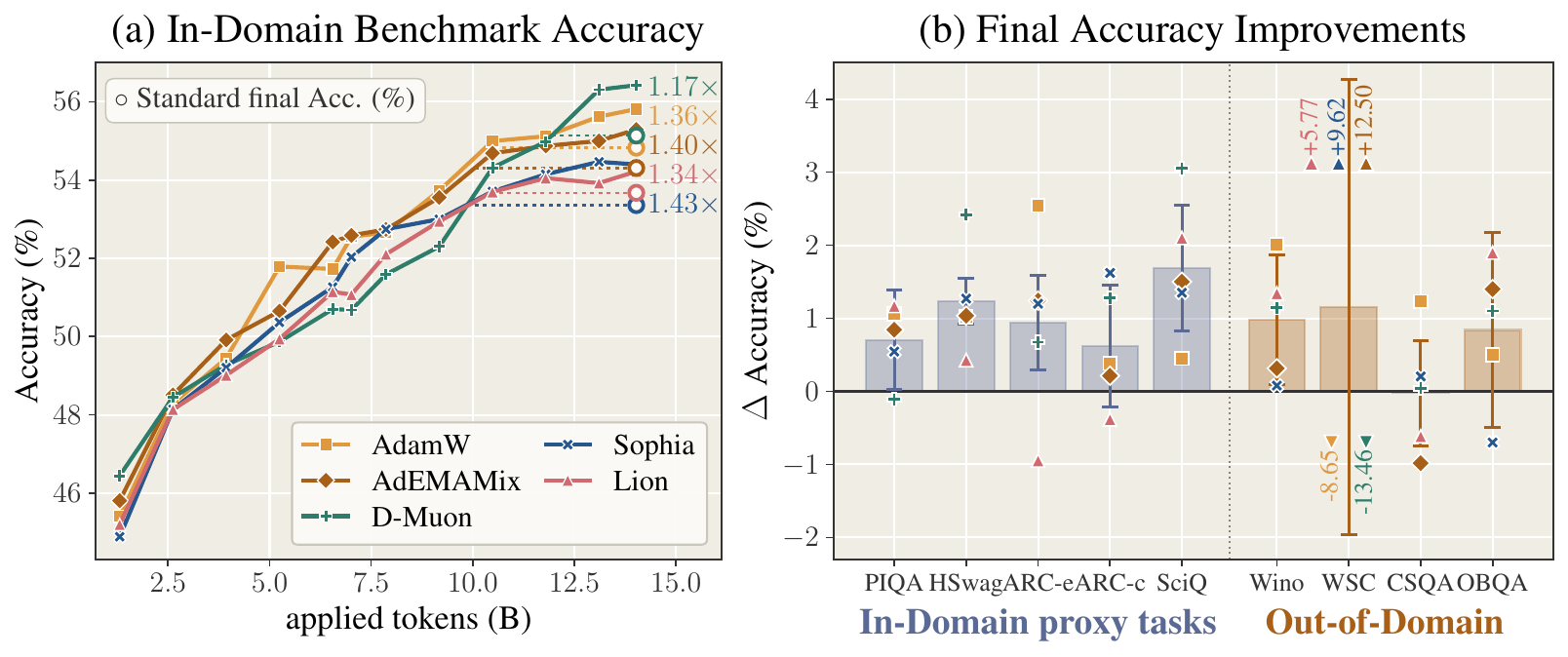} 
\vspace{-2.2em} 
\caption{\small \textbf{Downstream Accuracy at 720M scale.} \textit{OptiSelect} outperforms standard training across all optimizers on in-domain and OoD benchmarks.} 
\vspace{-1em} 
\label{fig:720m_acc} 
\end{wrapfigure}

\vspace{-0.5em}
\myparagraph{Optimizer ranking under \textit{OptiSelect}.}
\autoref{fig:optiselect_loss} and~\autoref{tab:loss_rank} report the downstream losses across optimizers and model scales. The optimizer ranking presented in ~\autoref{tab:loss_rank} demonstrates a scale-dependent impact from data selection: at 124M, the optimizer ranks on standard training are mostly preserved on \textit{OptiSelect}; while the ranking differs significantly at 720M.
In terms of the downstream accuracy (\autoref{fig:720m_acc}), despite the limited gain from online data selection, \DMuon consistently achieves the best averaged downstream accuracy, benefits from its extraordinary efficacy in optimization.

\myparagraph{Empirical--theory consistency.} According to~\autoref{fig:optiselect_loss} and~\autoref{tab:loss_rank}, the empirical selection
gains remarkably \textit{align with our theoretical analysis}: \textit{OptiSelect} consistently improves the downstream loss with \AdamW, \AdEMAMix and \Sophia\ at both 124M and 720M, whereas obtain no benefits with \Lion. \DMuon demonstrates a scale-dependent behavior: at 124M, the selection gain shrinks by the end of training; while at 720M, the benefit remains significant and persists throughout the training.

\myparagraph{\textit{OptiSelect} gains in sample efficiency, not compute.}
Figure~\ref{fig:optiselect_loss} and Figure~\ref{fig:720m_acc} expose a critical reflection:
\textit{OptiSelect}'s gains are measured upon optimizer \emph{steps} at equal \emph{applied} tokens while not the gains 
\begin{wrapfigure}{r}{0.68\textwidth}
\centering
\vspace{-1.5em}
\includegraphics[width=\linewidth]{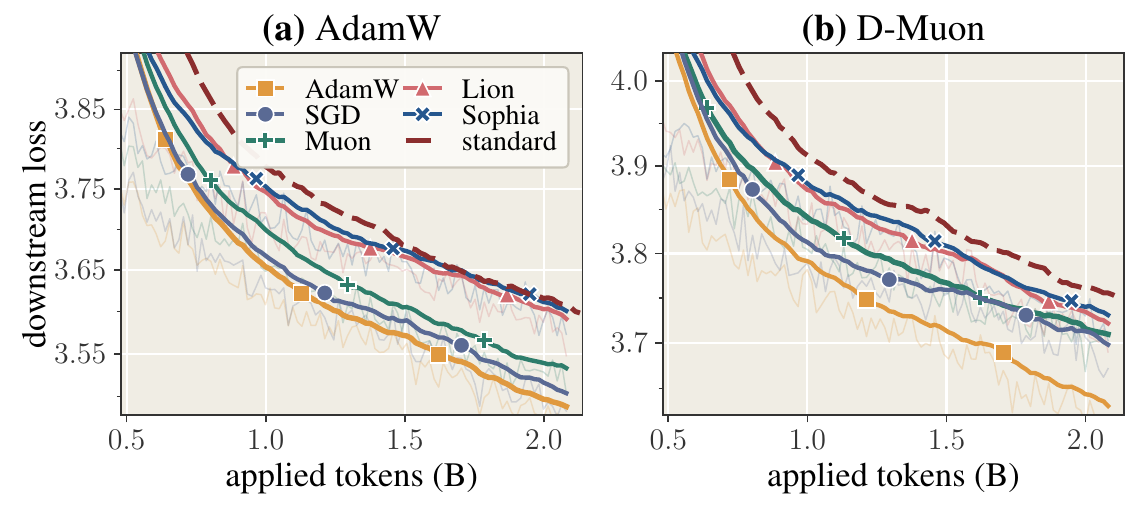}
\vspace{-2.2em}
\caption{\small \textbf{Scoring geometry ablations under \AdamW and \DMuon\ optimizers}.
The matched geometry works best under \AdamW, while both \AdamW and \SGD alternatives beat the matched scoring geometry on \DMuon.
}
\vspace{-1.5em}
\label{fig:geometry}
\end{wrapfigure}
per token \emph{read}.
At $\lambda = 2$, every selection step reads twice the tokens it applies, each with a full forward--backward pass, therefore a fair comparison with equal read-tokens 
requires \textit{OptiSelect} achieveing a $2\times$ acceleration in optimization steps.
However, comparing with standard training with matched read tokens, \textit{OptiSelect} falls behind in both downstream loss and accuracy across all five optimizers. Given the deficit shrinks at larger scale, \textit{OptiSelect} is still promising on scaled-up experiments.
\vspace{-1em}
\subsection{Which Scoring Geometry Should an Optimizer Use?}
\label{sec:geometry-ablation}
\vspace{-0.5em}
The inconsistency of optimizer ranking between standard training and \textit{OptiSelect} indicates a dichotomy of the preconditioning geometry for optimization and data selection: The geometry from the best-performed optimizer, \DMuon, is \emph{not the optimal for data selection}.
Inspired from the observation, we decouple the choice of optimizer and the geometry adopted for attribution scoring in~\eqref{eq:opus_score}. 
We run the scoring geometry ablation under two base optimizers \AdamW\ and \DMuon\ (\autoref{fig:geometry}). Each optimizer is combined with various $\cO_t$, i.e. the geometry, from each of the five optimizers. The \SGD\ scoring geometry recovers the GREATS algorithm from \citet{wang2024greats}.
\myparagraph{The optimizer-matched scoring geometry is not necessarily the best.} 
Under \AdamW, the best scoring geometry is also from \AdamW which matches the optimization dynamics. 
However, with \DMuon as base optimizer, the pairing geometry does not work well in data selection, while the geometry of \AdamW outperforms the others by a significant margin.
Notably, \textbf{the second-momentum}
\textbf{\textit{diagonal-adaptive operator from}} \AdamW \textbf{\textit{serves as the best scoring geometry}} in both \AdamW and \DMuon sweeps;
and \textit{scoring without preconditioner} (\SGD) beats most of the scoring-geometry. 
Additionally, \AdamW with the curvature-diagonal and sign-based geometries from \Sophia and \Lion can hardly outperform the standard training baseline with no selection. 
It indicates that a wrong geometry can hurt the performance of online data selection.
\begin{wrapfigure}{r}{0.44\textwidth}
\centering
\vspace{-0.5em}
\includegraphics[width=0.99\linewidth]{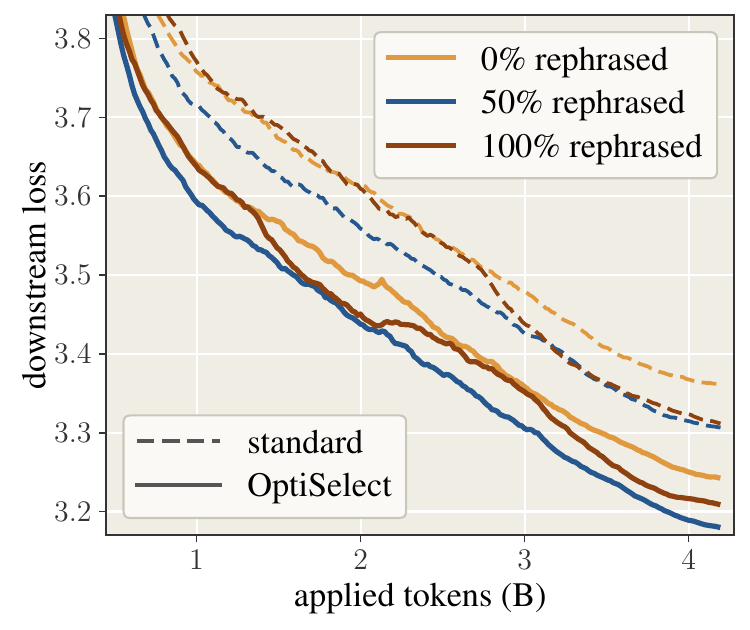}
\vspace{-1em}
\caption{\textbf{Downstream loss curves across rephrased mixtures},
where the benefits from selection are consistent.
}
\vspace{-2.5em}
\label{fig:finephrase_loss}
\end{wrapfigure}
\vspace{-2em}
\subsection{Does Selection Help with Rephrasing?}
\label{sec:finephrase}
\vspace{-0.5em}
Selection and format rephrasing are two techniques on data quality, while to what extend they are collaborative or conflict is unexplored.
FinePhrase~\citep{huggingface2025finephrase} rewrites documents from FineWeb-edu~\citep{lozhkov2024finewebedu} into various \textit{formats}, while leaving its content largely intact. 
We train \AdamW\ at 124M on mixtures containing $0\%$, $50\%$ and $100\%$ rephrased documents with consistent setup as in \S\ref{sec:opt_configs}.

\begin{wrapfigure}{r}{0.74\textwidth}
\centering
\vspace{-1.5em}
\includegraphics[width=\linewidth]{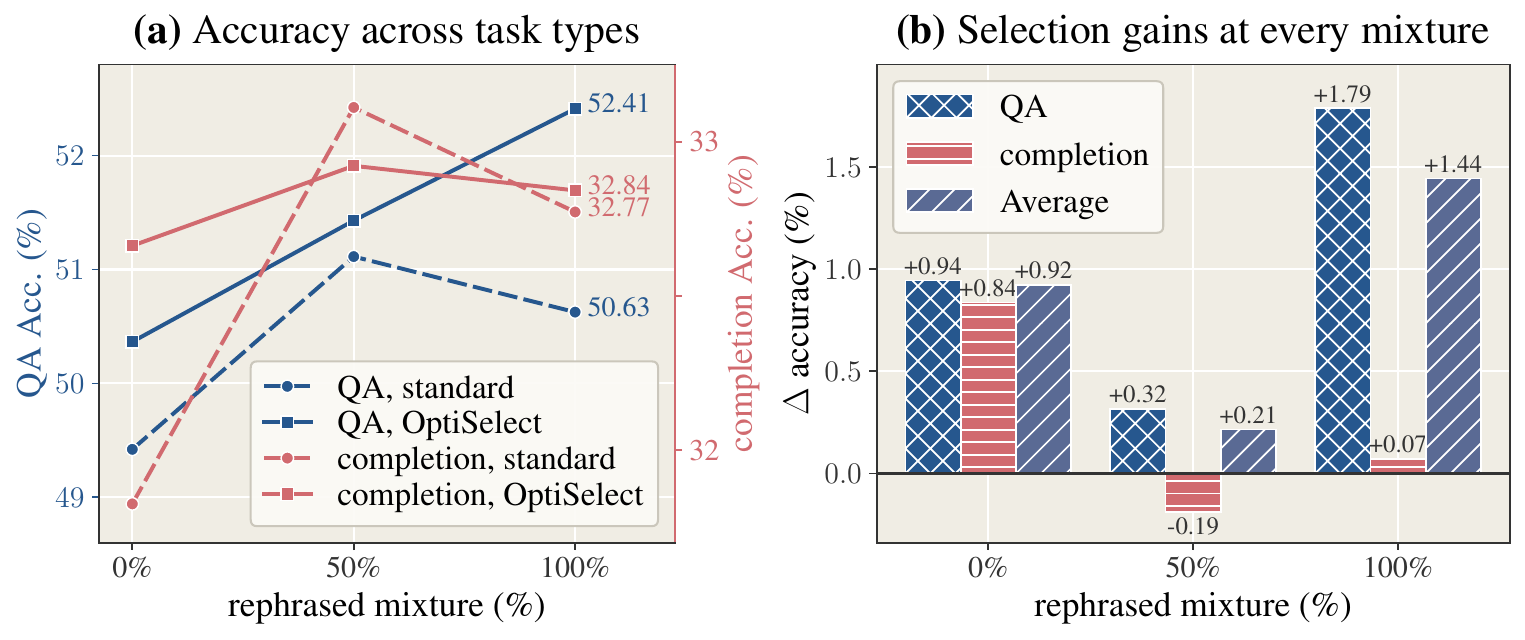}
\vspace{-2.2em}
\caption{\small\textbf{Task accuracy on format rephrasing $\times$ selection at 124M.}
\textbf{(a)} Accuracy against the rephrased mixture rates.
\textbf{(b)} The selection gain at each mixture.}
\label{fig:finephrase}
\vspace{-1.5em}
\end{wrapfigure}

\myparagraph{The selection benefit remains under data rephrasing.}
According to~\autoref{fig:finephrase_loss}, \textit{OptiSelect} consistently improves the downstream loss by a large margin comparing to the standard training. \textit{OptiSelect} also outperforms standard training across rephrasing mixtures on the average downstream accuracy~(\autoref{fig:finephrase}, \autoref{tab:finephrase_acc}). 
It indicates that rephrasing and selection are not substitutes or conflict against each other.

\myparagraph{\textit{Format mismatching} is a confound for downstream accuracy, but it depends on evaluation harness pattern.}
While the selection gets similar amount of gains on downstream loss along various rephrasing mixtures, the benefits on downstream accuracies move significantly along the \textit{formatting} 
axis. Specifically, we separate the downstream tasks into QA- and completion-formatted categories according to the prompting format adopted in the evaluation harness (Appendix \ref{app:rephrase-format}).
Under standard pretraining, the mixture with $50\%$ rephrased documents achieves the best downstream loss and accuracy scores in both QA- and completion-formatted tasks, aligned with the common practice in \citep{maini2024wrap} which samples real and synthetic data in a 1:1 ratio. 
In comparison, \textit{OptiSelect} greatly augments the gain on QA-formatted accuracies, reaches $52.51\%$ pp. On completion-formatted tasks, \textit{OptiSelect} obtains a great improvement ($+2\%$ pp.) on the original FineWeb corpus, while increasing rephrasing proportions does not bring more benefits. 
\vspace{-1em}
\subsection{More Candidates, Better Performance?} 
\vspace{-0.5em}
The candidate multiplier $\lambda$ is a critical hyperparameter for data selection, which determines the proportion of selected candidates. 
Instead of introducing as a magical hyperparameter settled in prior works \citep{wang2024greats, wang2026opus}, we quantify the effect of $\lambda$ under the \textit{selection gain principle} (\autoref{thm:selection_gain}). While the per-step gain $c_q\,\sigma_U$ with $q=1/\lambda$ is monotonically increasing by $\lambda$, the overhead on the streamed and scored tokens grow correspondingly. \autoref{lem:lambda_star} prices this trade-off and derives an optimal candidate multiplier $\lambda^{\star}$ in closed form.
\begin{takeaway}[]{sqrtred}
\begin{lemma}[Optimal candidate multiplier]
\label{lem:lambda_star}
Under the conditions of Theorem~\ref{thm:selection_gain} with Gaussian candidate utilities, the selection gain per \emph{streamed} token is
\begin{equation}
  \frac{c_q(\lambda)}{\lambda} \;=\; \phi\bigl(\Phi^{-1}(1-1/\lambda)\bigr),
\label{eq:gain_per_token}
\end{equation}
which attains its unique maximum at $\lambda^\star=2$, of value $\phi(0)=1/\sqrt{2\pi}$.
\end{lemma}
\end{takeaway}
\autoref{lem:lambda_star} proves that the selection gain grows as $\sqrt{2\ln\lambda}$, monotonically but not linearly. We provide the full proof in \ref{app:lambda_star_proof}. Therefore, the increment of selection gain cannot compensate for its extra costs when it past the peak value at $\lambda^{\star}=2$. \textit{A token-limited budget should stream exactly}
\begin{wrapfigure}{r}{0.65\textwidth}
\centering
\vspace{-1.2em}
\includegraphics[width=\linewidth]{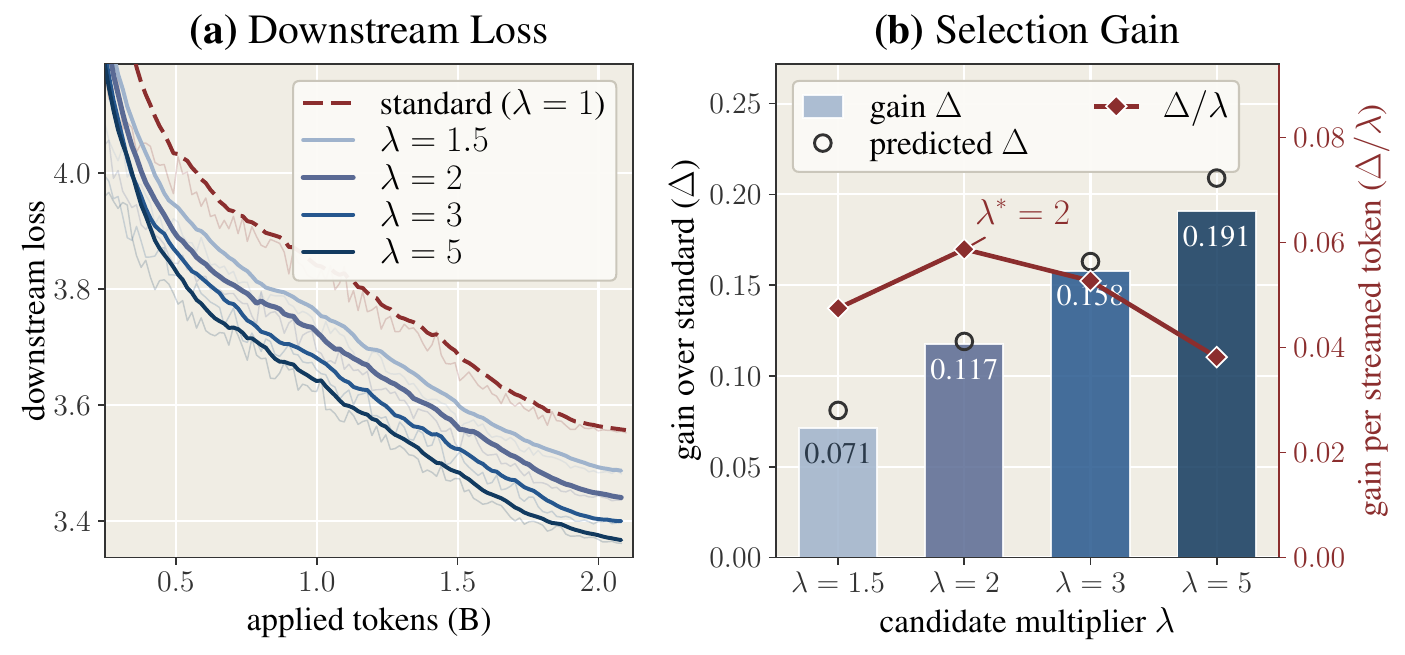}
\vspace{-2em}
\caption{\small \textbf{The candidate-multiplier sweep at 124M}.
\textbf{(a)}~\textbf{Downstream loss curves.}
\textbf{(b)}~\textbf{Selection gain over the standard}. Open circles mark the theoretical values of $c_q\sigma_U$ from~\autoref{thm:selection_gain}; the dark-red curve draws the gain per streamed token $\Delta/\lambda$ on the right axis, which peaks at $\lambda^{\star}=2$.}
\vspace{-1.5em}
\label{fig:lambda_ablation}
\end{wrapfigure}
\textit{ twice what it trains on}, which is the default used everywhere else in this paper. In~\autoref{fig:lambda_ablation}, we empirically measure the raw selection gain $\Delta$ as the downstream loss reduction above the standard training and the gain per streamed token as $\Delta / \lambda$, which validates the theory. Across various values of $\lambda$, the empirical raw selection gain $\Delta$ agrees with the theoretically computed values~(\autoref{fig:lambda_ablation}b), which keeps rising according to $\lambda$. In comparison, the measured $\Delta / \lambda$ demonstrates a concave curve, peaks at $\lambda^{\star} = 2$ while decreasing after the optimum. 

\vspace{-1em}
\section{Discussion and Limitations}
\vspace{-1em}
\myparagraph{The optimizer determines the landscapes of curriculum.
}
Under \textit{OptiSelect} the optimizer is not merely an optimization algorithm, it also serves as \emph{the designer of the curriculum}. The variance normalization of \AdamW builds a curriculum biased toward candidates whose gradients are large along rare dimensions, i.e. coordinates with small $[\hat{\bv}_t]_j$ are amplified;
\Muon's matrix-sign geometry builds a polar-tangent curriculum, which ignores candidates that only change singular values in the current polar basis and highlights those rotate singular subspaces. How to design an optimization algorithm that adapts to the data curriculum is a critical problem for future study. 

\myparagraph{What selection buys, and how it must be reported.}
Under a downstream proxy, \textit{OptiSelect} consistently improves the metrics it targets as proxy while may not necessarily benefit for held-out tasks. According to Figure \ref{fig:720m_acc} (b), \textit{OptiSelect} obtains benefits across 3 out of 4 Out-of-Domain benchmarks, while \textit{OptiSelect} yields little gains on Commonsense QA task.
Besides, selection is ahead at equal \emph{applied} tokens and behind at equal tokens \emph{read}, at 124M and at 720M alike: it converts extra forward--backward compute into fewer optimizer steps for the same quality, not into fewer tokens consumed. 
We report both matchings throughout and recommend that future evaluations of in-batch selection do the same; either number quoted alone misstates the method in one direction.

\myparagraph{Limitations and future work.}
Our benchmark covers only 124M and 720M models, which can only predict the behaviour at multi-billion scale by extrapolation. 
Besides, we only conducted each run on  a single random seed without claiming the variance across several seeds, mainly constrained from the high computation costs from pretraining. 
The scope of the proxy target is also limited to the general reasoning tasks, which are within the capability of small-scale models. We leave the exploration on more challenging domains, including Math, Code, and Agentic tasks for the future works.

\vspace{-1em}
\section{Conclusion}
\vspace{-1em}
In this work, we formalized \textit{OptiSelect} as the training paradigm with optimizer-dependent online in-batch selection. We provide both theoretical analysis and empirical benchmarking on how various optimization geometries can affect the preference in data selection. Our findings on \textit{OptiSelect} provide theoretical foundations and empirical insights for the next generation of pretraining optimizers, co-designed with the modern data deployment.

\clearpage
\newpage
\section*{AI Disclosure}
In this work, we \textbf{USED} generative AI tools for paper polishing and proof writing assistance, analyzing existing literature and coding assistance. 
We have \textbf{NOT} used generative AI tools for formulating research or proof ideas,  developing theoretical models or conceptual frameworks and synthetic data generation, synthetic evaluation are not applicable to this work. We have reviewed all AI-assisted work. We take responsibility for the final content of this work, including text, claims or artifacts produced with the aid of generative AI.

\section*{Reproducibility Statement}
Appendix~B specifies the optimizer update rules and scoring operators, including implementation approximations. Appendices~C--D provide the theoretical assumptions and proofs. Appendix~E documents model architectures, training configurations, selection proxies, and hyperparameter sweeps. Appendices~F--G report detailed experimental results. 
We will open-source our code at \url{https://anonymous.4open.science/r/OptiSelect-ICLR27-640B/} for reproducibility.

\section*{Ethics statement}
This work studies the interleaving dynamics between optimizers and data selection using public available datasets and evaluation protocols. 
We identify no additional ethical concerns specific to this study.

\clearpage
\newpage
\bigskip

\bibliographystyle{iclr2027_conference}
\bibliography{iclr2027_conference}

\newpage
\appendix
\addtocontents{toc}{\protect\setcounter{tocdepth}{2}}
{\centering\LARGE\bfseries Appendix\par}
\vspace{1.5em}
\tableofcontents
\clearpage
\newpage
\section{Extended Related Work}
\label{app:related_extended}

This section gives a detailed description of related work mentioned in \S\ref{sec:related}.
The organizing claim is unchanged: existing work either selects data while treating the optimizer as given, or studies optimizers while treating the data stream as given, and the two literatures barely cite each other.
OptiSelect sits at their intersection.

\subsection{Online Data Selection and Curriculum Learning}
\label{app:related_selection}

The idea that training order and sample weighting matter predates deep learning.
\citet{bengio2009curriculum} proposed easy-to-hard curricula for neural network training; \citet{loshchilov2015online} ranked samples by their current loss and sampled with exponential decay, reporting a $5\times$ speedup on MNIST; \citet{katharopoulos2019samplescreatedequaldeep} formalized the problem as importance sampling, training an auxiliary model to predict upper bounds on per-sample gradient norms.
These methods score samples by scalar losses or raw gradient norms; none accounts for the transformation the optimizer applies to a gradient before it becomes an update, which is the object this paper studies.

For LLM-scale pretraining, data selection has largely been \emph{static}: the corpus is curated once, before training, by a criterion that never sees the model state.
DSIR~\citep{xie2023dsir} performs importance resampling in a hashed n-gram feature space to match a target distribution; DCLM~\citep{li2024dclm} trains fastText classifiers for quality filtering and shows that classifier choice dominates other curation decisions at scale; FineWeb-Edu~\citep{penedo2024fineweb} distills LLM quality judgments into a lightweight scorer (the FineWeb family also supplies FineWeb-HQ, the corpus our benchmark streams); DoReMi~\citep{xie2023doremi} optimizes domain mixture weights with a small proxy model and group-DRO objective.
Static filters are cheap at training time and composable with any optimizer, but they cannot adapt to what the model currently knows, and they answer a different question from ours: \emph{what} to train on, rather than \emph{what next}.
A complementary static axis is rewriting rather than filtering: FinePhrase~\citep{huggingface2025finephrase} rephrases FineWeb documents into cleaner prose, changing the format of the text while preserving content.
Section~\ref{sec:finephrase} shows this axis composes with online selection rather than substituting for it: the selection gain is essentially unchanged at $0\%$, $50\%$ and $100\%$ rephrased shares.

\emph{Online} selection interleaves scoring with training.
MATES~\citep{yu2024mates} is the closest precursor to our setting at pretraining scale: it performs influence-guided selection during pretraining, but predicts influence with a small auxiliary model refreshed periodically, so the scores reflect the optimizer only indirectly through the checkpoints the auxiliary model is fitted to.
GREATS~\citep{wang2024greats} selects greedily within each batch using exact per-sample gradient inner products made affordable by the Ghost trick, at fine-tuning scale.
OPUS~\citep{wang2026opus} is the most principled online approach and the direct foundation of our benchmark: it computes optimizer-aware influence scores from per-sample gradients at every step and reports up to $2\times$ data efficiency at $\lambda = 2$.
OPUS, however, instantiates and validates the paradigm for \AdamW\ and \Muon\ only, and offers little analysis of \emph{how} the operator shapes selection; the paradigm's defining component---the optimizer---is exactly the axis it leaves unexplored, and the axis along which this paper varies everything.

\subsection{Influence Functions and Data Attribution}
\label{app:related_influence}

Classical influence functions~\citep{koh2017understanding} estimate the effect of removing training point $z$ on the loss at a test point $z'$ as $-\nabla_{\btheta}\cL(z')^\top H^{-1} \nabla_{\btheta}\cL(z)$, requiring an inverse-Hessian-vector product that is intractable at LLM scale and fragile in deep non-convex models.
A line of work replaces the Hessian with tractable surrogates.
TracIn~\citep{pruthi2020tracin} approximates influence by gradient dot products summed over checkpoints, which is exact for \SGD\ dynamics and explicitly ignores optimizer preconditioners.
TRAK~\citep{park2023trak} scales attribution to billion-parameter models by random projection and kernel linearization.
Data Shapley valuations can now be computed in a single training run~\citep{wang2025ghostsuite}, and LoGra~\citep{choe2024dataworthgptllmscale} makes influence queries affordable at GPT scale by projecting gradients onto low-rank subspaces during training.

Most relevant to us is the recent recognition that the optimizer's adaptive state fundamentally changes attribution.
LESS~\citep{xia2024less} extends TracIn to account for \Adam's second moment when selecting instruction-tuning data, finding that \Adam-aware selection consistently outperforms \SGD-aware selection under an \Adam-trained model.
TrackStar~\citep{chang2024scalableinfluencefacttracing} divides per-sample gradients by \Adam's $\sqrt{\hat{\bv}_t}$ before scoring, reporting that uncorrected scores preferentially retrieve long, high-magnitude examples rather than genuinely influential ones.
These works treat a single optimizer's correction as a fix to an attribution pipeline.
Our frozen-state operator $\cO_t$ (\S\ref{sec:scoring}) generalizes that move into a formalism: every optimizer induces its own influence geometry, the induced geometries differ qualitatively (diagonal, sign, polar-tangent), and the differences are large enough to reorder which optimizers benefit from selection at all (\S\ref{sec:experiments}).
Attribution also differs from our setting in \emph{when} it acts: it explains a trained model's behaviour retrospectively, whereas OptiSelect scores candidates prospectively, before the update they would cause, with the optimizer state frozen at the current step.

\subsection{Modern Optimizers for LLM Pretraining}
\label{app:related_optimizers}

\myparagraph{The families.}
\Adam~\citep{kingma2015adam} with decoupled weight decay~\citep{loshchilov2019adamw} remains the default for LLM pretraining.
\AdEMAMix~\citep{pagliardini2024ademamix} adds a third, much slower momentum buffer ($\beta_3 \approx 0.9999$, mixed in with weight $\alpha$), on the observation that gradients stay informative for tens of thousands of steps.
\Sophia~\citep{liu2023sophia} maintains a diagonal Hessian estimate via Hutchinson or Gauss--Newton--Bartlett probes and clips the preconditioned update coordinate-wise.
The sign family descends from \SignSGD\ and its momentum variant \Signum~\citep{bernstein2018signsgd}, motivated by communication compression; \Lion~\citep{chen2023lion}, discovered by program search, interpolates momentum and gradient inside the sign and stores the buffer with a separate decay.
\Muon~\citep{jordan2024muon} applies Newton--Schulz orthogonalization to the momentum matrix of each 2-D parameter, approximating the matrix-sign/polar-factor update; the \DMuon\ variant we benchmark follows \citet{semenov2025benchmarking} in handling 1-D parameters with \AdamW, and refinements continue to appear~\citep{li2026varianceadaptive}.
The Kronecker-factored line runs from Shampoo~\citep{gupta2018shampoo} through its scalable variants~\citep{anil2020scalable} to \SOAP~\citep{vyas2025soap}, which runs \Adam\ in Shampoo's eigenbasis.
Closest in spirit to our operator view, \citet{bernstein2024anthology} read modern optimizers as steepest descent under different norms; our frozen-state operator is the same structural move aimed at a different question---not what update geometry an optimizer imposes, but what \emph{influence} geometry it induces on data.

\myparagraph{Benchmarks.}
\citet{semenov2025benchmarking} benchmark eleven optimizers (\AdamW, \AdEMAMix, \MARS, \DMuon, \SOAP, \ADOPT, \Prodigy, \Sophia, \Lion, \Signum, \SFAdamW) on Llama-style models up to 720M parameters on FineWeb---the recipe our setup follows for comparability---finding that \AdEMAMix\ dominates at larger scale, sign-based methods degrade with model size, \SOAP\ wins short horizons but not long ones, and \Sophia\ diverges in small-batch regimes.
The concurrent study of \citet{wen2025fantasticpretrainingoptimizers} partially disputes \Muon's ranking and finds per-component learning-rate tuning critical.
Both benchmarks train on uniformly sampled batches; neither considers any form of data selection.
Our results give a concrete reason this omission matters: the benefit an optimizer draws from selection is an ordering of its own, stable across scale and unpredicted by standard-training quality (\S\ref{sec:experiments})---an optimizer chosen from these leaderboards may be the wrong choice for a selection pipeline.


\clearpage
\newpage
\providecommand{\finalbox}[1]{\begin{center}\fcolorbox{DarkBlue!40!white}{boxblue}{\parbox{0.94\linewidth}{#1}}\end{center}}

\section{Benchmarked Optimizers: Update Rules and Frozen-State Operators}
\label{app:operator_derivations}

This section specifies the optimizer update rules and the frozen-state operators used for data selection. 
For each optimizer discussed below, we provide (i)~pseudocode for the update rule, (ii)~the corresponding frozen-state scoring operator $\cO_t$, and (iii)~the approximations and normalization conventions used for scoring. We distinguish the implemented scoring maps from the exact nonlinear maps and derivative-based operators considered in the analysis.

\myparagraph{Derivation conventions.}
We use the following conventions:
\begin{itemize}[leftmargin=2.2em,topsep=2pt,itemsep=1pt,parsep=0pt]
\item[\textbf{(S1)}] \emph{Freeze the scoring state.} Hold the reference state $\cS_t$ fixed across candidates at step $t$, and write $\bx$ for the gradient argument. Where scoring also freezes a quantity that the optimizer normally updates using the incoming gradient, we state that approximation explicitly.
\item[\textbf{(S2)}] \emph{Separate candidate-independent additive terms.} When the frozen update is affine in $\bx$, separate its constant component from the candidate-dependent direction. An additive constant in a completed scalar score does not affect rankings or Boltzmann probabilities. Terms inside nonlinear maps cannot generally be removed this way; the scoring conventions for those maps are specified below.
\item[\textbf{(S3)}] \emph{Choose a linear scoring map.} A nonlinear batch update generally cannot be decomposed into independent per-sample directions. We therefore distinguish derivative-based linearizations from linear surrogates: the analysis considers the ideal polar derivative for \Muon, while the experiments use a right-preconditioned \Muon\ proxy and fixed-sign surrogates for \Lion/\Signum. Exact additivity of these scoring maps does not imply exact attribution under the full optimizer update.
\end{itemize}

Multiplying a completed score by a common positive scalar preserves hard rankings but rescales its variance and changes raw-score Boltzmann sampling at fixed temperature. Rescaling \(\cO_t\) need not preserve the rankings induced by~\eqref{eq:opus_score}: when \(\bG_t\) is rescaled consistently, the alignment term scales linearly and the redundancy term quadratically. The omitted prefactors therefore define the scoring convention rather than an invariance of the full utility or optimizer update.

\subsection{\texorpdfstring{\SGD}{SGD} with Momentum}
\label{app:op_sgd}

\begin{algorithm}[H]
\caption{\SGD\ with heavy-ball momentum, one step.}
\label{alg:op_sgd}
\small
\begin{algorithmic}[1]
\REQUIRE learning rate $\eta_t$; momentum $\mu$; first-order momentum buffer $\bb$ (init $\mathbf{0}$)
\STATE $\bb_t \gets \mu\,\bb_{t-1} + \hat{\bg}_t$
\STATE $\btheta_{t+1} \gets \btheta_t - \eta_t\,\bb_t$
\end{algorithmic}
\end{algorithm}

\myparagraph{Derivation.}
Substituting $\bx$ for $\hat{\bg}_t$ (S1), the update direction is $\bd(\bx) = \mu\,\bb_{t-1} + \bx$.
The frozen buffer term $\mu\,\bb_{t-1}$ is candidate-independent and drops (S2), leaving the identity map (S3):
\begin{equation}
  \cO_t^{\mathrm{\SGD}}(\bx) \;=\; \bx.
\label{eq:sgd_operator}
\end{equation}

\finalbox{\textbf{\SGD\ scoring operator.}\; $\cO_t^{\mathrm{\SGD}}(\bx) = \bx$: scoring with raw gradients.
With the momentum state fixed, the update direction is affine in the gradient. After removing the constant momentum term, the candidate-dependent map is exactly linear.}

\subsection{\texorpdfstring{\AdamW}{AdamW}}
\label{app:op_adamw}

\begin{algorithm}[H]
\caption{\AdamW~\citep{kingma2015adam,loshchilov2019adamw}, one step.}
\label{alg:op_adamw}
\small
\begin{algorithmic}[1]
\REQUIRE learning rate $\eta_t$; decays $(\beta_1,\beta_2)$; $\varepsilon$; decoupled weight decay $\lambda$; state with first- and second-moment estimates $(\bmv,\bv)$ (init $\mathbf{0}$)
\STATE $\bmv_t \gets \beta_1\,\bmv_{t-1} + (1-\beta_1)\,\hat{\bg}_t$
\STATE $\bv_t \gets \beta_2\,\bv_{t-1} + (1-\beta_2)\,\hat{\bg}_t^{\odot 2}$
\STATE $\hat{\bmv}_t \gets \bmv_t/(1-\beta_1^t)$;\quad $\hat{\bv}_t \gets \bv_t/(1-\beta_2^t)$
\STATE $\btheta_{t+1} \gets (1-\eta_t\lambda)\,\btheta_t - \eta_t\,\hat{\bmv}_t \oslash (\sqrt{\hat{\bv}_t}+\varepsilon)$
\end{algorithmic}
\end{algorithm}

\myparagraph{Derivation.}
For scoring, freeze $\bmv_{t-1}$ and the reference second-moment estimate $\hat{\bv}_t$ across candidates. This suppresses the dependence of the denominator on the candidate gradient in the full \AdamW\ update. Under this approximation, substituting $\bx$ gives
\[
  \bd(\bx)
  \;=\;
  \underbrace{\tfrac{1-\beta_1}{1-\beta_1^t}\,\bx \oslash (\sqrt{\hat{\bv}_t}+\varepsilon)}_{\text{candidate-dependent}}
  \;+\;
  \underbrace{\tfrac{\beta_1}{1-\beta_1^t}\,\bmv_{t-1} \oslash (\sqrt{\hat{\bv}_t}+\varepsilon) \;+\; \lambda\,\btheta_t}_{\text{constant for all candidates; drops by S2}} .
\]
After removing the frozen additive terms, we set the current-gradient coefficient \((1-\beta_1)/(1-\beta_1^t)\) to one as a scoring normalization, obtaining the diagonal map below.

\finalbox{\textbf{\AdamW\ scoring operator.}\; The scoring operator is linear, with exact per-sample decomposition under the frozen-denominator, unit-prefactor convention. $\cO_t^{\mathrm{\AdamW}}(\bx) = \bc_t \odot \bx$ with $\bc_t = \mathbf{1} \oslash (\sqrt{\hat{\bv}_t}+\varepsilon)$: inverse-square-root second-moment rescaling, assigning larger weights to coordinates with smaller estimated second moments.
}

\subsection{\texorpdfstring{\AdEMAMix}{AdEMAMix}}
\label{app:op_ademamix}

\begin{algorithm}[H]
\caption{\AdEMAMix~\citep{pagliardini2024ademamix}, one step. The runs use the linear $\alpha$- and $\beta_3$-warmup schedulers of the original paper.}
\label{alg:op_ademamix}
\small
\begin{algorithmic}[1]
\REQUIRE learning rate $\eta_t$; decays $(\beta_1,\beta_2,\beta_3)$; slow-buffer weight $\alpha$; $\varepsilon$; weight decay $\lambda$; state $(\bmv^{(1)},\bmv^{(2)},\bv)$
\STATE $\bmv^{(1)}_t \gets \beta_1\,\bmv^{(1)}_{t-1} + (1-\beta_1)\,\hat{\bg}_t$ \hfill\COMMENT{fast buffer, bias-corrected below}
\STATE $\bmv^{(2)}_t \gets \beta_3\,\bmv^{(2)}_{t-1} + (1-\beta_3)\,\hat{\bg}_t$ \hfill\COMMENT{slow buffer, uncorrected}
\STATE $\bv_t \gets \beta_2\,\bv_{t-1} + (1-\beta_2)\,\hat{\bg}_t^{\odot 2}$;\quad $\hat{\bmv}^{(1)}_t \gets \bmv^{(1)}_t/(1-\beta_1^t)$;\quad $\hat{\bv}_t \gets \bv_t/(1-\beta_2^t)$
\STATE $\btheta_{t+1} \gets (1-\eta_t\lambda)\,\btheta_t - \eta_t\,\bigl(\hat{\bmv}^{(1)}_t + \alpha\,\bmv^{(2)}_t\bigr) \oslash (\sqrt{\hat{\bv}_t}+\varepsilon)$
\end{algorithmic}
\end{algorithm}

\myparagraph{Derivation.}
Under the same frozen-denominator convention, the candidate enters both the fast and slow buffers, yielding the update direction:
\[
  \bd(\bx)
  \;=\;
  \Bigl(\tfrac{1-\beta_1}{1-\beta_1^t} + \alpha\,(1-\beta_3)\Bigr)\,
  \bx \oslash (\sqrt{\hat{\bv}_t}+\varepsilon)
  \;+\; \text{frozen terms},
\]
After removing the frozen additive terms and setting the displayed current-gradient coefficient to one, we obtain the same diagonal scoring form as \AdamW, evaluated at \AdEMAMix’s frozen second moment.

\finalbox{\textbf{\AdEMAMix\ scoring operator.}\; $\cO_t^{\mathrm{\AdEMAMix}}(\bx) = \bc_t \odot \bx$, with the same functional form as \AdamW\ under the stated normalization. The two methods need not produce identical scores: their update dynamics can lead to different parameter trajectories and second-moment estimates.}

\subsection{\texorpdfstring{\Sophia}{Sophia}}
\label{app:op_sophia}

\begin{algorithm}[H]
\caption{\Sophia-G~\citep{liu2023sophia}, one step. $B_H$ is the batch size used by the Hessian estimator; the trust scale is $\gamma = \rho\,B_H$.}
\label{alg:op_sophia}
\small
\begin{algorithmic}[1]
\REQUIRE learning rate $\eta_t$; decays $(\beta_1,\beta_h)$; trust parameter $\rho$; floor $\varepsilon$; Hessian interval $k_H$; weight decay $\lambda$; state $(\bmv,\bhv)$
\IF{$t \equiv 0 \pmod{k_H}$}
  \STATE draw $\hat{y} \sim p_{\btheta_t}(\cdot \mid x)$ for the mini-batch; \quad $\bg_s \gets \nabla_{\btheta}\,\ell(x,\hat{y};\btheta_t)$
  \STATE $\hat{\bhv} \gets B_H\,\bg_s \odot \bg_s$ \hfill\COMMENT{Gauss--Newton--Bartlett estimate of $\diag(\bH)$}
  \STATE $\bhv_t \gets \beta_h\,\bhv_{t-k_H} + (1-\beta_h)\,\hat{\bhv}$
\ENDIF
\STATE $\bmv_t \gets \beta_1\,\bmv_{t-1} + (1-\beta_1)\,\hat{\bg}_t$
\STATE $\btheta_{t+1} \gets (1-\eta_t\lambda)\,\btheta_t - \eta_t\,\clip\bigl(\bmv_t \oslash \max\{\gamma\,\bhv_t,\,\varepsilon\},\,1\bigr)$ \hfill\COMMENT{coordinate-wise clip at magnitude $1$}
\end{algorithmic}
\end{algorithm}

\Sophia estimates curvature from training gradients $\bg_s$,
\begin{align}
  &\hat{\bhv} \;=\; B_H\,\bg_s \odot \bg_s
  \quad\text{(GNB; the Hutchinson variant uses } \bu \odot (\bH\bu),\ \bu \sim \mathrm{Unif}\{\pm 1\}^{\dd}\text{)},
  \\\notag
  &\bhv_t \;=\; \beta_h\,\bhv_{t-k_H} + (1-\beta_h)\,\hat{\bhv}.
\label{eq:sophia_hessian}
\end{align}

\myparagraph{Derivation.}
Freeze $\bhv_t$ and substitute $\bx$ into the momentum (S1):
\[
  \bd(\bx)
  \;=\;
  \clip\Bigl(\bigl[\beta_1\,\bmv_{t-1} + (1-\beta_1)\,\bx\bigr] \oslash \max\{\gamma\,\bhv_t,\,\varepsilon\},\; 1\Bigr)
  \;+\; \lambda\,\btheta_t .
\]

\myparagraph{Implementation.}
Clipping makes the update nonlinear in $\bx$. For scoring, the implementation omits clipping and uses the \emph{unclipped} operator
\begin{equation}
  \cO_t^{\mathrm{\Sophia}}(\bx) \;=\; \bx \oslash \max\{\gamma\,\bhv_t,\,\varepsilon\}.
\label{eq:sophia_scoring_operator}
\end{equation}
Equation~\eqref{eq:sophia_scoring_operator} defines an unclipped scoring surrogate, with \(\bhv_t\) frozen, the frozen momentum contribution removed, and the current-gradient coefficient \(1-\beta_1\) set to one. Its per-sample decomposition is exact for this linear scoring map, not for the clipped optimizer update in general.

\finalbox{\textbf{\Sophia\ scoring operator.} $\cO_t^{\mathrm{\Sophia}}(\bx) \;=\; \bx \oslash \max\{\gamma\,\bhv_t,\,\varepsilon\}$. This linear surrogate assigns larger weights to coordinates with smaller estimated curvature, subject to the numerical floor.}

\subsection{\texorpdfstring{\Lion}{Lion}}
\label{app:op_lion}

\begin{algorithm}[H]
\caption{\Lion~\citep{chen2023lion}, one step. Define two decay coefficients: $\beta_1$ mixes inside the sign, $\beta_2$ updates the stored buffer.}
\label{alg:op_lion}
\small
\begin{algorithmic}[1]
\REQUIRE learning rate $\eta_t$; decays $(\beta_1,\beta_2)$; weight decay $\lambda$; state $\bmv$ (init $\mathbf{0}$)
\STATE $\bc_t \gets \beta_1\,\bmv_{t-1} + (1-\beta_1)\,\hat{\bg}_t$ \hfill\COMMENT{one-off interpolation; \emph{not} stored}
\STATE $\btheta_{t+1} \gets (1-\eta_t\lambda)\,\btheta_t - \eta_t\,\sign(\bc_t)$
\STATE $\bmv_t \gets \beta_2\,\bmv_{t-1} + (1-\beta_2)\,\hat{\bg}_t$
\end{algorithmic}
\end{algorithm}

\myparagraph{Derivation.}
Substituting $\bx$ gives the exact frozen-state operator directly:
\begin{equation}
  \cO_t^{\mathrm{\Lion}}(\bx) \;=\; \sign\bigl(\beta_1\,\bmv_{t-1} + (1-\beta_1)\,\bx\bigr).
\label{eq:lion_operator}
\end{equation}
When the frozen momentum dominates the candidate contribution, the exact operator usually agrees with the candidate-independent pattern $\bs_t = \sign(\bmv_{t-1})$. Candidate dependence then arises through coordinate sign flips. Under the Gaussian candidate model and momentum-dominance assumption of~\autoref{thm:sign_collapse}, these flips are rare tail events. The default $\beta_1=0.9$ alone does not establish this regime.

\myparagraph{Implementation.}
The implementation instead uses a \emph{fixed-sign linear surrogate}\footnote{The superscript $\mathrm{lin}$ denotes a linear surrogate, not a Taylor or Fr\'{e}chet linearization of the sign map, whose derivative is zero almost everywhere.}: the frozen pattern $\bs_t$ multiplies the candidate gradient coordinate-wise. The alignment component of the score is
$U_{z,\mathrm{align}}^{\mathrm{\Lion,lin}} = \eta_t\inner{\,\sign(\bmv_{t-1})\odot\bg_z}{\bg_\cV}$. The full utility also includes the redundancy term in~\eqref{eq:opus_score}. The collapse bound for the exact sign score does not directly apply to this surrogate.

\finalbox{\textbf{\Lion\ scoring operator.}\; Exact: $\cO_t(\bx) = \sign(\beta_1\bmv_{t-1} + (1-\beta_1)\bx)$.
Implemented: the linear surrogate map $\bx \mapsto \bs_t \odot \bx$, $\bs_t = \sign(\bmv_{t-1})$ read from the stored buffer.}

\subsection{\texorpdfstring{\Signum}{Signum}}
\label{app:op_signum}

\begin{algorithm}[H]
\caption{\Signum~\citep{bernstein2018signsgd}, one step.}
\label{alg:op_signum}
\small
\begin{algorithmic}[1]
\REQUIRE learning rate $\eta_t$; momentum decay $\mu$; state $\bmv$ (init $\mathbf{0}$)
\STATE $\bmv_t \gets \mu\,\bmv_{t-1} + (1-\mu)\,\hat{\bg}_t$
\STATE $\btheta_{t+1} \gets \btheta_t - \eta_t\,\sign(\bmv_t)$
\end{algorithmic}
\end{algorithm}

\myparagraph{Derivation.}
Substituting $\bx$ for the incoming gradient gives
\begin{equation}
  \cO_t^{\mathrm{\Signum}}(\bx) \;=\; \sign\bigl(\mu\,\bmv_{t-1} + (1-\mu)\,\bx\bigr).
\label{eq:signum_operator}
\end{equation}
This has the same form as the \Lion\ operator when $\beta_1=\beta_2=\mu$, with the same coefficient controlling both the direction used for the update and the stored momentum. The gradient-driven updates coincide under this identification; any weight-decay conventions must also be matched.
Unlike \Lion, what is signed is the freshly \emph{stored} buffer, so the frozen pattern's persistence across steps tracks $\mu$ rather than a separate memory coefficient.

\finalbox{\textbf{\Signum\ scoring operator.}\; Exact: $\cO_t(\bx) = \sign(\mu\bmv_{t-1} + (1-\mu)\bx)$.
Implemented: the fixed-sign linear surrogate $\bx\mapsto\bs_t\odot\bx$, as for \Lion, with $\bs_t = \sign(\bmv_{t-1})$. For plain \SignSGD\ ($\mu = 0$), the implementation uses \SGD\ scoring as a fallback; this is a scoring convention, not the exact sign-update map.}

\subsection{\texorpdfstring{\DMuon}{D-Muon}}
\label{app:op_dmuon}

\begin{algorithm}[H]
\caption{\DMuon~\citep{jordan2024muon,semenov2025benchmarking}, one optimizer step. Settings used throughout this paper: $\mu = 0.95$, $k_{\NS} = 5$, $r = 0.2$, $\lambda = 0.1$, and $\mathrm{nesterov} = \texttt{False}$.}
\label{alg:op_dmuon}
\small
\begin{algorithmic}[1]
\REQUIRE learning rate $\eta_t$; momentum $\mu$; Newton--Schulz steps $k_{\NS}$; RMS target $r$; weight decay $\lambda$; \AdamW\ hyperparameters $(\beta_1,\beta_2,\varepsilon_a)$ for the auxiliary path
\STATE \textbf{Partition:} $\bW$ takes the \Muon\ path iff $\bW$ is $\geq 2$-D \emph{and} $\dim_0(\bW) < 10^4$; embeddings and the LM head (vocabulary-sized first axis) therefore take the \AdamW\ path, as do all 1-D parameters.
\FOR{each \Muon-path weight $\bW \in \R^{m\times n}$ with gradient $\bG_t$}
  \STATE $\bM_t \gets \mu\,\bM_{t-1} + \bG_t$ \hfill\COMMENT{unnormalized heavy-ball buffer}
  \STATE $\bN_t \gets \bG_t + \mu\,\bM_t$ \textbf{ if } nesterov \textbf{ else } $\bM_t$ \hfill\COMMENT{matrix entering the sign map}
  \STATE $\bO_t \gets \NS_{k_{\NS}}(\bN_t) \approx \msign(\bN_t) = \bU\bV^\top$ \hfill\COMMENT{quintic iteration, \eqref{eq:ns_quintic}}
  \STATE $\bW \gets (1-\eta_t\lambda)\,\bW - \eta_t\,\sqrt{\max(m,n)}\;r\;\bO_t$
\ENDFOR
\STATE update the \AdamW-path parameters by bias-corrected \AdamW\ at the same $\eta_t$
\end{algorithmic}
\end{algorithm}

\DMuon\ does not compute an SVD. It applies $k_{\NS}$ steps of a fixed odd polynomial iteration to approximately orthogonalize the update matrix:
\begin{align}
  \bX_0 \;=\; \frac{\bN_t}{\norm{\bN_t}_F + 10^{-7}},
  \quad
  \bX_{j+1} \;=\; a\,\bX_j + \bigl(b\,\bX_j\bX_j^\top + c\,(\bX_j\bX_j^\top)^2\bigr)\bX_j,
  \quad
  \NS_{k_{\NS}}(\bN_t) \;\triangleq\; \bX_{k_{\NS}},
\label{eq:ns_quintic}
\end{align}
with $(a,b,c) = (3.4445,\,-4.7750,\,2.0315)$. The finite iteration is distinct from the ideal polar map used in the analysis below. 

\myparagraph{Derivation.}
On a 2-D parameter block, the frozen state is the momentum matrix $\bA_t = \bN_t = \bM_t$.
For the ideal polar map, a candidate perturbation with weight $\alpha$ gives $\bG \mapsto \msign(\bA_t + \alpha\,\bG)$, which is nonlinear in $\bG$. This is an idealized local perturbation model, not an identity for the finite Newton--Schulz update. 
The rigorous linear object is the Fr\'{e}chet derivative of the polar map at $\bA_t$,
\[
  \cO_t^{\mathrm{\Muon,lin}}(\bG) \;=\; \mathrm{D}\msign(\bA_t)[\bG],
\]
with full derivations and rank assumptions in Appendices~\ref{app:muon_prelim} and~\ref{app:proof_muon_spectral}. The derivative is defined per unit perturbation; a perturbation of size $\alpha$ contributes the corresponding scalar prefactor. 
Through the adjoint $\mathcal{T}_{\bA_t}$ of~\eqref{eq:muon_adjoint_operator}, the first-order score is again a linear functional of the candidate, supported on singular-subspace rotations and new column-space directions and \emph{exactly} zero on radial ones.

\myparagraph{Implementation.}
The implementation scores with the cheaper \emph{right-preconditioned proxy}
\begin{equation}
  \cO_t^{\mathrm{\Muon,rp}}(\bG) \;=\; \bG\,(\bA_t^\top\bA_t + \varepsilon\bI)^{-1/2} \;=\; \bG\,\bP_t,
\label{eq:muon_right_precond}
\end{equation}
$\bP_t$ is formed once per weight matrix by symmetric eigendecomposition and reused for $R$ scoring calls ($R=10$ in experiments). 
At a full-column-rank reference matrix and with $\varepsilon=0$, this proxy agrees with $\mathrm{D}\msign(\bA_t)$ on the column-space complement. Positive regularization changes that agreement. The proxy also retains radial candidate components that the ideal derivative annihilates; Appendix~\ref{app:muon_approx} gives the discrepancy in closed form.
\finalbox{\textbf{\DMuon\ scoring operator.}\; The ideal polar derivative $\cO_t^{\mathrm{\Muon,lin}}(\bG) = \mathrm{D}\msign(\bA_t)[\bG]$ annihilates radial perturbations under the assumptions of Theorem~\ref{thm:muon_spread}. This radial-invariance result does not extend to the implemented proxy.
Implemented: $\cO_t^{\mathrm{\Muon,rp}}(\bG) = \bG(\bA_t^\top\bA_t+\varepsilon\bI)^{-1/2}$, with the preconditioner refreshed every $R=10$ steps.
The auxiliary \AdamW\ path's parameters are excluded from scoring, as are all vocabulary-sized layers.}

\subsection{Classification of the Frozen-State Operators}
\label{app:operator_classification}

\autoref{tab:operator_summary} summarizes the operators discussed in this appendix, including the \SGD\ and \Signum\ reference methods. The following remark states what linearity of a frozen scoring map guarantees and distinguishes that property from fidelity to the optimizer update. \looseness=-1

\begin{remark}[Distributivity: when the per-sample decomposition is exact]
\label{rem:distributivity}
The batch direction represented by a frozen linear scoring map decomposes exactly into per-sample directions:
\[
  \cO_t\!\left(\frac{1}{B}\sum_z \bg_{z,t}\right)
  =
  \frac{1}{B}\sum_z \cO_t(\bg_{z,t}).
\]
This holds for \SGD, \AdamW, \AdEMAMix, and \Sophia's unclipped scoring operator~\eqref{eq:sophia_scoring_operator}. With a fixed scoring reference, these linear maps also induce linear scalar scores. By contrast, the exact clipped \Sophia\ update, coordinate-wise sign maps, and the finite \Muon\ matrix-sign map are nonlinear. We use the following linear scoring approximations: the fixed-sign surrogate for \Lion/\Signum, and the Fréchet derivative in the \Muon\ analysis or the right-preconditioned proxy in the benchmark. Their per-sample decomposition is exact for the chosen scoring map, but only approximates attribution under the nonlinear optimizer update, to first order for $\cO_t^{\mathrm{\Muon,lin}}$ and heuristically for the deployed sign and \Muon\ proxies. 
\end{remark}

\begin{table}[!t]
\caption{Properties of the frozen-state maps discussed in this appendix. Linearity guarantees additivity of the represented batch direction, not exact attribution under the full optimizer update. ``Magnitude information'' refers to retaining continuous amplitude information rather than only sign patterns; it does not imply norm preservation. $^\S$The \Sophia\ row describes its unclipped scoring surrogate. $^\ddagger$The \Muon\ row describes the ideal polar derivative; the implemented right-preconditioned proxy is also linear but retains radial components. $^\ast$The \Lion/\Signum\ rows describe the exact sign maps; their implemented fixed-sign surrogates are linear and retain candidate amplitude on coordinates with nonzero frozen signs.}
\label{tab:operator_summary}
\centering
\small
\begin{tabularx}{\linewidth}{@{}lXcp{0.24\linewidth}@{}}
\toprule
\textbf{Optimizer} & \textbf{Operator family} & \textbf{Linear in $\bx$?} & \textbf{Magnitude information} \\
\midrule
\multicolumn{4}{l}{\emph{Operators analyzed}} \\
\SGD          & Unpreconditioned (identity)     & \checkmark & \checkmark \\
\AdamW        & Diagonal adaptive (second moment)    & \checkmark & \checkmark \\
\AdEMAMix     & Diagonal adaptive (second moment)    & \checkmark & \checkmark \\
\Sophia       & Diagonal adaptive (curvature)   & \checkmark$^\S$ & \checkmark \\
\Muon         & Matrix sign / polar tangent      & \checkmark$^\ddagger$ & Tangential; radial discarded \\
\Lion$^\ast$  & Sign-based                      & $\times$   & $\times$ \\
\Signum$^\ast$ & Sign-based                      & $\times$   & $\times$ \\
\bottomrule
\end{tabularx}
\end{table}

\clearpage
\newpage
\section{The Selection-Gain Principle: Proof and Empirical Evidence}
\label{app:selection_gain_proof}


\subsection{Proof of~\autoref{thm:selection_gain}}
\label{app:sg_proof}

Let us separate two ingredients for the proof of~\autoref{thm:selection_gain}. Smoothness of the validation
objective relates the realized one-step loss reduction to the mean
utility of the trained-on batch, while the Gaussian candidate model
determines the expected utility of the uniformly sampled and
top-selected batches.

\begin{proof}
Condition on the parameters $\btheta_t$, the optimizer state, and the
frozen scoring reference at step~$t$. Let
\[
  N
  \;\triangleq\;
  |\tilde{\cB}_t|
  =
  \lambda B,
  \qquad
  q
  \;\triangleq\;
  \frac{B}{N}
  =
  \frac{1}{\lambda}.
\]
For
$a\in\{\mathrm{rand},\mathrm{sel}\}$, denote by
$\cB_t^a\subseteq\tilde{\cB}_t$ the batch returned by the corresponding
sampling rule and by
\begin{equation}
  \bd_t^a
  \;\triangleq\;
  \cO_t\!\left(
    \frac{1}{B}
    \sum_{z\in\cB_t^a}
    \bg_{z,t}
  \right)
\label{eq:sg_batch_direction}
\end{equation}
the resulting batch direction.

We first make explicit the regularity conditions required for the
one-step comparison. Assume that $\nabla F$ is
$L_{F,t}$-Lipschitz on a neighborhood containing every line segment
\[
  \left\{
    \btheta_t-s\eta_t\bd_t^a
    :
    0\leq s\leq1
  \right\},
  \qquad
  a\in\{\mathrm{rand},\mathrm{sel}\},
\]
and that the two batch directions satisfy
\begin{equation}
  \max_{
    a\in\{\mathrm{rand},\mathrm{sel}\}
  }
  \E\!\left[
    \norm{\bd_t^a}_2^2
  \right]
  \;\leq\;
  D_t^2
  <
  \infty.
\label{eq:sg_direction_moment}
\end{equation}
Assume further that the frozen map $\cO_t$ used in the present
argument is linear and is the same map used both to score candidates
and to form the batch direction. Consequently,
\begin{equation}
  \bd_t^a
  =
  \frac{1}{B}
  \sum_{z\in\cB_t^a}
  \cO_t(\bg_{z,t}).
\label{eq:sg_linearity}
\end{equation}

The descent lemma now gives, for either sampling rule,
\begin{equation}
  F(\btheta_t)
  -
  F(\btheta_t-\eta_t\bd_t^a)
  =
  \eta_t
  \inner{
    \bg_{\cV,t}
  }{
    \bd_t^a
  }
  +
  \rho_t^a,
\label{eq:sg_smooth_expansion}
\end{equation}
where
\begin{equation}
  |\rho_t^a|
  \;\leq\;
  \frac{
    L_{F,t}\eta_t^2
  }{2}
  \norm{\bd_t^a}_2^2.
\label{eq:sg_smooth_remainder}
\end{equation}
Let us clarify that this argument relies solely on local smoothness, requiring neither positive-definiteness nor a local quadratic-model assumption.

To connect~\eqref{eq:sg_smooth_expansion} to the candidate
utilities, write the frozen reference used in~\eqref{eq:opus_score} as
\begin{equation}
  \hat{\bg}_{\cV,t}
  \;\triangleq\;
  \bg_{\cV,t}
  -
  \eta_t\boldsymbol{r}_t,
\label{eq:sg_absorbed_reference}
\end{equation}
where $\boldsymbol{r}_t$ is the frozen redundancy direction. For
\[
  \mathbf{u}_{z,t}
  \;\triangleq\;
  \cO_t(\bg_{z,t}),
\]
the adopted utility score can then be written as
\begin{equation}
  U(z,t)
  =
  \eta_t
  \inner{
    \mathbf{u}_{z,t}
  }{
    \hat{\bg}_{\cV,t}
  }.
\label{eq:sg_utility_absorbed}
\end{equation}
Averaging~\eqref{eq:sg_utility_absorbed} over
$\cB_t^a$ and using~\eqref{eq:sg_linearity} gives
\begin{align}
  \bar U_{\cB_t^a}
  &\triangleq
  \frac{1}{B}
  \sum_{z\in\cB_t^a}
  U(z,t)
  \nonumber\\
  &=
  \eta_t
  \inner{
    \bd_t^a
  }{
    \bg_{\cV,t}
  }
  -
  \eta_t^2
  \inner{
    \bd_t^a
  }{
    \boldsymbol{r}_t
  }.
\label{eq:sg_batchmean_identity}
\end{align}
Combining~\eqref{eq:sg_smooth_expansion} and
\eqref{eq:sg_batchmean_identity} therefore yields
\begin{equation}
  F(\btheta_t)
  -
  F(\btheta_t-\eta_t\bd_t^a)
  =
  \bar U_{\cB_t^a}
  +
  \varepsilon_t^a,
\label{eq:sg_batchmean}
\end{equation}
where
\begin{equation}
  \varepsilon_t^a
  =
  \rho_t^a
  +
  \eta_t^2
  \inner{
    \bd_t^a
  }{
    \boldsymbol{r}_t
  }.
\end{equation}
By~\eqref{eq:sg_direction_moment} and Cauchy--Schwarz,
\begin{equation}
  \left|
    \E[\varepsilon_t^a]
  \right|
  \;\leq\;
  \eta_t^2 C_t,
  \qquad
  C_t
  \;\triangleq\;
  \frac{L_{F,t}}{2}D_t^2
  +
  \norm{\boldsymbol{r}_t}_2D_t.
\label{eq:sg_uniform_error}
\end{equation}
Thus, uniformly over the two sampling rules,
\begin{equation}
  R_t^a
  \;\triangleq\;
  \E\!\left[
    F(\btheta_t)
    -
    F(\btheta_t-\eta_t\bd_t^a)
  \right]
  =
  \E\!\left[
    \bar U_{\cB_t^a}
  \right]
  +
  \varepsilon_{t,a},
  \qquad
  |\varepsilon_{t,a}|
  \leq
  \eta_t^2C_t.
\label{eq:sg_expected_batchmean}
\end{equation}
Hence, up to a controlled second-order error, comparing one-step loss
reductions is equivalent to comparing the expected mean utilities of
the retained batches.

It remains to evaluate these two expectations. Let
$U_1,\ldots,U_N
  \stackrel{\mathrm{i.i.d.}}{\sim}
  \mathcal{N}\!\left(
    \mu_U(t),
    \sigma_U^2(t)
  \right)$
be the candidate utilities, and let $U_{(1)}\leq\cdots\leq U_{(N)}$
denote their order statistics. Under uniform sampling, conditioning on
the candidate pool gives
\begin{equation}
  \E\!\left[
    \bar U_{\cB_t^{\mathrm{rand}}}
    \,\middle|\,
    U_1,\ldots,U_N
  \right]
  =
  \frac{1}{N}
  \sum_{i=1}^{N}U_i.
\end{equation}
Taking expectation once more yields the exact identity
\begin{equation}
  \E\!\left[
    \bar U_{\cB_t^{\mathrm{rand}}}
  \right]
  =
  \mu_U(t).
\label{eq:sg_random_mean}
\end{equation}

For top-$B$ selection,
\begin{equation}
  \bar U_{\cB_t^{\mathrm{sel}}}
  =
  \frac{1}{B}
  \sum_{k=N-B+1}^{N}
  U_{(k)}.
\end{equation}
Assume first that $\sigma_U(t)>0$ and write
\[
  U_i
  =
  \mu_U(t)
  +
  \sigma_U(t)Z_i,
  \qquad
  Z_i
  \stackrel{\mathrm{i.i.d.}}{\sim}
  \mathcal{N}(0,1).
\]
Then
\begin{equation}
  \E\!\left[
    \bar U_{\cB_t^{\mathrm{sel}}}
  \right]
  =
  \mu_U(t)
  +
  c_{q,N}\sigma_U(t),
\label{eq:sg_finite_pool_superquantile}
\end{equation}
where the exact finite-pool coefficient is
\begin{equation}
  c_{q,N}
  \;\triangleq\;
  \E\!\left[
    \frac{1}{B}
    \sum_{k=N-B+1}^{N}
    Z_{(k)}
  \right].
\label{eq:sg_finite_pool_constant}
\end{equation}
The case $\sigma_U(t)=0$ is immediate: all utilities are identical and
both sampling rules have zero selection gain.

Equations~\eqref{eq:sg_expected_batchmean},
\eqref{eq:sg_random_mean}, and
\eqref{eq:sg_finite_pool_superquantile} give the exact finite-pool
relations
\begin{align}
  R_t^{\mathrm{rand}}
  &=
  \mu_U(t)
  +
  \varepsilon_{t,\mathrm{rand}},
  \nonumber\\
  R_t^{\mathrm{sel}}
  &=
  \mu_U(t)
  +
  c_{q,N}\sigma_U(t)
  +
  \varepsilon_{t,\mathrm{sel}},
\label{eq:sg_exact_finite_pool_reductions}
\end{align}
with
\[
  |\varepsilon_{t,\mathrm{rand}}|,
  |\varepsilon_{t,\mathrm{sel}}|
  \leq
  \eta_t^2C_t.
\]
Consequently,
\begin{equation}
  G_t
  \;\triangleq\;
  R_t^{\mathrm{sel}}
  -
  R_t^{\mathrm{rand}}
  =
  c_{q,N}\sigma_U(t)
  +
  \varepsilon_{t,G},
  \qquad
  |\varepsilon_{t,G}|
  \leq
  2\eta_t^2C_t.
\label{eq:sg_exact_finite_pool_gain}
\end{equation}

The population coefficient in~\autoref{thm:selection_gain} is obtained by letting
$N\to\infty$ while keeping $q=B/N\in(0,1)$ fixed. Consistency of
upper-trimmed L-statistics gives
\begin{equation}
  c_{q,N}
  \longrightarrow
  c_q
  \;\triangleq\;
  \frac{1}{q}
  \int_{1-q}^{1}
  \Phi^{-1}(u)\,\mathrm{d}u.
\label{eq:sg_population_superquantile}
\end{equation}
Equivalently, with
\[
  z_q
  \;\triangleq\;
  \Phi^{-1}(1-q),
\]
continuity of the Gaussian law gives
\begin{equation}
  c_q
  =
  \E\!\left[
    Z
    \,\middle|\,
    Z\geq z_q
  \right].
\end{equation}
Since $\phi'(z)=-z\phi(z)$,
\begin{align}
  \E\!\left[
    Z\mathbf{1}\{Z\geq z_q\}
  \right]
  &=
  \int_{z_q}^{\infty}
  z\phi(z)\,\mathrm{d}z
  \nonumber\\
  &=
  \left[
    -\phi(z)
  \right]_{z_q}^{\infty}
  =
  \phi(z_q).
\label{eq:sg_partial_expectation}
\end{align}
Dividing by
$\Pr(Z\geq z_q)=q$ yields
\begin{equation}
  c_q
  =
  \frac{\phi(z_q)}{q}.
\label{eq:sg_cq}
\end{equation}
In particular, when $\lambda=2$ and hence $q=1/2$,
\begin{equation}
  c_{1/2}
  =
  2\phi(0)
  =
  \sqrt{\frac{2}{\pi}}
  \approx
  0.79788.
\end{equation}

Thus, under the population top-$q$ idealization,~\eqref{eq:sg_exact_finite_pool_gain} becomes
\begin{equation}
  G_t
  =
  c_q\sigma_U(t)
  +
  \bigO(\eta_t^2C_t),
\end{equation}
which is~\eqref{eq:selection_gain}. For a finite candidate pool,
the mathematically exact statement is~\eqref{eq:sg_exact_finite_pool_gain}. Alternatively, invoking the
standard fixed-$q$ Gaussian L-statistic bound
$c_{q,N}=c_q+\bigO(N^{-1/2})$ gives
\begin{equation}
  G_t
  =
  c_q\sigma_U(t)
  +
  \bigO\!\left(
    \eta_t^2C_t
    +
    \frac{
      \sigma_U(t)
    }{
      \sqrt{\lambda B}
    }
  \right).
\label{eq:sg_gain_with_finite_pool}
\end{equation}
Because $c_q$ is defined after centering and standardization, it is
dimensionless and invariant under every positive affine
transformation of the utility scores. 
\end{proof}

\begin{corollary}[Relative selection gain]
\label{cor:selection_ordering}
Under the assumptions of~\autoref{thm:selection_gain}, suppose that
$\mu_U(t)>0$ and that the aggregate learning-rate and finite-pool
remainder is $o(\mu_U(t))$. Then
\begin{equation}
  \frac{G_t}{R_t^{\mathrm{rand}}}
  =
  c_q\,\frac{\sigma_U(t)}{\mu_U(t)}
  +
  o(1)
  =
  c_q\,\mathrm{CV}\!\left[U(\cdot,t)\right]
  +
  o(1),
\label{eq:relative_gain}
\end{equation}
or, equivalently,
$R_t^{\mathrm{sel}}/R_t^{\mathrm{rand}}
 = 1+c_q\,\mathrm{CV}[U(\cdot,t)]+o(1)$.
Thus, for a fixed retained fraction $q$, the relative benefit of
selection is governed by the coefficient of variation of the
optimizer-induced utility scores; the complete finite-pool and
remainder bounds are given in Appendix~\ref{app:sg_proof}.
\end{corollary}

\begin{proof}[Proof of~\autoref{cor:selection_ordering}]
Write
\begin{equation}
  R_t^{\mathrm{rand}}
  =
  \mu_U(t)
  +
  \varepsilon_{t,\mathrm{rand}},
  \qquad
  G_t
  =
  c_{q,N}\sigma_U(t)
  +
  \varepsilon_{t,G},
\end{equation}
where
\[
  |\varepsilon_{t,\mathrm{rand}}|
  \leq
  \eta_t^2C_t,
  \qquad
  |\varepsilon_{t,G}|
  \leq
  2\eta_t^2C_t.
\]
Suppose that $\mu_U(t)>0$ and
$\eta_t^2C_t\leq\mu_U(t)/2$. Direct substitution gives
\begin{align}
  \frac{
    G_t
  }{
    R_t^{\mathrm{rand}}
  }
  &=
  \frac{
    c_{q,N}\sigma_U(t)
    +
    \varepsilon_{t,G}
  }{
    \mu_U(t)
    +
    \varepsilon_{t,\mathrm{rand}}
  }
  \nonumber\\
  &=
  c_{q,N}
  \frac{
    \sigma_U(t)
  }{
    \mu_U(t)
  }
  +
  \cO\!\left(
    \frac{
      \eta_t^2C_t
    }{
      \mu_U(t)
    }
    \left[
      1
      +
      c_{q,N}
      \frac{
        \sigma_U(t)
      }{
        \mu_U(t)
      }
    \right]
  \right).
\label{eq:relative_gain_finite_pool}
\end{align}
Hence, to leading order,
\begin{equation}
  \frac{
    G_t
  }{
    R_t^{\mathrm{rand}}
  }
  =
  c_{q,N}
  \operatorname{CV}\!\left[
    U(\cdot,t)
  \right]
  +
  o(1).
\end{equation}
In the population limit,
$c_{q,N}\to c_q$, and therefore
\begin{equation}
  \frac{
    G_t
  }{
    R_t^{\mathrm{rand}}
  }
  =
  c_q
  \operatorname{CV}\!\left[
    U(\cdot,t)
  \right]
  +
  o(1).
\end{equation}
Equivalently, the multiplicative one-step improvement satisfies
\begin{equation}
  \frac{
    R_t^{\mathrm{sel}}
  }{
    R_t^{\mathrm{rand}}
  }
  =
  1
  +
  c_q
  \operatorname{CV}\!\left[
    U(\cdot,t)
  \right]
  +
  o(1).
\end{equation}
\end{proof}

\begin{remark}[Linearity and attribution error]
\label{rem:selection_gain_linearity}
The preceding loss-reduction argument applies when the same frozen linear map, with matching scale, is used to score candidates and form the batch update. Linearity holds for the specified \AdamW, \AdEMAMix, and unclipped \Sophia\ scoring maps; agreement with the full optimizer update is a separate assumption.

For any mismatched optimizer update satisfying the corresponding smoothness and moment conditions, let
\begin{equation}
  \boldsymbol{\Delta}_t^a
  \;\triangleq\;
  \bd_{t,\mathrm{upd}}^a
  -
  \frac{1}{B}
  \sum_{z\in\cB_t^a}
  \cO_{t,\mathrm{score}}(\bg_{z,t})
\end{equation}
denote the discrepancy between the direction actually applied and the
additive direction represented by the scoring map. The same proof then
acquires the additional error
\begin{equation}
  \eta_t
  \E\!\left[
    \inner{
      \bg_{\cV,t}
    }{
      \boldsymbol{\Delta}_t^a
    }
  \right].
\end{equation}
Thus, the $\bigO(\eta_t^2)$ conclusion extends to a nonlinear update only
when this attribution error is itself $\bigO(\eta_t^2)$. It is zero for
the exact linear families and is controlled to first order for the
Fr\'{e}chet-linearized polar map. For the deployed fixed-sign and
right-preconditioned proxies, the theorem describes the gain predicted
by the additive scoring geometry unless the corresponding update
mismatch is bounded separately, as discussed in
Remark~\ref{rem:distributivity}.
\end{remark}

\begin{remark}[Scope of the gain comparison]
\label{rem:gain_scope}
Theorem~\ref{thm:selection_gain} compares selection with uniform
sampling for a fixed optimizer and scoring geometry, not absolute
convergence across optimizers. For $\mu_U(t)>0$, in the population
approximation and with negligible remainder terms, the relative
extra one-step reduction is $c_q\operatorname{CV}[U(\cdot,t)]$;
the total improvement factor is
$1+c_q\operatorname{CV}[U(\cdot,t)]$.
Score spread alone does not establish alignment with the validation
objective: useful selection requires both spread and ranking
fidelity.
\end{remark}

\begin{remark}[Temperature and the empirical $2\times$ gain]
\label{rem:gain_temperature}
Boltzmann sampling without replacement interpolates between top-$B$
selection as $\tau\to0$ and uniform sampling as $\tau\to\infty$.
For nondegenerate scores, let $\kappa_t(\tau)\in[0,1]$ denote the
fraction of the expected top-$B$ score gain retained. In the Gaussian
population approximation,
\[
G_t \simeq \kappa_t(\tau)c_q\sigma_U(t).
\]
With standardized utilities, $\kappa_t$ depends on temperature and
distributional shape, not raw score scale alone.

Assuming approximately constant $\kappa_t\approx\bar\kappa$ and
$\operatorname{CV}[U]\approx\rho$, and neglecting trajectory changes,
a heuristic applied-token speedup is $1+\bar\kappa c_q\rho$.
For $q=1/2$ and $\bar\kappa\approx1$, the $2\times$ efficiency
reported by~\citet{wang2026opus} corresponds to
$\rho\approx1/c_q\approx1.25$. Indeed, this is a consistency calculation, not a convergence or compute-efficiency guarantee.
\end{remark}

\clearpage
\newpage
\section{Detailed Proofs for \textit{Discriminability} (\S\ref{sec:theory})}
\label{app:theory_proof}

This section provides proofs for each theorem in \S\ref{sec:theory} with detailed derivations and supporting assumptions (\S\ref{app:proof_sign}--\S\ref{app:proof_muon_spectral}). 

\subsection{Sign Collapse (Theorem~\ref{thm:sign_collapse})}
\label{app:proof_sign}

Throughout this subsection, we condition on the step-$t$ optimizer
state and on the frozen scoring reference $\hat{\bg}_{\cV}$.
Consequently, all expectations, probabilities, and variances are taken
only over the candidate gradient
\[
  \bg_z
  \sim
  \mathcal{N}(
    \boldsymbol{\mu},
    \boldsymbol{\Sigma}
  ),
  \qquad
  \boldsymbol{\Sigma}
  =
  \diag(
    \sigma_1^2,\ldots,\sigma_\dd^2
  ).
\]
Coordinates with
$[\hat{\bg}_{\cV}]_j\neq 0$ but $\sigma_j=0$ contribute only
candidate-independent constants to both scores and therefore contribute
zero variance. They may be removed without affecting any
discriminability statement. Hence, without loss of generality, we take
$\sigma_j>0$ on
$\mathrm{supp}(\hat{\bg}_{\cV})$. We also fix an arbitrary
$\{-1,+1\}$-valued convention for $\sign(0)$; under the preceding
non-degeneracy condition, the argument of the exact sign map equals
zero with probability zero.

\paragraph{Variance of fixed linear diagonal scores.}
We first record the variance identity used for the linearized score and,
subsequently, for the diagonal-operator comparison of
Theorem~\ref{thm:diag_advantage}. Importantly, this identity applies
only to a candidate-independent \emph{linear} diagonal map; it does not
apply directly to the exact nonlinear \Lion\ score.

Let $\boldsymbol{\pi}\in\mathbb{R}^{\dd}$ be fixed conditional on the
optimizer state, and define
\begin{equation}
  U_z^{(\boldsymbol{\pi})}
  \;\triangleq\;
  \eta_t
  \inner{
    \boldsymbol{\pi}\odot\bg_z
  }{
    \hat{\bg}_{\cV}
  }
  \;=\;
  \eta_t
  \bg_z^\top
  \bigl(
    \boldsymbol{\pi}
    \odot
    \hat{\bg}_{\cV}
  \bigr).
\end{equation}
Since only $\bg_z$ is random,
\begin{align}
  \Var\bigl[
    U_z^{(\boldsymbol{\pi})}
  \bigr]
  &=
  \eta_t^2
  \bigl(
    \boldsymbol{\pi}
    \odot
    \hat{\bg}_{\cV}
  \bigr)^\top
  \boldsymbol{\Sigma}
  \bigl(
    \boldsymbol{\pi}
    \odot
    \hat{\bg}_{\cV}
  \bigr)
  \nonumber\\
  &=
  \eta_t^2
  \sum_{j=1}^{\dd}
  \pi_j^2
  \sigma_j^2
  [\hat{\bg}_{\cV}]_j^2.
\label{eq:diag_score_variance}
\end{align}
In particular,
\begin{equation}
  \Var\bigl[
    U_z^{\mathrm{\SGD}}
  \bigr]
  =
  \eta_t^2
  \sum_{j=1}^{\dd}
  \sigma_j^2
  [\hat{\bg}_{\cV}]_j^2
  =
  \eta_t^2
  \norm{
    \boldsymbol{\Sigma}^{1/2}
    \hat{\bg}_{\cV}
  }_2^2.
\end{equation}

Whenever
$\norm{\boldsymbol{\Sigma}^{1/2}\hat{\bg}_{\cV}}_2>0$, define
\begin{equation}
  w_j
  \;\triangleq\;
  \frac{
    \sigma_j^2
    [\hat{\bg}_{\cV}]_j^2
  }{
    \sum_{k=1}^{\dd}
    \sigma_k^2
    [\hat{\bg}_{\cV}]_k^2
  },
  \qquad
  \sum_{j=1}^{\dd}w_j=1.
\label{eq:gaussian_weights}
\end{equation}
Then every comparison among fixed linear diagonal operators satisfies
\begin{equation}
  \frac{
    \Var[
      U_z^{(\boldsymbol{\pi})}
    ]
  }{
    \Var[
      U_z^{\mathrm{\SGD}}
    ]
  }
  =
  \sum_{j=1}^{\dd}
  w_j\pi_j^2.
\label{eq:spread_ratio}
\end{equation}
If
$\norm{\boldsymbol{\Sigma}^{1/2}\hat{\bg}_{\cV}}_2=0$, the
\SGD\ and linearized sign scores are both deterministic, so their
variance equality remains valid, although the normalized ratio
\eqref{eq:spread_ratio} is undefined.

\begin{assumption}[Momentum dominance]
\label{ass:momentum_dominance}
Let $0<\beta_1<1$. There exists $\kappa\geq 1$ such that
\begin{equation}
  \beta_1
  |[\bmv_{t-1}]_j|
  \;\geq\;
  \kappa(1-\beta_1)
  \bigl(
    \sigma_j+|\mu_j|
  \bigr),
  \qquad
  \forall
  j\in
  \mathrm{supp}(
    \hat{\bg}_{\cV}
  ).
\end{equation}
\end{assumption}

To interpret the assumption, define the coordinate-wise dominance
statistic
\begin{equation}
  \kappa_{t,j}
  \;\triangleq\;
  \frac{
    \beta_1
    |[\bmv_{t-1}]_j|
  }{
    (1-\beta_1)
    \bigl(
      \sigma_j+|\mu_j|
    \bigr)
  }.
\label{eq:coordinate_dominance_statistic}
\end{equation}
Assumption~\ref{ass:momentum_dominance} is equivalent to requiring
$\kappa_{t,j}\geq\kappa$ on the support of the scoring reference.
It is therefore a state-dependent and directly measurable condition,
rather than a universal property of sign-based optimization.

As a heuristic interpretation, under stationary i.i.d.\ gradients the
stored momentum buffer approaches the expectation
$\E[\bmv_t]\to\boldsymbol{\mu}$; for \Lion, this remains true even
though the stored buffer uses the separate decay coefficient
$\beta_2$. Replacing the realized buffer by its stationary mean gives
the mean-state approximation
\begin{equation}
  \kappa_{t,j}
  \approx
  \frac{
    \beta_1|\mu_j|
  }{
    (1-\beta_1)
    \bigl(
      \sigma_j+|\mu_j|
    \bigr)
  }.
\end{equation}
For $\beta_1=0.9$, the prefactor
$\beta_1/(1-\beta_1)=9$ permits a substantial dominance margin when
the gradient retains a persistent mean component. This stationary
approximation is only interpretive and is not used in the proof below.

\begin{proposition}[Linearized sign scoring is spread-neutral]
\label{prop:sign_neutral}
For every fixed sign pattern
$\bs_t\in\{-1,+1\}^{\dd}$, the linearized sign score
\begin{equation}
  U_z^{\mathrm{\Lion,lin}}
  \;\triangleq\;
  \eta_t
  \inner{
    \bg_z
  }{
    \bs_t
    \odot
    \hat{\bg}_{\cV}
  }
\end{equation}
satisfies
\begin{equation}
  \Var\bigl[
    U_z^{\mathrm{\Lion,lin}}
  \bigr]
  =
  \eta_t^2
  \sum_{j=1}^{\dd}
  \sigma_j^2
  [\hat{\bg}_{\cV}]_j^2
  =
  \eta_t^2
  \norm{
    \boldsymbol{\Sigma}^{1/2}
    \hat{\bg}_{\cV}
  }_2^2
  =
  \Var\bigl[
    U_z^{\mathrm{\SGD}}
  \bigr].
\tag{\ref{eq:var_lion}}
\end{equation}
\end{proposition}

\begin{proof}[Proof of
Proposition~\ref{prop:sign_neutral}]
The linearized score is a fixed linear functional of the candidate:
\begin{equation}
  U_z^{\mathrm{\Lion,lin}}
  =
  \eta_t
  \sum_{j=1}^{\dd}
  [\bg_z]_j
  [\bs_t]_j
  [\hat{\bg}_{\cV}]_j.
\end{equation}
Applying the fixed-diagonal variance identity
\eqref{eq:diag_score_variance} with
$\pi_j=[\bs_t]_j$ gives
\begin{align}
  \Var\bigl[
    U_z^{\mathrm{\Lion,lin}}
  \bigr]
  &=
  \eta_t^2
  \sum_{j=1}^{\dd}
  [\bs_t]_j^2
  \sigma_j^2
  [\hat{\bg}_{\cV}]_j^2
  \nonumber\\
  &=
  \eta_t^2
  \sum_{j=1}^{\dd}
  \sigma_j^2
  [\hat{\bg}_{\cV}]_j^2,
\end{align}
because $[\bs_t]_j^2=1$ for every coordinate. This is exactly the
\SGD\ score variance. Equivalently, the diagonal reflection
$\diag(\bs_t)$ preserves the covariance geometry because
\[
  \diag(\bs_t)
  \boldsymbol{\Sigma}
  \diag(\bs_t)
  =
  \boldsymbol{\Sigma}.
\]
\end{proof}

\begin{proposition}[Exponential collapse of the exact score]
\label{prop:exponential_collapse}
Suppose that, for some $\kappa\geq1$,
\begin{equation}
  \beta_1|[\bmv_{t-1}]_j|
  \;\geq\;
  \kappa(1-\beta_1)
  \bigl(\sigma_j+|\mu_j|\bigr),
  \qquad
  \forall j\in\mathrm{supp}(\hat{\bg}_{\cV}).
\label{eq:momentum_dominance}
\end{equation}
Then the exact \Lion\ score
\begin{equation}
  U_z^{\mathrm{\Lion}}
  \;\triangleq\;
  \eta_t
  \inner{
    \sign\!\bigl(
      \beta_1\bmv_{t-1}
      +(1-\beta_1)\bg_z
    \bigr)
  }{
    \hat{\bg}_{\cV}
  }
\end{equation}
satisfies
\begin{equation}
  \Var\bigl[U_z^{\mathrm{\Lion}}\bigr]
  \;\leq\;
  8e^{-\kappa^2/2}
  \eta_t^2
  \norm{\hat{\bg}_{\cV}}_2^2.
\end{equation}
Consequently, whenever
$\norm{\boldsymbol{\Sigma}^{1/2}\hat{\bg}_{\cV}}_2>0$,
{%
\setlength{\abovedisplayskip}{2pt plus 1pt minus 1pt}%
\setlength{\belowdisplayskip}{2pt plus 1pt minus 1pt}%
\setlength{\abovedisplayshortskip}{0pt plus 1pt}%
\setlength{\belowdisplayshortskip}{1pt plus 1pt minus 1pt}%
\begin{equation}
  \Var\bigl[U_z^{\mathrm{\Lion}}\bigr]
  \;\leq\;
  \Var\bigl[U_z^{\mathrm{\Lion,lin}}\bigr]
  = \Var\bigl[U_z^{\mathrm{\SGD}}\bigr]
  \quad\text{if}\quad
  \kappa^2
  \;\geq\;
  2\log\!\left(
    \frac{
      8\norm{\hat{\bg}_{\cV}}_2^2
    }{
      \norm{
        \boldsymbol{\Sigma}^{1/2}
        \hat{\bg}_{\cV}
      }_2^2
    }
  \right).
  \label{equ:rank_lion}
\end{equation}
}%
The same conclusions hold for \Signum\ after replacing $\beta_1$
with its momentum coefficient $\mu$.
\end{proposition}

\begin{proof}[Proof of
Proposition~\ref{prop:exponential_collapse}]
\textit{Step 1: define coordinate-wise sign flips}

We first apply exact decomposition into coordinate-wise sign flips.
Denote
\begin{equation}
  [\bs_t]_j
  \;\triangleq\;
  \sign(
    [\bmv_{t-1}]_j
  ),
\end{equation}
and the exact candidate-dependent sign
\begin{equation}
  \xi_{j,z}
  \;\triangleq\;
  \sign\!\left(
    \beta_1[\bmv_{t-1}]_j
    +(1-\beta_1)[\bg_z]_j
  \right).
\end{equation}
For each coordinate, introduce the flip indicator, which is the probability that the candidate gradient flips the sign of the momentum buffer at $j^{th}$ coordinate.
\begin{equation}
  F_{j,z}
  \;\triangleq\;
  \mathbf{1}\{
    \xi_{j,z}
    \neq
    [\bs_t]_j
  \},
  \qquad
  p_j
  \;\triangleq\;
  \Pr_z(
    F_{j,z}=1
  ).
\label{eq:flip_indicator}
\end{equation}
Since both $\xi_{j,z}$ and $[\bs_t]_j$ take values in
$\{-1,+1\}$,
\begin{equation}
  \E[\xi_{j,z}] = [\bs_t]_j \bigl(1-2p_j\bigr), \E[\xi_{j,z}^2]=1.
\label{eq:sign_flip_identity}
\end{equation}
Therefore, we have
\begin{equation}
    \mathrm{Var}[\xi_{j,z}] = 1-(1-2p_j)^2 = 4p_j(1-p_j) \le 4p_j.
\label{eq:sign_flip_var}
\end{equation}
According to our Gaussian candidate modelling, $\boldsymbol{\Sigma}$ is diagonal and $\bg_z$ is Gaussian,
so the coordinates $\{[\bg_z]_j\}_{j=1}^{\dd}$ are independent.
Therefore, the flip indicators are independent across coordinates.
We can write the variance of the exact score as:
\begin{align}
  \Var\bigl[
    U_z^{\mathrm{\Lion}}
  \bigr]&= \Var[\eta_t
  \sum_{j=1}^{\dd}
  \xi_{j,z}
  [\hat{\bg}_{\cV}]_j] \nonumber\\
  &=\eta_t^2
  \sum_{j=1}^{\dd}
  [\hat{\bg}_{\cV}]_j^2
  \Var[
    \xi_{j,z}
  ]
  \nonumber\\
  &=4\eta_t^2
  \sum_{j=1}^{\dd}
  [\hat{\bg}_{\cV}]_j^2
  p_j(1-p_j) \leq 4\eta_t^2
  \sum_{j=1}^{\dd}
  [\hat{\bg}_{\cV}]_j^2
  p_j.
\label{eq:exact_lion_flip_variance}
\end{align}
Thus, we find that the candidate-dependent variation in the
nonlinear sign score is carried by coordinate-wise sign reversals.

\emph{Step 2: sign flips is a Gaussian tail events under momentum dominance.}

$\forall j\in\mathrm{supp}(\hat{\bg}_{\cV})$, a flip occurs when the candidate gradient satisfies:
\begin{equation}
  \{F_{j,z}=1\}
  =
  \left\{
    [\bs_t]_j[\bg_z]_j
    <
    -a_{t,j}
  \right\}, a_{t,j}\;\triangleq\;
  \frac{
    \beta_1
    |[\bmv_{t-1}]_j|
  }{
    1-\beta_1
  }.
\label{eq:exact_flip_event}
\end{equation}
Since
\[
  [\bs_t]_j[\bg_z]_j
  \sim
  \mathcal{N}\!\left(
    [\bs_t]_j\mu_j,
    \sigma_j^2
  \right),
\]
the flip probability can be written in a Gaussian cumulative distribution function $\Phi(\cdot)$:
\begin{equation}
  p_j
  = \bP[\left\{
    [\bs_t]_j[\bg_z]_j
    <
    -a_{t,j}
  \right\} ]
  =
  \Phi\!\left(
    \frac{
      -a_{t,j}
      -[\bs_t]_j\mu_j
    }{
      \sigma_j
    }
  \right).
\label{eq:exact_flip_probability}
\end{equation}
According to Assumption~\ref{ass:momentum_dominance}, we have $a_{t,j} \geq \kappa \bigl(\sigma_j+|\mu_j|\bigr)$, therefore,
\begin{align}
  \frac{
    -a_{t,j}
    -[\bs_t]_j\mu_j
  }{
    \sigma_j
  }
  &\leq
  -\kappa
  -
  \frac{
    \kappa|\mu_j|
    +[\bs_t]_j\mu_j
  }{
    \sigma_j
  } \leq -\kappa.
\label{eq:flip_standardized_threshold}
\end{align}
The second inequality is from $\kappa|\mu_j| + [\bs_t]_j\mu_j \geq (\kappa-1)|\mu_j| \geq 0.$

Combining \eqref{eq:exact_flip_probability} and
\eqref{eq:flip_standardized_threshold} gives the sharper one-sided
bound
\begin{equation}
  p_j
  \leq
  \Phi(-\kappa).
\label{eq:flip_probability_phi_bound}
\end{equation}
In particular, for $Z\sim\mathcal{N}(0,1)$, according to Gaussian Chernoff tail bound, we have
\begin{equation}
  p_j
  \leq
  \Phi(-\kappa)
  \leq
  \Pr(
    |Z|\geq\kappa
  )
  \leq
  2e^{-\kappa^2/2}.
\label{eq:flip_probability_exp_bound}
\end{equation}

Substituting \eqref{eq:flip_probability_exp_bound} into \eqref{eq:exact_lion_flip_variance}, we obtain
\begin{align}
  \Var\bigl[
    U_z^{\mathrm{\Lion}}
  \bigr]
  &\leq
  4\eta_t^2
  \sum_{j=1}^{\dd}
  [\hat{\bg}_{\cV}]_j^2
  p_j
  \nonumber\\
  &\leq
  8e^{-\kappa^2/2}
  \eta_t^2
  \sum_{j=1}^{\dd}
  [\hat{\bg}_{\cV}]_j^2
  =
  8e^{-\kappa^2/2}
  \eta_t^2
  \norm{
    \hat{\bg}_{\cV}
  }_2^2,
\end{align}
which proves Eq.~\eqref{eq:var_lion_true}.

Finally, when
$\norm{
  \boldsymbol{\Sigma}^{1/2}
  \hat{\bg}_{\cV}
}_2>0$, combining
Eqs.~\eqref{eq:var_lion} and~\eqref{eq:var_lion_true} yields
\begin{equation}
  \frac{
    \Var[
      U_z^{\mathrm{\Lion}}
    ]
  }{
    \Var[
      U_z^{\mathrm{\Lion,lin}}
    ]
  }
  \leq
  8e^{-\kappa^2/2}
  \frac{
    \norm{
      \hat{\bg}_{\cV}
    }_2^2
  }{
    \norm{
      \boldsymbol{\Sigma}^{1/2}
      \hat{\bg}_{\cV}
    }_2^2
  }.
\label{eq:exact_to_linearized_sign_ratio}
\end{equation}
Therefore,
\begin{equation}
  \Var\bigl[
    U_z^{\mathrm{\Lion}}
  \bigr]
  \leq
  \Var\bigl[
    U_z^{\mathrm{\Lion,lin}}
  \bigr]
\end{equation}
whenever
\begin{equation}
  \kappa^2
  \geq
  2\log\!\left(
    \frac{
      8
      \norm{
        \hat{\bg}_{\cV}
      }_2^2
    }{
      \norm{
        \boldsymbol{\Sigma}^{1/2}
        \hat{\bg}_{\cV}
      }_2^2
    }
  \right).
\end{equation}

The proof uses only the generic sign-operator form
\[
  \sign\!\left(
    \alpha\bmv_{t-1}
    +(1-\alpha)\bg_z
  \right).
\]
The same proof applies to \Signum\ by replacing
$\beta_1$ with Signum's scalar momentum coefficient $\mu$, which is
distinct from the gradient-mean vector
$\boldsymbol{\mu}$.
\end{proof}

The two parts of~\autoref{thm:sign_collapse} concern distinct
scoring rules. \autoref{prop:sign_neutral} analyzes the
fixed-sign linear score denoted by
$U_z^{\mathrm{\Lion,lin}}$, whereas~\eqref{eq:var_lion_true} analyzes the exact nonlinear score
$U_z^{\mathrm{\Lion}}$. Accordingly, the fixed-diagonal spread
identity~\eqref{eq:spread_ratio} may be applied to the former but not
to the latter. Conversely, the momentum-dominance bound for the exact
score does not by itself constitute a variance bound for the
implemented linearized score.

\subsection{Diagonal-Adaptive Advantage of \AdamW and \Sophia (\autoref{thm:diag_advantage})}
\label{app:proof_thm3}

The amplification half of the ranking rests on one further empirical condition and one convexity lemma, from which the two auxiliary propositions and the theorem follow.

\begin{assumption}[Weighted variance normalization]
\label{ass:variance_normalization}
With the weights $w_j$ of~\eqref{eq:gaussian_weights}, the second-moment estimates satisfy $\sum_j w_j\,[\hat{\bv}_t]_j \leq 1$, and $[\hat{\bv}_t]_j$ is non-constant on the support of~$w$.
\end{assumption}

\autoref{ass:variance_normalization} asks that the coordinates carrying validation-relevant variance not sit where the second moments are large.
It is empirically mild since gradient clipping caps the incoming squared gradients, and the downstream proxy concentrates $w$ on rare-feature coordinates, precisely those with $[\hat{\bv}_t]_j \ll 1$.
We start the comparison on \AdamW\ and \Sophia\ from~\autoref{lem:precond_amplification}:

\begin{lemma}[Preconditioner amplification]
\label{lem:precond_amplification}
Let $w_j > 0$ be weights with $\sum_j w_j = 1$, and let $c_j > 0$ be positive scalars that are not all equal.
Then $\sum_j w_j\,c_j^2 > \bigl(\sum_j w_j\,c_j\bigr)^2$.
Moreover, for $c_j = [\bc_t]_j = 1/(\sqrt{[\hat{\bv}_t]_j}+\varepsilon)$ under~\autoref{ass:variance_normalization},
\[
  \sum_j w_j\,[\bc_t]_j^2 \;>\; \frac{1}{(1+\varepsilon')^{2}}
  \qquad\text{with}\quad
  \varepsilon' \;\triangleq\; \frac{\varepsilon}{\min_j\sqrt{[\hat{\bv}_t]_j}},
\]
and in the regime $\varepsilon \ll \min_j\sqrt{[\hat{\bv}_t]_j}$ relevant to \AdamW ($\varepsilon = 10^{-8}$),
\[
  \sum_j w_j\,[\bc_t]_j^2 \;>\; 1 - \bigO(\varepsilon').
\]
\end{lemma}

\begin{proof}[Proof of Lemma~\ref{lem:precond_amplification} (Preconditioner amplification)]
The first statement is an immediate consequence of Jensen's inequality applied to the strictly convex function $\varphi(x) = x^2$: $\E_w[\varphi(c_J)] > \varphi(\E_w[c_J])$ when $c_J$ is non-constant under the distribution $w$.

For the second statement, write $\sqrt{[\hat{\bv}_t]_j}+\varepsilon = \sqrt{[\hat{\bv}_t]_j}\,\bigl(1+\varepsilon/\sqrt{[\hat{\bv}_t]_j}\bigr) \leq \sqrt{[\hat{\bv}_t]_j}\,(1+\varepsilon')$, so that $[\bc_t]_j^2 \geq \frac{1}{(1+\varepsilon')^2}\cdot\frac{1}{[\hat{\bv}_t]_j}$ coordinate-wise; it therefore suffices to bound $\sum_j w_j/[\hat{\bv}_t]_j$.
By Jensen's inequality applied to the convex function $x \mapsto 1/x$ on $\R_{>0}$,
\[
  \sum_j \frac{w_j}{[\hat{\bv}_t]_j}
  \;\geq\;
  \frac{1}{\sum_j w_j\,[\hat{\bv}_t]_j}
  \;\geq\; 1,
\]
where the last step uses~\autoref{ass:variance_normalization}; the inequality is strict since $[\hat{\bv}_t]_j$ is non-constant on the support of~$w$.
The $\varepsilon\to 0$ statement follows since $\varepsilon' = \cO(\varepsilon)$ when $\varepsilon \ll \min_j\sqrt{[\hat{\bv}_t]_j}$.
We also discuss the role of the normalization condition; without it the conclusion can fail (e.g., $w = (\tfrac12,\tfrac12)$, $\hat{\bv}_t = (9,\,0.81)$ gives $\sum_j w_j[\bc_t]_j^2 \approx 0.67$), so~\autoref{ass:variance_normalization} is not merely technical.
\end{proof}

\begin{proposition}[Variance preconditioning amplifies spread]
\label{prop:adam_spread}
The \AdamW\ score $U_z^{\mathrm{\Adam}} = \eta_t\inner{\bg_z}{\bc_t\odot\bg_\cV}$ has
\begin{equation}
  \Var[U_z^{\mathrm{\Adam}}]
  \;=\; \eta_t^2
  \sum_{j=1}^{\dd} \sigma_j^2\, [\bc_t]_j^2\, [\bg_\cV]_j^2
  \;=\; \eta_t^2
  \norm{\boldsymbol{\Sigma}^{1/2}(\bc_t\odot\bg_\cV)}_2^2,
\label{eq:var_adam}
\end{equation}
so its spread ratio~\eqref{eq:spread_ratio} is $\sum_j w_j\,[\bc_t]_j^2$, which under~\autoref{ass:variance_normalization}, \autoref{lem:precond_amplification} bounds below:
\[
  \frac{\Var[U_z^{\mathrm{\Adam}}]}{\Var[U_z^{\mathrm{\SGD}}]}
  \;=\; \sum_{j=1}^{\dd} w_j\,[\bc_t]_j^2
  \;>\; 1 - \bigO(\varepsilon').
\]
\end{proposition}

\begin{proof}[Proof of~\autoref{prop:adam_spread} (Variance preconditioning amplifies spread)]
The \Adam influence score is $U_z^{\mathrm{\Adam}} = \eta_t\inner{\bc_t \odot \bg_z}{\bg_\cV} = \eta_t\sum_j [\bc_t]_j[\bg_z]_j[\bg_\cV]_j$.
By independence of the coordinates of $\bg_z$:
$\Var[U_z^{\mathrm{\Adam}}]
= \eta_t^2 \sum_j [\bc_t]_j^2\,[\bg_\cV]_j^2\,\sigma_j^2$, yielding~\eqref{eq:var_adam}.
Dividing by $\Var[U_z^{\mathrm{\SGD}}] = \eta_t^2\sum_j \sigma_j^2[\bg_\cV]_j^2$ gives the spread ratio $\sum_j w_j[\bc_t]_j^2$ with the weights~\eqref{eq:gaussian_weights}, and~\autoref{lem:precond_amplification} under~\autoref{ass:variance_normalization} bounds it below by $1 - \bigO(\varepsilon')$.
\end{proof}

\begin{proposition}[Curvature preconditioning amplifies spread]
\label{prop:sophia_spread}
The \Sophia\ score under the unclipped scoring operator~\eqref{eq:sophia_scoring_operator} is the diagonal-form score with pattern $\pi_j^{\mathrm{\Sophia}} = 1/\max\{\gamma[\bhv_t]_j,\,\varepsilon\}$.
Suppose the curvature estimates satisfy the normalization analogous to~\autoref{ass:variance_normalization}: $\sum_j w_j\,\gamma[\bhv_t]_j \leq 1$, with $[\bhv_t]_j$ non-constant on the support of~$w$.
Then, with $\varepsilon'' \triangleq \varepsilon/\min_j \gamma[\bhv_t]_j$,
\[
  \frac{\Var[U_z^{\mathrm{\Sophia}}]}{\Var[U_z^{\mathrm{\SGD}}]}
  \;=\; \sum_{j=1}^{\dd} w_j\,\bigl(\pi_j^{\mathrm{\Sophia}}\bigr)^2
  \;>\; \frac{1}{(1+\varepsilon'')^{2}}
  \;=\; 1 - \bigO(\varepsilon'')
\]
in the regime $\varepsilon \ll \min_j \gamma[\bhv_t]_j$.
The amplification enters through the \emph{squared} inverse curvature, concentrated on exactly the flat directions (small $[\bhv_t]_j$) that the validation-relevant weights $w_j$ occupy; the condition is measurable along training in the same way as Assumption~\ref{ass:variance_normalization}.
\end{proposition}

\begin{proof}
Since $\max\{\gamma[\bhv_t]_j, \varepsilon\} \leq \gamma[\bhv_t]_j\,(1 + \varepsilon/\gamma[\bhv_t]_j) \leq \gamma[\bhv_t]_j\,(1+\varepsilon'')$, we have $\bigl(\pi_j^{\mathrm{\Sophia}}\bigr)^2 \geq (1+\varepsilon'')^{-2}\,(\gamma[\bhv_t]_j)^{-2}$ coordinate-wise.
Applying Jensen's inequality twice, to $x \mapsto x^2$ (strictly, since $[\bhv_t]_j$ is non-constant on the support of $w$) and to the convex map $x \mapsto 1/x$,
\[
  \sum_j w_j\,(\gamma[\bhv_t]_j)^{-2}
  \;>\; \Bigl(\sum_j w_j\,(\gamma[\bhv_t]_j)^{-1}\Bigr)^{2}
  \;\geq\; \Bigl(\sum_j w_j\,\gamma[\bhv_t]_j\Bigr)^{-2}
  \;\geq\; 1,
\]
where the last step uses the normalization.
Combining the two displays gives the claim; the spread-ratio identity is~\eqref{eq:spread_ratio}.
\end{proof}

Combining~\autoref{prop:adam_spread}--\ref{prop:sophia_spread} with~\autoref{prop:exponential_collapse}~\eqref{equ:rank_lion}, we complete the proof of~\autoref{thm:diag_advantage}.

\newpage
\subsection{Radial Collapse of \Muon\ (\autoref{thm:muon_spread})}
\label{app:proof_muon_spectral}

\subsubsection{Preliminaries: the ideal polar map}
\label{app:muon_prelim}
Throughout, all matrices are real, and $\msign$ denotes the ideal
polar-factor map, not the conventional primary matrix-sign function.
The exact identities below do not describe the finite Newton--Schulz
computation in~\eqref{eq:ns_quintic}.
We condition on the momentum matrix $\bA_t$, the scoring reference
$\bR=\hat{\bG}_\cV\in\R^{m\times n}$, and all scoring coefficients.
All probabilistic statements are conditional on these frozen quantities;
only the candidate gradient $\bG_z$ is random.

\begin{assumption}[Regular polar state]
\label{ass:muon_full_rank}
The matrix $\bA_t\in\R^{m\times n}$, $m\ge n\ge1$, has full column
rank. Fix a compact SVD
\begin{equation}
\bA_t=\bU\boldsymbol{\Sigma}\bV^\top,
\qquad
\boldsymbol{\Sigma}=\diag(\sigma_1,\ldots,\sigma_n),
\qquad
\sigma_1\ge\cdots\ge\sigma_n>0,
\label{eq:muon_svd}
\end{equation}
where $\bU^\top\bU=\bV^\top\bV=\bI_n$ and $\bV$ is square.
Repeated singular values are allowed.
\end{assumption}

The polar decomposition is $\bA_t=\bQ\bH$, where
\begin{equation}
\bQ=\msign(\bA_t)
=\bA_t(\bA_t^\top\bA_t)^{-1/2}=\bU\bV^\top,
\qquad
\bH=\bV\boldsymbol{\Sigma}\bV^\top\succ0.
\label{eq:msign_rect}
\end{equation}
In particular, $\bQ^\top\bQ=\bI_n$.
The polar-derivative statements for full-row-rank wide matrices follow
by transposition. The right-preconditioned proxy is analyzed below
in the displayed $m\ge n$ orientation.

\begin{proposition}[Spectral-norm steepest descent]
\label{prop:msign_steepest}
For full-column-rank $\bA=\bU\boldsymbol{\Sigma}\bV^\top$,
$\msign(\bA)=\bU\bV^\top$ is the unique maximizer of
$\inner{\bA}{\bD}_F$ over $\norm{\bD}_{\mathrm{op}}\le1$.
\end{proposition}
\begin{proof}
For any feasible $\bD$, its alignment with the singular vectors satisfies
\[
\inner{\bA}{\bD}_F
=\sum_{i=1}^n\sigma_i\mathbf u_i^\top\bD\mathbf v_i
\le\sum_{i=1}^n\sigma_i,
\]
with equality at $\bD=\bU\bV^\top$.
Since every $\sigma_i>0$, equality requires
$\mathbf u_i^\top\bD\mathbf v_i=1$ for every $i$.
Together with $\norm{\bD\mathbf v_i}_2\le1$, this implies
$\bD\mathbf v_i=\mathbf u_i$.
The vectors $\mathbf v_i$ form a basis, so $\bD=\bU\bV^\top$.
\end{proof}

\subsubsection{Polar derivative and radial geometry}
Fix an orthonormal complement
$\bU_\perp\in\R^{m\times(m-n)}$ so that
$[\bU\ \bU_\perp]$ is orthogonal. Define
\begin{align}
\bB_z&=\bU^\top\bG_z\bV,
&\bC_z&=\bU_\perp^\top\bG_z\bV,
\label{eq:muon_candidate_coordinates}\\
\bR_\parallel&=\bU^\top\bR\bV,
&\bR_\perp&=\bU_\perp^\top\bR\bV.
\label{eq:muon_reference_coordinates}
\end{align}
When $m=n$, complement-block terms are absent.
Unless indicated otherwise, $i,j$ range over $1,\ldots,n$ and
$a$ over $1,\ldots,m-n$.

The radial subspace and its Frobenius-orthogonal complement are
\[
\begin{aligned}
\mathcal R_t
&=\{\bU\bS\bV^\top:\bS\in\R^{n\times n},\ \bS=\bS^\top\},\\
\mathcal R_t^\perp
&=\{\bX\in\R^{m\times n}:\bQ^\top\bX+\bX^\top\bQ=\mathbf0\}.
\end{aligned}
\]
Equivalently, $\mathcal R_t=\{\bQ\bS:\bS=\bS^\top\}$:
radial perturbations change the positive polar factor at fixed $\bQ$.
Writing $\operatorname{sym}(\bX)=(\bX+\bX^\top)/2$, define
\begin{equation}
\Pi^{\mathrm{rad}}\bM
=\bU\operatorname{sym}(\bU^\top\bM\bV)\bV^\top,
\qquad
\Pi^{\mathrm{tan}}=\mathrm{Id}-\Pi^{\mathrm{rad}}.
\label{eq:muon_radial_projectors}
\end{equation}
These are the orthogonal projections onto $\mathcal R_t$ and
$\mathcal R_t^\perp$, respectively, since symmetric and skew-symmetric
matrices are Frobenius-orthogonal and the two column-space blocks
are orthogonal. Set $\bG_z^a=\Pi^a\bG_z$ for
$a\in\{\mathrm{rad},\mathrm{tan}\}$.

\begin{theorem}[Exact first-order geometry of matrix-sign scoring]
\label{thm:muon_spectral_full}
Under Assumption~\ref{ass:muon_full_rank}, let
$\mathcal L_t=\mathrm D\msign(\bA_t)$.
For any $\bG\in\R^{m\times n}$, set
$\bB=\bU^\top\bG\bV$ and $\bC=\bU_\perp^\top\bG\bV$. Then
\begin{equation}
\begin{aligned}
\mathcal L_t[\bG]
&=\bU\boldsymbol{\Omega}(\bB)\bV^\top
+\bU_\perp\bC\boldsymbol{\Sigma}^{-1}\bV^\top,\\
[\boldsymbol{\Omega}(\bB)]_{ij}
&=\frac{[\bB]_{ij}-[\bB]_{ji}}{\sigma_i+\sigma_j},
\qquad 1\le i,j\le n.
\end{aligned}
\label{eq:muon_derivative_explicit}
\end{equation}
The derivative is self-adjoint for the Frobenius inner product:
\begin{equation}
\mathcal T_{\bA_t}\triangleq\mathcal L_t^*=\mathcal L_t.
\label{eq:muon_adjoint_operator}
\end{equation}
Moreover, with $\delta_t=\sigma_n/2$, there exists $C_t<\infty$,
depending only on $\bA_t$, such that
\begin{equation}
\begin{aligned}
\Phi_z(\alpha)
&\triangleq\inner{\msign(\bA_t+\alpha\bG_z)}{\bR}_F\\
&=\inner{\bQ}{\bR}_F
+\alpha\inner{\bG_z}{\mathcal L_t[\bR]}_F+r_z(\alpha),\\
|r_z(\alpha)|
&\le C_t\norm{\bR}_F\alpha^2\norm{\bG_z}_F^2
\end{aligned}
\label{eq:muon_expansion_full}
\end{equation}
for every real $\alpha$ and every $\bG_z$ satisfying
$|\alpha|\norm{\bG_z}_F\le\delta_t$.
\end{theorem}

\begin{proof}
The polar map is smooth on the open set of full-column-rank matrices.
Differentiate $\bA_t=\bQ\bH$ in direction $\bG$ to obtain
$\bG=\dot{\bQ}\bH+\bQ\dot{\bH}$.
Since $\bQ^\top\bQ=\bI_n$ and $\dot{\bH}$ is symmetric,
$\bK=\bQ^\top\dot{\bQ}$ is skew-symmetric, and
\[
\begin{aligned}
\bK\bH+\bH\bK&=\bQ^\top\bG-\bG^\top\bQ,\\
(\bI_m-\bQ\bQ^\top)\dot{\bQ}
&=(\bI_m-\bQ\bQ^\top)\bG\bH^{-1}.
\end{aligned}
\]
In the $\bV$ basis, the first identity becomes
$\boldsymbol{\Omega}\boldsymbol{\Sigma}
+\boldsymbol{\Sigma}\boldsymbol{\Omega}=\bB-\bB^\top$,
where $\boldsymbol{\Omega}=\bV^\top\bK\bV$.
Solving entrywise and using
$\dot{\bQ}=\bQ\bK+(\bI_m-\bQ\bQ^\top)\dot{\bQ}$
gives~\eqref{eq:muon_derivative_explicit}.

In the orthogonal block coordinates $(\bB,\bC)$, the map on each
pair $([\bB]_{ij},[\bB]_{ji})$, $i<j$, has the symmetric matrix
\[
\frac{1}{\sigma_i+\sigma_j}
\begin{pmatrix}1&-1\\-1&1\end{pmatrix};
\]
diagonal entries are sent to zero, and each $[\bC]_{aj}$ is multiplied
by $\sigma_j^{-1}$. Thus $\mathcal L_t$ is self-adjoint.
Finally, the closed Frobenius ball of radius $\delta_t=\sigma_n/2$
around $\bA_t$ lies in the full-column-rank set.
The second derivative is bounded on this compact ball, so Taylor's
theorem and Cauchy--Schwarz give~\eqref{eq:muon_expansion_full}.
\end{proof}

\begin{corollary}[Exact radial nullspace]
\label{cor:muon_nullspace}
Under Assumption~\ref{ass:muon_full_rank},
\[
\ker\mathcal L_t=\mathcal R_t,
\qquad
\mathcal L_t=\mathcal L_t\Pi^{\mathrm{tan}}
=\Pi^{\mathrm{tan}}\mathcal L_t.
\]
\end{corollary}
\begin{proof}
The two terms in~\eqref{eq:muon_derivative_explicit} are orthogonal
and vanish simultaneously exactly when $\bB=\bB^\top$ and $\bC=0$.
This gives the kernel and the first projection identity; self-adjointness
gives the second.
\end{proof}

\begin{corollary}[Exact radial invariance]
\label{cor:muon_radial_invariance}
Under Assumption~\ref{ass:muon_full_rank}, for every symmetric
$\bS\in\R^{n\times n}$ and every real $\alpha$,
\begin{equation}
\msign(\bA_t+\alpha\bU\bS\bV^\top)=\bU\bV^\top
\quad\text{whenever}\quad
\boldsymbol{\Sigma}+\alpha\bS\succ0.
\label{eq:muon_radial_invariance}
\end{equation}
\end{corollary}
\begin{proof}
The factorization
$\bA_t+\alpha\bU\bS\bV^\top
=(\bU\bV^\top)
[\bV(\boldsymbol{\Sigma}+\alpha\bS)\bV^\top]$
is its polar decomposition.
\end{proof}
A diagonal $\bS$ changes only singular values in the chosen basis.
A general symmetric $\bS$ can also change the eigenvectors of the
positive polar factor while leaving $\bQ$ unchanged.
The finite invariance above concerns purely radial perturbations;
for arbitrary candidates differing radially, equality is guaranteed
only for their first-order scores at the frozen state.

\subsubsection{Score variance and proof of \autoref{thm:muon_spread}}
Fix $\eta_t>0$ and $\alpha\ge0$, and assume
$\mathbb E\norm{\bG_z}_F^2<\infty$.
Define
\[
\begin{aligned}
U_z^{\Muon}
&=\eta_t\alpha\inner{\mathcal L_t[\bG_z]}{\bR}_F
=\eta_t\alpha\inner{\bG_z}{\mathcal L_t[\bR]}_F,\\
U_z^a&=\eta_t\inner{\bG_z^a}{\bR}_F,
\qquad a\in\{\mathrm{rad},\mathrm{tan}\},\\
U_z^{\SGD}&=\eta_t\inner{\bG_z}{\bR}_F
=U_z^{\mathrm{rad}}+U_z^{\mathrm{tan}}.
\end{aligned}
\]
Thus $U_z^{\Muon}$ is the candidate-dependent first-order term of
$\eta_t\Phi_z(\alpha)$.

\begin{corollary}[General covariance formula]
\label{cor:muon_spread}
Under the frozen-state and second-moment conditions above, let
$\boldsymbol{\Gamma}=\Cov(\operatorname{vec}\bG_z)$ and
$\mathbf t=\operatorname{vec}(\mathcal L_t[\bR])$. Then
\begin{equation}
\Var(U_z^{\Muon})
=\eta_t^2\alpha^2\mathbf t^\top\boldsymbol{\Gamma}\mathbf t.
\label{eq:muon_general_variance}
\end{equation}
No coordinate-uncorrelation assumption is needed.
\end{corollary}
\begin{proof}
Apply the covariance identity for the linear functional
$U_z^{\Muon}=\eta_t\alpha\,\mathbf t^\top
\operatorname{vec}(\bG_z)$.
\end{proof}

For the coordinate formulas and \autoref{thm:muon_spread}, additionally
assume that the entries of the combined collection $(\bB_z,\bC_z)$
are pairwise uncorrelated, including cross-block pairs, and write
\[
\tau_{ij}^2=\Var([\bB_z]_{ij}),
\qquad
\nu_{aj}^2=\Var([\bC_z]_{aj}).
\]
These are entry variances, not variances of antisymmetric differences.
Uncorrelation is assumed in the fixed singular-vector basis and does
not follow merely from diagonal covariance in the original parameter
coordinates. Neither Gaussianity nor a zero-mean assumption is required.

\begin{proof}[Proof of \autoref{thm:muon_spread}]
Corollary~\ref{cor:muon_nullspace} gives
$\mathcal L_t[\bG_z]=\mathcal L_t[\bG_z^{\mathrm{tan}}]$,
so changing only the radial component leaves the score unchanged.
By self-adjointness and~\eqref{eq:muon_derivative_explicit},
\[
U_z^{\Muon}
=\eta_t\alpha\!\left[
\sum_{i\ne j}[\bB_z]_{ij}
\frac{[\bR_\parallel]_{ij}-[\bR_\parallel]_{ji}}
{\sigma_i+\sigma_j}
+\sum_{a,j}[\bC_z]_{aj}\frac{[\bR_\perp]_{aj}}{\sigma_j}
\right].
\]
Pairwise uncorrelation yields
\begin{equation}
\Var(U_z^{\Muon})
=\eta_t^2\alpha^2\!\left[
\sum_{i\ne j}\tau_{ij}^2
\left(
\frac{[\bR_\parallel]_{ij}-[\bR_\parallel]_{ji}}
{\sigma_i+\sigma_j}
\right)^2
+\sum_{a,j}\nu_{aj}^2
\left(\frac{[\bR_\perp]_{aj}}{\sigma_j}\right)^2
\right].
\label{eq:muon_coordinate_variance}
\end{equation}
Likewise, the tangential projection gives
\[
\Var(U_z^{\mathrm{tan}})
=\eta_t^2\!\left[
\sum_{i\ne j}\tau_{ij}^2
\left(
\frac{[\bR_\parallel]_{ij}-[\bR_\parallel]_{ji}}{2}
\right)^2
+\sum_{a,j}\nu_{aj}^2[\bR_\perp]_{aj}^2
\right].
\]
Since $2/(\sigma_i+\sigma_j)\le1/\sigma_n$ and
$1/\sigma_j\le1/\sigma_n$, termwise comparison gives
\[
\Var(U_z^{\Muon})
\le\frac{\alpha^2}{\sigma_n^2}\Var(U_z^{\mathrm{tan}}),
\]
which is~\eqref{eq:muon_ranking_slot}.
\end{proof}

\paragraph{Comparison with \SGD.}
Orthogonality of the matrix projections does not imply uncorrelation
of the projected scalar scores. Indeed,
\[
\Var(U_z^{\SGD})
=\Var(U_z^{\mathrm{tan}})+\Var(U_z^{\mathrm{rad}})
+2\Cov(U_z^{\mathrm{tan}},U_z^{\mathrm{rad}}),
\]
and expansion under the coordinate assumptions above gives
\[
\Cov(U_z^{\mathrm{tan}},U_z^{\mathrm{rad}})
=\frac{\eta_t^2}{4}\sum_{i<j}
(\tau_{ij}^2-\tau_{ji}^2)
\bigl([\bR_\parallel]_{ij}^2-[\bR_\parallel]_{ji}^2\bigr).
\]
For example, transpose-pair variance symmetry
$\tau_{ij}^2=\tau_{ji}^2$ makes this covariance zero; together with
$\alpha\le\sigma_n$, it is sufficient for
$\Var(U_z^{\Muon})\le\Var(U_z^{\SGD})$.
There is no unconditional ordering: for $\bA_t=s\bI_2$, $s>0$,
a nonzero skew-symmetric reference $\bR$, and independent unit-variance
entries of $\bG_z$, the ratio is $\alpha^2/s^2$, which exceeds one
when $0<s<\alpha$.

\subsection{Right-Preconditioned \Muon\ Proxy and Exact Error}
\label{app:muon_approx}
For fixed $\varepsilon\ge0$, write
\[
\bP_\varepsilon
=(\bA_t^\top\bA_t+\varepsilon\bI_n)^{-1/2}
=\bV\bD_\varepsilon\bV^\top,
\qquad
\bD_\varepsilon
=\diag\!\left((\sigma_j^2+\varepsilon)^{-1/2}\right)_{j=1}^n.
\]
The score with the normalization of~\eqref{eq:operator_zoo} is
$U_z^{\mathrm{rp}}=\eta_t\inner{\bG_z\bP_\varepsilon}{\bR}_F$.
Since
\[
\bG_z\bP_\varepsilon
=\bU\bB_z\bD_\varepsilon\bV^\top
+\bU_\perp\bC_z\bD_\varepsilon\bV^\top,
\]
taking the inner product with $\bR$ and using the same coordinate
uncorrelation assumption gives
\begin{equation}
\Var(U_z^{\mathrm{rp}})
=\eta_t^2\!\left[
\sum_{i,j}\tau_{ij}^2
\frac{[\bR_\parallel]_{ij}^2}{\sigma_j^2+\varepsilon}
+\sum_{a,j}\nu_{aj}^2
\frac{[\bR_\perp]_{aj}^2}{\sigma_j^2+\varepsilon}
\right].
\label{eq:muon_rp_variance}
\end{equation}
Without coordinate uncorrelation, the corresponding identity is
$\Var(U_z^{\mathrm{rp}})=\eta_t^2\mathbf p^\top
\boldsymbol{\Gamma}\mathbf p$, where
$\mathbf p=\operatorname{vec}(\bR\bP_\varepsilon)$.
Both scores are fixed linear functionals of
$\operatorname{vec}(\bG_z)$; if this vector is Gaussian, the scores
are jointly Gaussian, possibly degenerate.

For any $\bG=\bU\bB\bV^\top+\bU_\perp\bC\bV^\top$,
subtracting the proxy from~\eqref{eq:muon_derivative_explicit} gives
\begin{equation}
\begin{aligned}
\mathcal L_t[\bG]-\bG\bP_\varepsilon
={}&\bU\bigl(\boldsymbol{\Omega}(\bB)-\bB\bD_\varepsilon\bigr)\bV^\top\\
&+\bU_\perp\bC
\bigl(\boldsymbol{\Sigma}^{-1}-\bD_\varepsilon\bigr)\bV^\top.
\end{aligned}
\label{eq:muon_proxy_error_exact}
\end{equation}
The two blocks are orthogonal, hence
\begin{equation}
\begin{aligned}
\norm{\mathcal L_t[\bG]-\bG\bP_\varepsilon}_F^2
={}&\norm{\boldsymbol{\Omega}(\bB)-\bB\bD_\varepsilon}_F^2\\
&+\norm{\bC(\boldsymbol{\Sigma}^{-1}-\bD_\varepsilon)}_F^2.
\end{aligned}
\label{eq:muon_proxy_error_norm}
\end{equation}
At $\varepsilon=0$, the proxy agrees with the derivative on the
column-space complement, but generally not on the parallel block.
For $\varepsilon>0$, the complement weights also differ when that
block is present. Since $\bP_\varepsilon$ is invertible, the proxy
annihilates no nonzero matrix, including radial perturbations.
Their scalar alignments can nevertheless vanish for particular
references. Thus radial invariance of the ideal polar map is not
a radial-collapse guarantee for the deployed proxy.

\paragraph{Normalization and scope.}
The coefficient $\alpha$ is the candidate's weight in the local
matrix perturbation, determined by the batch and momentum conventions;
it is not universally $1/B$.
The displayed proxy score omits this weight, as in
\eqref{eq:operator_zoo}. For equally weighted scores,
\[
|U_z^{\Muon}-\alpha U_z^{\mathrm{rp}}|
\le\eta_t\alpha\norm{\bR}_F
\norm{\mathcal L_t[\bG_z]-\bG_z\bP_\varepsilon}_F.
\]
Multiplying any score by a fixed scalar multiplies its variance by the
square of that scalar; any additional block-update rescaling must
therefore be retained in calibrated comparisons.
All variance identities above concern the specified linear scores,
not the finite nonlinear alignment $\eta_t\Phi_z(\alpha)$.
The Taylor estimate in~\eqref{eq:muon_expansion_full} is local and
must be controlled separately when relating their variances.

\clearpage
\newpage
\subsection{Optimal Candidate Multiplier (\autoref{lem:lambda_star})}
\label{app:lambda_star_proof}
We provide the detailed proof of \autoref{lem:lambda_star} here. 
\begin{proof}
Following~\autoref{thm:selection_gain}, the per-step selection gain is $G_t = c_q\,\sigma_U(t)$ with $q = 1/\lambda$.
A step draws $\tilde{B} = \lambda B$ candidate rows and trains on $B$ of them. Denote sequence length $L$, \textit{OptiSelect} \emph{streams and scores} $\lambda B L$ tokens while \emph{training the model on} $BL$.
Therefore, under a budget counted in tokens read, we quantify the selection gain per-read-token as
\begin{equation}
  \frac{G_t}{\lambda B L} \;\propto\; \frac{c_q(\lambda)}{\lambda} \;=:\; f(\lambda),
  \qquad \lambda \in (1,\infty).
\label{eq:lambda_objective}
\end{equation}

Let $U \sim \mathcal{N}(\mu, \sigma^2)$, define the density, CDF and upper tail functions for $z \sim \mathcal{N}(0,1)$ as
\[
  \phi(z) \;=\; \tfrac{1}{\sqrt{2\pi}}\,e^{-z^{2}/2},
  \qquad
  \Phi(z) \;=\; \int_{-\infty}^{z}\phi(u)\,\mathrm{d} u,
  \qquad
  \bar{\Phi} \;=\; 1-\Phi,
\] 
with $\Phi^{-1}$ denoting the quantile function.

Let hard top-$q$ selection retain the largest $q$-fraction, i.e.\ those $U$ above the $(1-q)$-quantile $u_q = \mu + \sigma\,\Phi^{-1}(1-q)$.
Writing $Z = (U-\mu)/\sigma \sim \mathcal{N}(0,1)$ and $a = \Phi^{-1}(1-q)$, the truncated-normal mean is
\begin{equation}
  \E[Z \mid Z > a] \;=\; \frac{1}{\Pr[Z>a]}\int_a^{\infty} z\,\phi(z)\,\mathrm{d} z
  \;=\; \frac{[-\phi(z)]_a^{\infty}}{1-\Phi(a)}
  \;=\; \frac{\phi(a)}{1-\Phi(a)},
\label{eq:trunc_mean}
\end{equation}
using $\phi'(z) = -z\phi(z)$, so that $\int_a^\infty z\phi(z)\mathrm{d} z = \phi(a)$.
By construction $1-\Phi(a) = q = \frac{1}{\lambda}$, hence selection lifts the mean utility by $\sigma\,\phi(a)/q$ and
\begin{equation}
  c_q \;=\; \frac{\phi\bigl(\Phi^{-1}(1-q)\bigr)}{q}.
\label{eq:cq_closed_form}
\end{equation}

Therefore, plug in~\eqref{eq:lambda_objective}, we have:
\begin{equation}
  f(\lambda) \;=\; \frac{c_q(\lambda)}{\lambda} \;=\; \phi\bigl(\Phi^{-1}(1-1/\lambda)\bigr).
\label{eq:f_identity}
\end{equation}

Write $z(\lambda) = \Phi^{-1}(1-1/\lambda)$, so $\Phi(z) = 1 - 1/\lambda$. Differentiating implicitly,
\[
  \phi(z)\,z'(\lambda) \;=\; \frac{1}{\lambda^2}
  \qquad\Longrightarrow\qquad
  z'(\lambda) \;=\; \frac{1}{\lambda^2\,\phi(z)} \;>\; 0 .
\]
Then, plug in $\phi'(z) = -z\,\phi(z)$ with the chain rule, we have:
\begin{equation}
  f'(\lambda) \;=\; \phi'(z)\,z'(\lambda) \;=\; \bigl(-z\,\phi(z)\bigr)\cdot\frac{1}{\lambda^2\phi(z)}
  \;=\; -\,\frac{z(\lambda)}{\lambda^{2}}
  \;=\; -\,\frac{\Phi^{-1}(1-1/\lambda)}{\lambda^{2}} .
\label{eq:f_derivative}
\end{equation}

We then find the optimum of~\eqref{eq:f_identity} by introducing an involution $\lambda \mapsto \lambda/(\lambda-1)$, which maps $(1,\infty)$ onto itself and corresponds to $q \mapsto 1-q$.
Because $\Phi^{-1}(q) = -\Phi^{-1}(1-q)$ and $\phi$ is even,~\eqref{eq:f_identity} gives
\[
  f\!\left(\frac{\lambda}{\lambda-1}\right) \;=\; \phi\bigl(\Phi^{-1}(1/\lambda)\bigr) \;=\; \phi\bigl(-\Phi^{-1}(1-1/\lambda)\bigr) \;=\; f(\lambda),
\]
so the efficiency curve is symmetric under that involution, i.e. $f(1.5) = f(3)$ and $f(1.25) = f(5)$ exactly.
An involution of an interval has exactly one fixed point here, $\lambda = \lambda/(\lambda-1) \iff \lambda = 2$, and a strictly unimodal function invariant under it must attain its extremum there. 

Therefore, the multiplier $\lambda^{\star}=2$ sets the unique optimum.

\end{proof}

\clearpage
\newpage
\section{Experiment Configurations}
\label{app:exp_configs}
\subsection{Model Architecture}
\label{app:model_configs}
\paragraph{Configurations.} 
We follow Llama architecture with RMSNorm pre-normalization, SwiGLU and RoPE positional embedding for all the LLM pretraining runs in Section \ref{sec:experiments}. We apply GPT2 BPE tokenizer with vocabulary size $50{,}304$.
\begin{table}[ht!]
\caption{Model architecture configurations.}
\label{tab:model_configs}
\centering
\small
\begin{tabular}{lccccc}
\toprule
\textbf{Scale} & $d_{\mathrm{model}}$ & \textbf{layers} & \textbf{heads} & \textbf{seq.\ len.} & \textbf{tokens/step} \\
\midrule
124M & 768  & 12 & 12 & 512 & 32{,}768 \\
720M & 2048 & 12 & 16 & 512 & 131{,}072 \\
\bottomrule
\end{tabular}
\end{table}

\paragraph{Infrastructure.}
All runs use single GH200 nodes (4 GPUs, DDP) on the Clariden cluster, inside an NGC PyTorch container. FineWeb-HQ and FinePhrase datasets are loaded in streaming.

\subsection{Training Arguments}
\label{app:train_args}

On both 124M and 720M model runs, we apply cosine schedule decaying to $0.01\times\eta_{\max}$ with $2{,}000$ warmup steps and weight decay $0.1$, the optimal learning rate are determined by a grid sweep per-optimizer and per-scale, according to Appendix \ref{app:lr_sweep}. 
The 124M models are trained with a global batch size $64\times512=32{,}768$ tokens across 4 GH200 GPUs, while the 720M models are trained with a global batch size $256\times512=131{,}072$ tokens On both standard training and \textit{OptiSelect}, the models are train from scratch for $128$k steps, consuming 4.2B tokens in total.


\subsection{Selection Proxy Details}
\label{app:proxy_details}

The proxy set holds $4{,}096$ densely packed $512$-token sequences split evenly over ARC-Easy, ARC-Challenge, HellaSwag, PIQA, and SciQ, each document rendered as the task prompt concatenated with its \emph{correct} completion. 
The proxy gradient $\bg_{\cV,t}$ is recomputed every refresh interval ($1000$ steps) on the corresponding model checkpoint. The proxy refresh interval is determined by the ablation of Appendix~\ref{app:refresh_sweep}.

\subsection{Learning Rate Sweep}
\label{app:lr_sweep}

Switching a run to OptiSelect changes which data the optimizer sees, and could
in principle move the optimizer's best step size with it.
If it did, every OptiSelect run would need its own learning-rate search and the
cost of using the method would roughly double.
This appendix settles the question at 124M, and the answer is what licenses the
cheaper protocol used at 720M.

\paragraph{LR Sweep on 124M runs}
On 124M models, we grid-search the learning rate over four geometrically spaced rates across 7 optimizers, on OptiSelect and standard training respectively.

\begin{table}[ht!]
\caption{\textbf{Downstream loss (``dwn'') and FineWeb Validation loss (``val'') across a sweep of learning rates at 124M.}  Bold marks the optimal learning rate on the \textit{downstream loss}, which are taken for experiments.}
\label{tab:lr_sweep}
\centering
\small
\setlength{\tabcolsep}{4pt}
\begin{tabular}{llcccc}
\toprule
\textbf{Optimizer} & \textbf{Mode} & \multicolumn{4}{c}{\textbf{val / dwn} at $\eta = \{0.25,\,0.5,\,1,\,2\}\times$ default} \\
\midrule
\multirow{2}{*}{\AdamW}
 & std & 3.692 / 3.963 & 3.610 / 3.865 & 3.579 / \textbf{3.840} & 3.642 / 3.902 \\
 & sel & 3.733 / 3.850 & 3.650 / 3.771 & 3.616 / \textbf{3.745} & 3.949 / 4.080 \\
\addlinespace[2pt]
\multirow{2}{*}{\AdEMAMix}
 & std & 3.680 / 3.950 & 3.617 / 3.885 & 3.599 / \textbf{3.863} & 4.206 / 4.446 \\
 & sel & 3.724 / 3.855 & 3.659 / 3.780 & 3.635 / \textbf{3.763} & 4.484 / 4.646 \\
\addlinespace[2pt]
\multirow{2}{*}{\Sophia}
 & std & 3.796 / 4.062 & 3.687 / 3.947 & 3.647 / \textbf{3.908} & 3.730 / 3.979 \\
 & sel & 3.828 / 3.950 & 3.724 / 3.856 & 3.679 / \textbf{3.804} & 3.737 / 3.891 \\
\addlinespace[2pt]
\multirow{2}{*}{\DMuon}
 & std & 3.777 / 4.043 & 3.658 / 3.931 & 3.607 / \textbf{3.870} & 3.622 / 3.877 \\
 & sel & 3.789 / 3.971 & 3.675 / 3.851 & 3.620 / \textbf{3.794} & 3.613 / 3.795 \\
\addlinespace[2pt]
\multirow{2}{*}{\Lion}
 & std & 3.611 / 3.867 & 3.585 / \textbf{3.852} & 3.603 / 3.862 & 5.697 / 5.946 \\
 & sel & 3.638 / 3.841 & 3.609 / \textbf{3.834} & 3.629 / 3.835 & 5.562 / 5.742 \\
\addlinespace[2pt]
\multirow{2}{*}{\Signum}
 & std & 3.840 / 4.099 & 3.740 / 3.996 & 3.709 / \textbf{3.964} & 3.867 / 4.107 \\
 & sel & 3.858 / 4.059 & 3.761 / 3.967 & 3.727 / \textbf{3.934} & 3.819 / 4.009 \\
\addlinespace[2pt]
\multirow{2}{*}{\SGD}
 & std & 6.996 / 7.190 & 6.878 / 7.073 & 6.728 / 6.930 & 6.558 / \textbf{6.763} \\
 & sel & 6.992 / 7.160 & 6.885 / 7.054 & 6.722 / 6.905 & 6.506 / \textbf{6.648} \\
\addlinespace[2pt]
\bottomrule
\end{tabular}
\end{table}

\begin{figure}[ht!]
\centering
\includegraphics[width=0.88\linewidth]{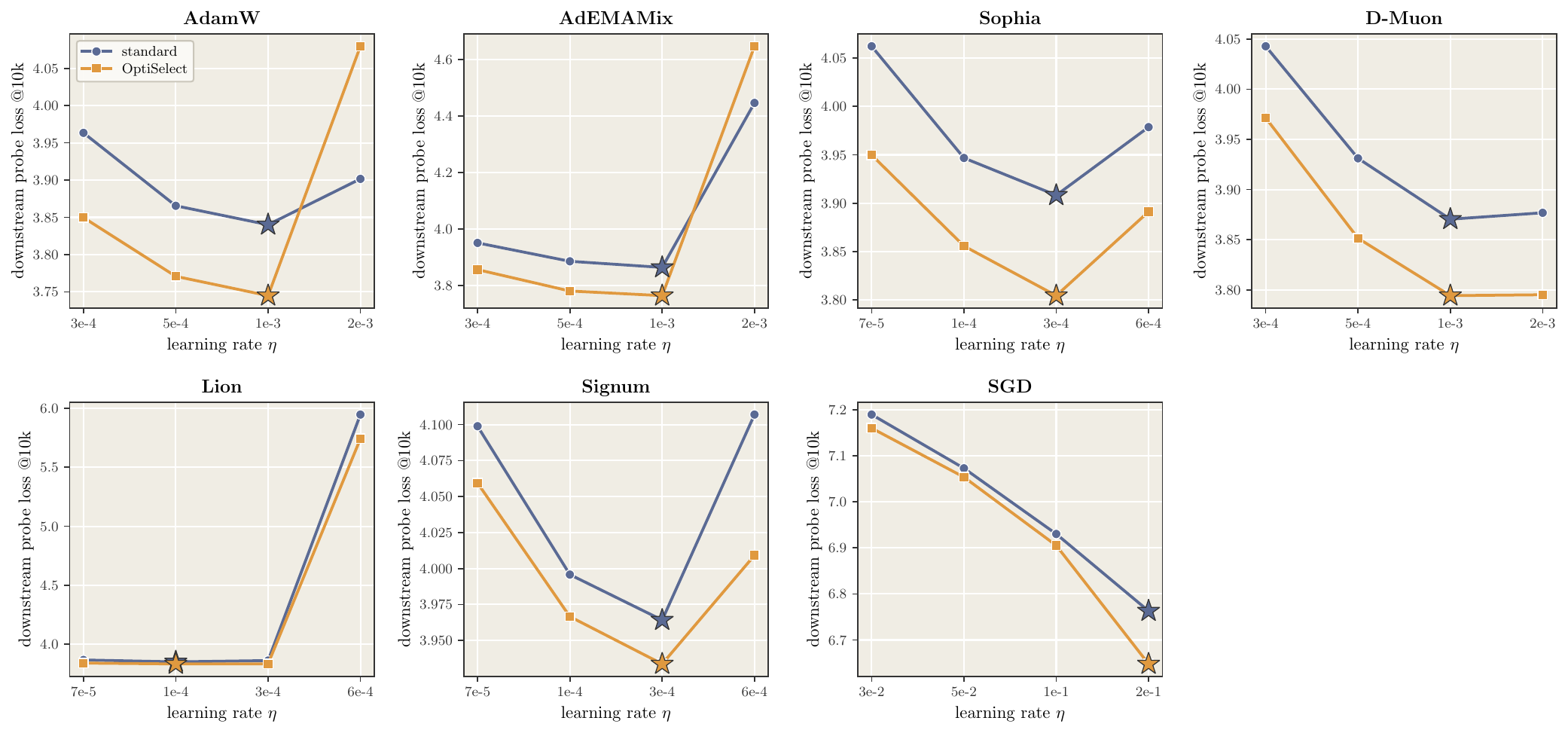}
\caption{\textbf{Downstream loss in 124M learning-rate sweep.} Stars $\star$ mark the optimum for each optimizer.}
\label{fig:lrsweep_downstream}
\end{figure}

According to~\autoref{tab:lr_sweep} and~\autoref{fig:lrsweep_downstream}, we find \textbf{The optimal learning rate is transferrable on most optimizers.} All seven optimizers select the \emph{same} learning rate under standard training and \textit{OptiSelect} paradigm according to its downstream performance.
Therefore, a run can adopt the optimal learning rate from its standard-training rate in the \textit{OptiSelect} mode.

\begin{figure}[ht!]
\centering
\includegraphics[width=0.5\linewidth]{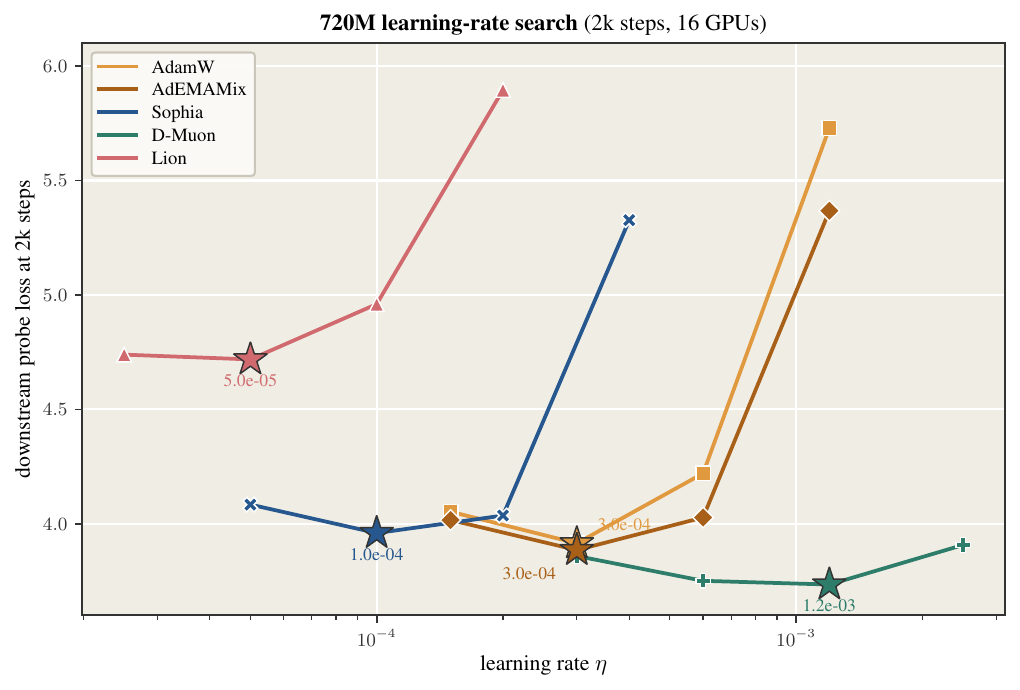}
\caption{\textbf{Downstream loss in 720M learning-rate search.} Stars $\star$ mark the optimum for each optimizer.}
\label{fig:lr720}
\end{figure}

\paragraph{720M Learning-Rate Search}
Given the transferability of learning rate, 
we swept five optimizers in \emph{standard} training mode at the full 720M configuration ($16$ GPUs, $256$ sequences per step, $2$k steps $=262$M tokens).
We select the optimum across four grids according to~\autoref{fig:lr720} and directly apply the same optimal learning rate on \textit{OptiSelect}.

\clearpage
\newpage
\subsection{Proxy Refresh Interval}
\label{app:refresh_sweep}

\begin{figure}[ht!]
\centering
\includegraphics[width=\linewidth]{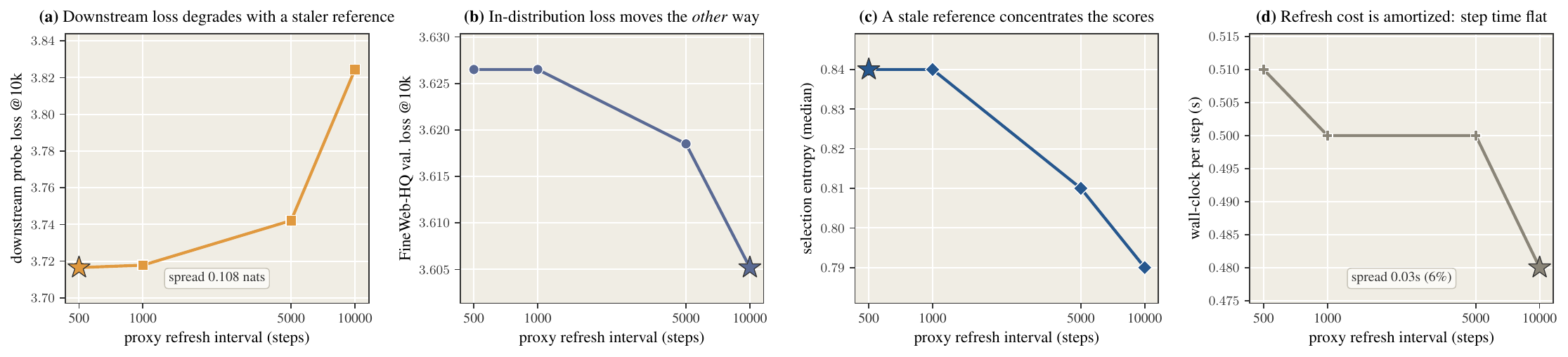}
\caption{\textbf{Ablation on Proxy-refresh Frequency.}
\AdamW\ at $\eta = 10^{-3}$, $10$k steps, varying only the refresh interval $\Delta$; stars mark the best cell.
(a)~\textbf{The downstream loss} degrades monotonically proxy-fresh frequency; while
(b)~\textbf{Held-out FineWeb-HQ loss} moves in an \emph{opposite} way.
(c)~\textbf{Selection entropy} falls as $\Delta$ grows, i.e.\ a stale reference makes the scores \emph{look} more decisive while aligning them with a stale target.
(d)~\textbf{Step time is flat to within $0.03$\,s ($6\%$)}: the refresh is amortized over $\Delta$ steps and never dominates.}
\label{fig:refresh_sweep}
\end{figure}

Recomputing $\bg_{\cV,t}$ costs a backward pass over the packed proxy set, so the engine caches it and reuses it for $\Delta$ steps.
This trades a per-step cost against a staleness bias: between refreshes the reference gradient is evaluated at parameters the model has since left.
We sweep $\Delta \in \{500, 1000, 5000, 10000\}$ on \AdamW\ at its best rate ($\eta = 10^{-3}$, $10$k steps, everything else held fixed) and report the results in~\autoref{fig:refresh_sweep}, \autoref{tab:refresh_sweep}. Since the cost is essentially free and the benefit saturates by $\Delta = 1000$ ($3.7178$ vs.\ $3.7165$ nats at $\Delta = 500$), we set $\Delta = 1000$ as the default.

\begin{table}[ht!]
\centering
\caption{Proxy refresh sweep, \AdamW\ at $\eta = 10^{-3}$, $10$k steps at 124M. $t_{\mathrm{step}}$ is measured as median wall-clock per optimizer step.}
\label{tab:refresh_sweep}
\begin{tabular}{lcccc}
\toprule
refresh $\Delta$ (steps) & $\mathbf{500}$ & $\mathbf{1000}$ & $5000$ & $10000$ \\
\midrule
downstream loss $\downarrow$   & \textbf{3.7165} & \textbf{3.7178} & 3.7421 & 3.8244 \\
FineWeb-HQ val.\ loss          & 3.6265 & 3.6265 & 3.6185 & \textbf{3.6052} \\
selection entropy              & 0.84 & 0.84 & 0.81 & 0.79 \\
$t_{\mathrm{step}}$ (s)        & 0.51 & 0.50 & 0.50 & 0.48 \\
\bottomrule
\end{tabular}
\end{table}

\clearpage
\newpage
\section{Extended Experimental Results}
\label{app:results}

We present an extended analysis on the experimental results from section \S\ref{sec:experiments}, at both 124M and 720M model scales.

\subsection{Complete scores at 124M}
\begin{figure}[ht!]
\centering
\includegraphics[width=\linewidth]{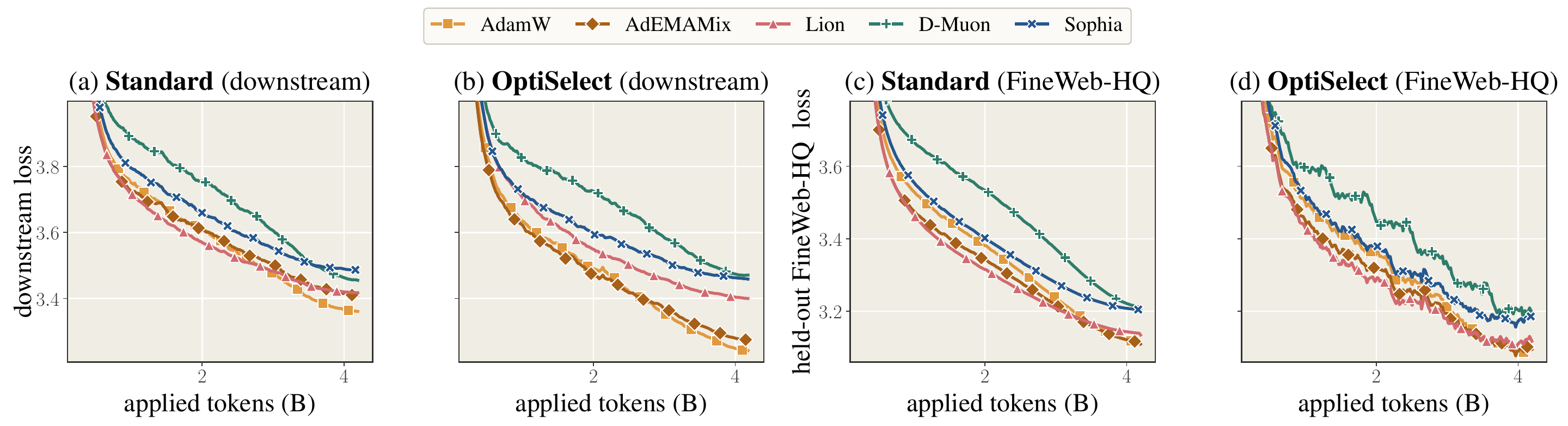}
\caption{\textbf{Downstream loss and Held-out Fineweb-HQ loss compared across all optimizers at 720M.} (i)~\textbf{Standard training}, (ii)~\textit{\textbf{OptiSelect}}.
}
\label{fig:full124m_all}
\end{figure}

\paragraph{Losses.} We directly compare across various optimizers under standard training and \textit{OptiSelect} in terms of average loss across all downstream tasks (\autoref{fig:full124m_all}a,b) and validation loss on a held-out set from FineWeb-HQ (\autoref{fig:full124m_all}c,d). The full loss values are presented in \autoref{tab:loss124}.

\autoref{fig:full124m_all} demonstrates that data selection improves downstream loss with every optimizers, while the ranking of optimizers is largely preserved at 124M, with \AdamW and \AdEMAMix takeing the leading position. 
In comparison, \Lion and \DMuon falls behind other optimizers in rank despite its improvements on raw downstream performance.
However, \textit{OptiSelect} does not affect the optimizer ranking upon training distribution, i.e. the ranks upon held-out FineWeb-HQ loss are constant between standard training and \textit{OptiSelect}.   

\paragraph{Accuracy.} 
\autoref{tab:acc124m} provides the complete accuracy scores across 9 in-domain and out-of-domain tasks.  
\textit{OptiSelect} outperforms standard training with the equal amount of trained tokens in both in-domain and out-of-domain tasks, while falling behind with an equal amount of read tokens ($\lambda\times$ of trained tokens). Across optimizers, \AdEMAMix achieves the best accuracy across in-domain tasks, while \AdamW achieves the best score on out-of-domain ones. In comparison, \Lion underperforms standard baseline in the average in-domain accuracy and \Sophia underperforms in the average accuracies of out-of domain tasks. 

\begin{table}[ht!]
\caption{\textbf{Downstream loss and Held-out Fineweb-HQ loss on 124M models.}}
\label{tab:loss124}
\centering
\small
\setlength{\tabcolsep}{4pt}
\begin{tabular}{l rrr rrr}
\toprule
& \multicolumn{3}{c}{\textbf{Held-out FineWeb-HQ loss}$\downarrow$} & \multicolumn{3}{c}{\textbf{Downstream loss}$\downarrow$} \\
\cmidrule(lr){2-4}\cmidrule(lr){5-7}
\textbf{Optimizer} & std@$128$k & sel@$128$k & $\Delta$  & std@$128$k & sel@$128$k & $\Delta$ \\
\midrule
\AdamW & \textbf{3.107} & 3.094 & -0.013 & \textbf{3.361} & \textbf{3.241} & -0.119 \\
\AdEMAMix & 3.108 & \textbf{3.093} & -0.014 & 3.410 & 3.274 & -0.135 \\
\Sophia & 3.198 & 3.179 & -0.019 & 3.486 & 3.458 & -0.028 \\
\Lion & 3.133 & 3.116 & -0.016 & 3.417 & 3.399 & -0.018 \\
\DMuon & 3.204 & 3.195 & -0.008  & 3.455 & 3.471 & \textcolor{red}{+0.015} \\
\bottomrule
\end{tabular}
\end{table}

\begin{table}[ht!]
\caption{\textbf{124M zero-shot accuracy (\%) per task.} \textit{OptiSelect} outperforms standard training with equal amount of trained tokens, while falls behind with the equal amount of read tokens ($\lambda\times$ of trained tokens). Bold marks the best performance scores. Red color marks the slots where \textit{OptiSelect} underperformed.}
\label{tab:acc124m}
\centering
\footnotesize
\setlength{\tabcolsep}{3pt}
\adjustbox{max width=0.99\textwidth}{
\begin{tabular}{ll rrrrr c rrrr c}
\toprule
& & \multicolumn{6}{c}{\textbf{In-distribution}} & \multicolumn{5}{c}{\textbf{Out-of-distribution}} \\
\cmidrule(lr){3-8}\cmidrule(lr){9-13}
\textbf{Optimizer} & \textbf{Checkpoint} & PIQA & HSwag & ARC-e & ARC-c & SciQ & \textbf{mean} & Wino & WSC & CSQA & OBQA & \textbf{mean} \\
\midrule
\AdamW & std@$128$k & 64.80 & 31.83 & 44.11 & 23.38 & 65.40 & 45.90 & 51.14 & 36.54 & 19.66 & 28.80 & 34.03 \\
 & sel@$128$k & 64.80 & 32.56 & 44.32 & 25.51 & 65.20 & \textbf{46.48} & 52.96 & 41.35 & 19.74 & 29.00 & \textbf{35.76} \\
 & sel@$64$k & 62.08 & 30.56 & 42.93 & 23.72 & 62.40 & 44.34 & 50.91 & 36.54 & 19.49 & 28.00 & 33.73 \\
\addlinespace[2pt]
\AdEMAMix & std@$128$k & 63.71 & 31.98 & 43.01 & 25.51 & 65.40 & 45.92 & 50.83 & 36.54 & 19.57 & 29.40 & 34.09 \\
 & sel@$128$k & 65.18 & 32.50 & 43.81 & 25.26 & 66.80 & \textbf{46.71} & 54.14 & 36.54 & 20.64 & 26.60 & \textbf{34.48} \\
 & sel@$64$k & 62.51 & 29.91 & 41.88 & 23.46 & 61.80 & 43.91 & 51.38 & 36.54 & 19.57 & 29.00 & 34.12 \\
\addlinespace[2pt]
\Sophia & std@$128$k & 62.95 & 31.27 & 41.71 & 24.32 & 58.80 & 43.81 & 49.64 & 38.46 & 19.66 & 27.80 & \textbf{33.89} \\
 & sel@$128$k & 62.40 & 31.26 & 43.22 & 24.15 & 62.80 & \textbf{44.77} & 50.28 & 36.54 & 19.57 & 26.40 & \textcolor{red}{33.20} \\
 & sel@$64$k & 62.35 & 29.58 & 41.12 & 25.34 & 60.20 & 43.72 & 50.59 & 36.54 & 19.57 & 28.00 & 33.68 \\
\addlinespace[2pt]
\Lion & std@$128$k & 63.44 & 32.39 & 43.18 & 23.21 & 65.00 & \textbf{45.44} & 50.75 & 36.54 & 19.49 & 27.60 & 33.60 \\
 & sel@$128$k & 63.71 & 32.10 & 42.47 & 25.51 & 62.20 & \textcolor{red}{45.20} & 51.46 & 37.50 & 19.57 & 28.00 & \textbf{34.13} \\
 & sel@$64$k & 61.86 & 30.17 & 40.99 & 24.23 & 59.70 & 43.39 & 52.41 & 36.54 & 19.57 & 27.60 & 34.03 \\
\addlinespace[2pt]
\DMuon & std@$128$k & 64.58 & 32.15 & 41.96 & 24.15 & 62.60 & 45.09 & 52.88 & 36.54 & 19.74 & 25.00 & 33.54 \\
 & sel@$128$k & 63.00 & 32.62 & 42.47 & 24.06 & 64.20 & \textbf{45.27} & 50.99 & 40.38 & 19.66 & 28.00 & \textbf{34.76} \\
& sel@$64$k$^{*}$ & 61.15 & 29.84 & 40.19 & 23.63 & 60.00 & 42.97 & 49.57 & 38.46 & 19.57 & 27.20 & 33.70 \\
\bottomrule
\end{tabular}
}
\end{table}

\clearpage
\newpage
\subsection{Complete scores at 720M}
\begin{figure}[ht!]
\centering
\includegraphics[width=\linewidth]{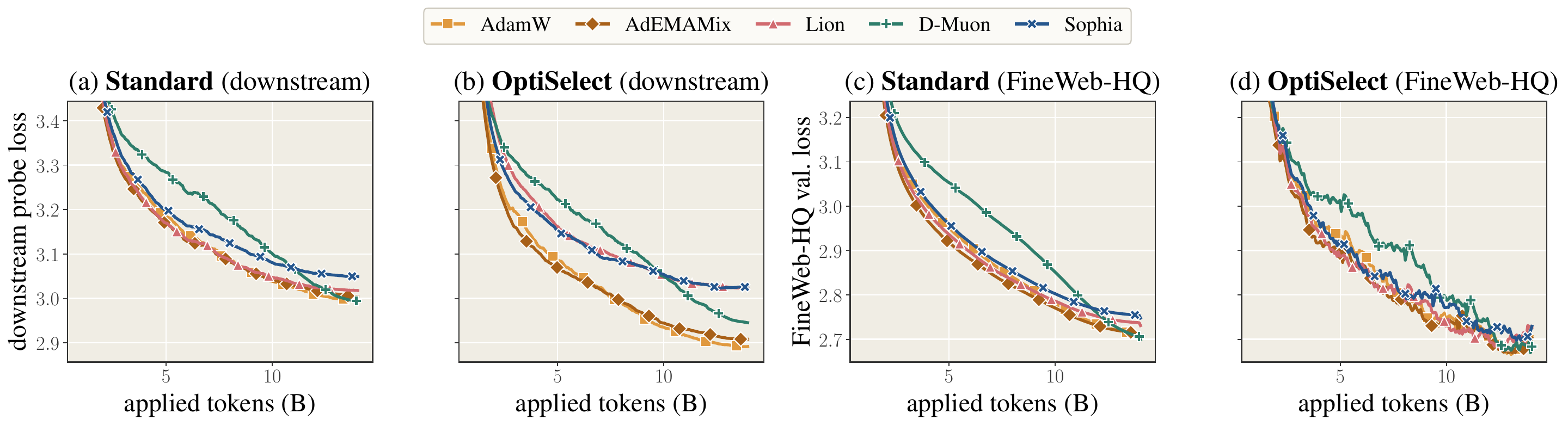}
\caption{\textbf{Downstream loss and Held-out Fineweb-HQ loss compared across all optimizers at 720M.} (i)~\textbf{Standard training}, (ii)~\textit{\textbf{OptiSelect}}.
}
\label{fig:full720m_all}
\end{figure}

\paragraph{Losses.}
According to \autoref{fig:full720m_all} and \autoref{tab:loss720}. At 720M, \DMuon ranks at the first under standard training while falling behind of \AdamW and \AdEMAMix with \textit{OptiSelect} in downstream losses. Whereas for held-out FineWeb-HQ loss, \Muon keeps its leading position while \AdamW, \AdEMAMix and \Lion demonstrate a degradation compared with standard training. 

\paragraph{Accuracy.}
\autoref{tab:acc720m} provides the complete accuracy scores across 9 in-domain and out-of-domain tasks at 720M. All the optimizers outperform standard training in in-domain accuracies, while \Muon and \AdEMAMix underperform in average out-of-domain accuracy. 

\begin{table}[ht!]
\caption{\textbf{Downstream loss and Held-out Fineweb-HQ loss on 720M models.}}
\label{tab:loss720}
\centering
\small
\setlength{\tabcolsep}{4pt}
\begin{tabular}{l rrr rrr}
\toprule
& \multicolumn{3}{c}{\textbf{Held-out FineWeb-HQ}} & \multicolumn{3}{c}{\textbf{Downstream probe}} \\
\cmidrule(lr){2-4}\cmidrule(lr){5-7}
\textbf{Optimizer} & std@$107$k & sel@$107$k & $\Delta$  & std@$107$k & sel@$107$k & $\Delta$ \\
\midrule
\AdamW & 2.704 & 2.714 & \textcolor{red}{+0.010}  & 2.995 & 2.891 & -0.104  \\
\AdEMAMix & 2.705 & 2.710 & \textcolor{red}{+0.005} & 3.004 & \textbf{2.908} & -0.096 \\
\Sophia & 2.745 & 2.736 & -0.009  & 3.049 & 3.026 & -0.023  \\
\Lion & 2.729 & 2.731 & \textcolor{red}{+0.002} & 3.018 & 3.022 & \textcolor{red}{+0.004} \\
\DMuon & 2.696 & \textbf{2.684} & -0.012 & 2.993 & 2.944 & -0.049 \\
\bottomrule
\end{tabular}
\end{table}

\begin{table}[ht!]
\caption{\textbf{720M zero-shot accuracy (\%), per task.}}
\label{tab:acc720m}
\centering
\footnotesize
\setlength{\tabcolsep}{3pt}
\adjustbox{max width=0.99\textwidth}{
\begin{tabular}{ll rrrrr c rrrr c}
\toprule
& & \multicolumn{6}{c}{\textbf{In-distribution}} & \multicolumn{5}{c}{\textbf{Out-of-distribution}} \\
\cmidrule(lr){3-8}\cmidrule(lr){9-13}
\textbf{Optimizer} & \textbf{Checkpoint} & PIQA & HSwag & ARC-e & ARC-c & SciQ & \textbf{mean} & Wino & WSC & CSQA & OBQA & \textbf{mean} \\
\midrule
\AdamW & std@$107$k & 69.45 & 46.03 & 53.89 & 28.37 & 75.85 & 54.72 & 50.87 & 45.19 & 19.49 & 32.50 & \textbf{37.01} \\
 & sel@$107$k & 70.51 & 47.05 & 56.44 & 28.75 & 76.30 & \textbf{55.81} & 52.88 & 36.54 & 20.72 & 33.00 & \textcolor{red}{35.79} \\
 & sel@$53.5$k & 68.72 & 41.35 & 52.31 & 27.39 & 73.00 & 52.55 & 49.96 & 36.54 & 19.74 & 30.20 & 34.11 \\
\addlinespace[2pt]
\AdEMAMix & std@$107$k & 69.29 & 46.04 & 54.19 & 27.60 & 74.50 & 54.32 & 52.49 & 42.31 & 19.57 & 32.60 & 36.74 \\
 & sel@$107$k & 70.13 & 47.07 & 55.43 & 27.82 & 76.00 & \textbf{55.29} & 52.80 & 54.81 & 18.59 & 34.00 & 40.05 \\
 & sel@$53.5$k & 68.61 & 42.27 & 51.52 & 27.13 & 73.40 & 52.59 & 51.22 & 59.62 & 20.07 & 32.40 & \textbf{40.83} \\
\addlinespace[2pt]
\Sophia & std@$107$k & 69.04 & 44.14 & 52.21 & 26.96 & 73.65 & 53.20 & 50.91 & 37.50 & 21.09 & 32.90 & 35.60 \\
 & sel@$107$k & 69.59 & 45.41 & 53.41 & 28.58 & 75.00 & \textbf{54.40} & 50.99 & 47.12 & 21.29 & 32.20 & \textbf{37.90} \\
 & sel@$53.5$k & 67.68 & 41.66 & 52.48 & 27.13 & 71.20 & 52.03 & 51.38 & 36.54 & 20.88 & 31.40 & 35.05 \\
\addlinespace[2pt]
\Lion & std@$107$k & 69.94 & 44.95 & 53.85 & 27.69 & 72.30 & 53.75 & 50.59 & 36.54 & 20.84 & 30.70 & 34.67 \\
 & sel@$107$k & 71.11 & 45.38 & 52.90 & 27.30 & 74.40 & \textbf{54.22} & 51.93 & 42.31 & 20.23 & 32.60 & \textbf{36.77} \\
 & sel@$53.5$k & 67.85 & 41.62 & 48.57 & 26.54 & 70.80 & 51.07 & 50.67 & 38.46 & 20.31 & 30.40 & 34.96 \\
\addlinespace[2pt]
\DMuon & std@$107$k & 70.67 & 47.49 & 54.84 & 28.24 & 73.55 & 54.96 & 52.92 & 58.65 & 20.43 & 32.50 & \textbf{41.13} \\
 & sel@$107$k & 70.57 & 49.91 & 55.51 & 29.52 & 76.60 & \textbf{56.42} & 54.06 & 45.19 & 20.48 & 33.60 & \textcolor{red}{38.33} \\
 & sel@$53.5$k & 67.57 & 41.09 & 47.64 & 26.79 & 70.30 & 50.68 & 52.96 & 36.54 & 19.25 & 31.00 & 34.94 \\
\bottomrule
\end{tabular}
}
\end{table}

\clearpage
\newpage
\section{Data Selection under Rephrasing}
\label{app:rephrase}
\subsection{Details on the Downstream Tasks}
\label{app:rephrase-format}
We categorize all five downstream tasks into two groups according their probing formats:
(1) \textit{\textbf{QA-formatted}}: 
A task is \emph{QA-formatted} if the prompt poses an explicit question and terminates with an answer slot, so the model is scored on the Question-Answering ability;
(2) \textit{\textbf{completion-formatted}} A task is is \emph{completion-formatted} if it feeds a plain textual stem to the model, and every option is a continuation of it. The model is scored on its ability on continuation.

According to the evaluation harness on each task, we categorize ARC-Easy, ARC-Challenge, PIQA, SciQ, CommonSense QA, Openbook QA as \textbf{QA-formatted}; and HellaSwag, WinoGrande, WSC are \textbf{completion-formatted}.

\subsection{Extended Results on Rephrasing}
We present the complete results on the \textit{OptiSelect} $\times$ Format rephrasing experiments in \S\ref{sec:finephrase}. According to \autoref{tab:finephrase_acc}, \textit{OptiSelect} consistently outperforms standard training across various rephrasing mixtures in terms of average accuracy on in-domain proxy tasks and out-of-domain benchmarks. 
Notably, both standard training and \textit{OptiSelect} achieve the best performance across all 9 tasks at $50\%$ rephrasing rates, which mixes original FineWeb and FinePhrase documents by $1:1$ ratio.
\begin{table}[ht!]
\caption{\textbf{Format rephrasing $\times$ \textit{OptiSelect}, zero-shot accuracy (\%).} }
\label{tab:finephrase_acc}
\centering
\small
\setlength{\tabcolsep}{3pt}
\adjustbox{max width=0.99\textwidth}{
\begin{tabular}{ll rrrrr c rrrr c c}
\toprule
& & \multicolumn{6}{c}{\textbf{In-Domain}} & \multicolumn{5}{c}{\textbf{Out-of-Domain}} & \\
\cmidrule(lr){3-8}\cmidrule(lr){9-13}
\textbf{Rephrased} & \textbf{Mode} & PIQA & HSwag & ARC-e & ARC-c & SciQ & \textbf{mean} & Wino & WSC & CSQA & OBQA & \textbf{mean} & overall-mean\\
\midrule
$0\%$ & standard & \textbf{64.74} & 29.59 & 47.81 & \textbf{20.65} & \textbf{75.10} & 47.58 & 51.14 & 36.54 & 19.66 & 16.60 & 30.98 & 40.20\\
 & OptiSelect & \textbf{64.74} & \textbf{30.01} & \textbf{50.04} & 20.31 & 74.10 & \textbf{47.84} & \textbf{52.01} & \textbf{44.23} & \textbf{19.74} & \textbf{17.40} & \textbf{33.35} & \textbf{41.40}\\
\addlinespace[2pt]
$50\%$ & standard & 64.47 & 29.87 & 51.94 & 23.04 & 75.90 & 49.04 & 50.28 & 38.46 & \textbf{19.57} & \textbf{18.00} & 31.58 & 41.28\\
 & OptiSelect & \textbf{64.91} & \textbf{30.13} & \textbf{53.24} & \textbf{23.29} & \textbf{77.00} & \textbf{49.72} & \textbf{52.17} & \textbf{40.38} & \textbf{19.57} & 17.60 & \textbf{32.43} & \textbf{42.03}\\
\addlinespace[2pt]
$100\%$ & standard & 63.71 & 30.05 & 54.00 & \textbf{23.21} & 75.60 & 49.31 & 49.96 & \textbf{36.54} & 19.49 & 18.40 & 31.10 & 41.22\\
 & OptiSelect & \textbf{64.91} & \textbf{30.23} & \textbf{54.38} & 22.70 & \textbf{77.70} & \textbf{49.98} & \textbf{52.17} & \textbf{36.54} & \textbf{19.74} & \textbf{19.40} & \textbf{31.96} & \textbf{41.97}\\
\bottomrule
\end{tabular}
}
\end{table}

\end{document}